\documentclass[journal]{IEEEtran}
\usepackage{graphicx}
\usepackage{amsmath, amssymb}
\usepackage{caption, subcaption}
\usepackage{multirow}
\usepackage{soul,color}
\usepackage{cite}
\usepackage[linesnumbered,ruled,vlined]{algorithm2e}
\begin{document}
%%%%%%%%%%%%%%%%%%%%%%%%%%%%%%%
\title{Importance of Disjoint Sampling in Conventional and Transformer Models for Hyperspectral Image Classification}
%%%%%%%%%%%%%%%%%%%%%%%%%%%%%%%
\author{Muhammad Ahmad, Manuel Mazzara, Salvatore Distifano
\thanks{M. Ahmad is with the Department of Computer Science, National University of Computer and Emerging Sciences, Islamabad, Chiniot-Faisalabad Campus, Chiniot 35400, Pakistan, and Dipartimento di Matematica e Informatica---MIFT, University of Messina, Messina 98121, Italy; (e-mail: mahmad00@gmail.com)}
\thanks{M. Mazzara is with the Institute of Software Development and Engineering, Innopolis University, Innopolis, 420500, Russia. (e-mail: m.mazzara@innopolis.ru)}
\thanks{S. Distefano is with  Dipartimento di Matematica e Informatica---MIFT, University of Messina, Messina 98121, Italy. (e-mail: sdistefano@unime.it)}
}
%%%%%%%%%%%%%%%%%%%%%%%%%%%%%%%
\markboth{Journal of \LaTeX\ Class Files,}
{Ahmad \MakeLowercase{\textit{et al.}}:}
%%%%%%%%%%%%%%%%%%%%%%%%%%%%%%%
\maketitle
%%%%%%%%%%%%%%%%%%%%%%%%%%%%%%%
\begin{abstract}
Disjoint sampling is critical for rigorous and unbiased evaluation of state-of-the-art (SOTA) models. When training, validation, and test sets overlap or share data, it introduces a bias that inflates performance metrics and prevents accurate assessment of a model's true ability to generalize to new examples. This paper presents an innovative disjoint sampling approach for training SOTA models on Hyperspectral image classification (HSIC) tasks. By separating training, validation, and test data without overlap, the proposed method facilitates a fairer evaluation of how well a model can classify pixels it was not exposed to during training or validation. Experiments demonstrate the approach significantly improves a model's generalization compared to alternatives that include training and validation data in test data. By eliminating data leakage between sets, disjoint sampling provides reliable metrics for benchmarking progress in HSIC. Researchers can have confidence that reported performance truly reflects a model's capabilities for classifying new scenes, not just memorized pixels. This rigorous methodology is critical for advancing SOTA models and their real-world application to large-scale land mapping with Hyperspectral sensors. The source code is be available at https://github.com/mahmad00/Disjoint-Sampling-for-Hyperspectral-Image-Classification.
\end{abstract}
%%%%%%%%%%%%%%%%%%%%%%%%%%%%%%%
%%%%%%%%%%%%%%%%%%%%%%%%%%%%%%%
\IEEEpeerreviewmaketitle
%%%%%%%%%%%%%%%%%%%%%%%%%%%%%%%
\section{Introduction}

\IEEEPARstart{H}{yperspectral Image Classification (HSIC)} plays a pivotal role in various domains such as remote sensing \cite{ahmad2021hyperspectral}, earth observation \cite{lodhi2018hyperspectral}, urban planning \cite{li2024HD}, agriculture \cite{lu2020recent}, forestry \cite{adao2017hyperspectral}, target/object detection \cite{li2023lrr}, mineral exploration \cite{bedini2017use}, environmental monitoring \cite{weber2018hyperspectral, stuart2019hyperspectral}, climate change \cite{pande2023application} food processing \cite{khan2021hyperspectral, khan2020hyperspectral}, bakery products \cite{saleem2020prediction}, bloodstain identification \cite{butt2022fast, zulfiqar2021hyperspectral}, and meat processing \cite{ayaz2020hyperspectral, ayaz2020myoglobin}. The rich spectral information provided by HSIs, often spanning hundreds of narrow bands, presents both challenges and opportunities for effective classification \cite{hong2024spectralgpt}. In recent years, Convolutional Neural Networks (CNNs) \cite{ahmad2020fast}, Graph-based CNNs (GCNNs) \cite{9170817, 10409250} and Transformers \cite{yao2023extended, ahmad2024waveformer} models have demonstrated remarkable success in various computer vision tasks, prompting researchers to explore their potential in HSI analysis. However, the computational complexity and memory requirements associated with processing the entire hyperspectral (HS) data pose significant challenges \cite{ahmad2023sharpend}. In this context, this work focuses on the importance of disjoint sampling in training, validation, and testing such models for HSIC.

Random sampling for data splitting can lead to several issues. It can result in non-representative training, validation, and test sets, causing models to overfit or underfit the data. Different random splits produce inconsistent results, making it hard to draw meaningful conclusions \cite{AHMAD2021166267,AHMAD201786}. Random sampling offers no control over data distribution, introducing bias in imbalanced datasets. It hinders the reproducibility of experimental results and limits the exploration of data relationships. Disjoint sampling is an important yet often overlooked consideration when evaluating spatial-spectral classification models. As demonstrated by the works \cite{10409250, ahmad2022disjoint, 8046029, ahmad2019spatial, rs15225331, ahmad2021hyperspectral, hong2024multimodal}, traditional evaluations using overlapping training and test samples can lead to biased results and unfair assessments of model performance. Their work presents a systematic evaluation of conventional and 3D convolutional neural network (CNN) models trained on disjoint HS samples and evaluated on disjoint validation sets. In addition to reporting overall accuracy, they analyze statistical measures like precision, recall, and F1-score to better validate the model's generalized performance. Their results show that models trained on disjoint samples generalize better to unseen data, compared to models trained on the entire dataset. Their findings highlight the importance and feasibility of using disjoint sampling approaches to robustly evaluate conventional and 3D CNN models, which have large capacities and could overfit standard evaluations without adequate disjoint training and testing. 

Even though these methodologies \cite{10409250, 8046029, rs15225331} meticulously employ disjoint sets for training, validation, and testing their models, there's a notable inconsistency in their approach when it comes to generating land-cover maps. Specifically, many of these methods deviate from the disjoint sampling principle by utilizing the entire HSI dataset for land-cover classification (Thematic Maps). This practice introduces a conflict between the reported accuracy and the methodology employed. To address this inconsistency, it is essential to advocate for the use of disjoint test sets exclusively for generating land-cover maps. By doing so, the evaluation process aligns more closely with the principles of unbiased model assessment. It ensures that the model is confronted with truly unseen data during the map generation phase, fostering a more accurate representation of its real-world performance. This refined approach contributes to the transparency and reliability of reported accuracies, ultimately enhancing the credibility of land-cover mapping results in hyperspectral imaging studies.

Moreover, disjoint sampling is essential for training and evaluating deep models in HSI-based land cover classification. This method involves carefully selecting diverse and representative samples from various regions, land cover types, and environmental conditions to overcome biased or non-representative training data limitations \cite{hong2023cross}. It ensures the model learns robust features, enhancing classification performance and adaptability to unseen data. Additionally, disjoint sampling facilitates fair and accurate model evaluation by keeping training, validation, and testing samples separate \cite{yao2022semi}. Furthermore, it enhances model interpretability as diverse samples help the model learn distinctive spectral characteristics associated with different land cover classes. This interpretability is crucial for building trust in the model's predictions and ensuring successful deployment in real-world applications. Furthermore, disjoint sampling is crucial in training SOTA models for HSIC, notably for CNN and Spatial-Spectral Transformer-based models. It enhances generalization, ensures fair evaluation, and enables result interpretability. This sampling strategy empowers researchers to maximize the potential of HS data, creating accurate models for diverse applications. The use of disjoint training, validation, and test samples is imperative in HSIC for various reasons, such as:

\textbf{Unbiased Evaluation and Hyperparameter Tuning:} It is crucial to evaluate HSIC models using completely separate and disjoint data for training, validation, and testing in order to properly assess a model's true ability to generalize to new unknown examples \cite{ahmad2022disjoint}. The validation set plays a critical role in systematically developing a customized model for optimal performance. During training, numerous hyperparameters shape characteristics such as architecture and complexity but are not informed by the training data, necessitating impartial evaluation of candidate models to select the best-performing configuration. Through an iterative process of training models with varying hyperparameters on the training set and then evaluating their generalization ability solely on the held-out validation set, the ideal hyperparameter values producing a model specifically tailored to the nuances captured in the full dataset can be identified. By maintaining a rigorous separation of samples, we obtain an honest measure of the model's predictive power on genuinely new observations and can be confident that it has learned to generalize, not memorize, enabling accurate classification of instances well beyond the initial data used to build the model \cite{ahmad2021hyperspectral}.

\textbf{Preventing Data Leakage and Mitigating Overfitting:} Maintaining disjoint samples for training, validation, and testing is crucial to obtaining an accurate evaluation of a model's true generalization performance \cite{10409250, yao2022semi}. Utilizing disjoint subsets of the data for each stage of model development plays a critical role in enhancing generalization performance. By training on distinct partitions in an iterative fashion, the model is compelled to deduce underlying patterns common across diverse examples rather than potentially noisome idiosyncrasies within a single fixed training sample \cite{10423094, Adari2024}. This discourages the memorization of spurious characteristics unique to a single snapshot of data and instead cultivates the ability to accurately handle a broader array of presentations, both seen and novel. In this paper, we have made the following \textbf{\textit{contributions}}: 

\begin{enumerate}
    \item We present a novel approach for generating disjoint train, validation, and test splits for HSIC. By ensuring the disjoint splits, the approach eliminates data leakage between subsets which can bias performance evaluations.
    \item The proposed technique also provides a practical implementation for creating disjoint train, validation, and test splits from ground truth data in HSIC. This allows researchers to obtain unbiased performance evaluations and reliable comparisons between HSIC models.
    \item By offering a standardized approach for creating evaluation splits, the proposed technique enhances the reproducibility and transparency of HSIC research. It fosters a more rigorous and standardized evaluation of classification models in the HSI domain.
\end{enumerate}

In the subsequent section, we present our proposed methodology. Section III provides the experimental results and their discussion, and Section IV concludes the paper.

%%%%%%%%%%%%%%%%%%%%%%%%%%
\section{Proposed Methodology}

Let's consider HSI data composed of $B$ spectral bands, each with a spatial resolution of $M \times N$ pixels. The HSI data cube, denoted as $X \in \mathbb{R}^{(M \times N \times B)}$, is initially partitioned into overlapping 3D patches \cite{ahmad2024waveformer, 8736016}. Each patch is centered at a spatial location $(\alpha, \beta)$ and covers a spatial extent of $S \times S$ pixels across all $B$ bands. The total number of 3D patches ($m$) extracted from $X$ (i.e., $X \in \mathbb{R}^{(S \times S \times B)}$) is given by $(M-S+1) \times (N-S+1)$. A patch located at $(\alpha, \beta)$ is represented as $P_{\alpha, \beta}$ and spans spatially from $\alpha - \frac{S-1}{2}$ to $\alpha + \frac{S-1}{2}$ in width and $\beta - \frac{S-1}{2}$ to $\beta + \frac{S-1}{2}$ in height. The labeling of these patches is determined by the label assigned to the central pixel within each patch as described in Algorithm 1. 

%%%%%%%%%%%%%%%%%%%%%%%%%%
\begin{algorithm}
    \caption{Create 3D HSI Patches}
    \SetKwInOut{Input}{Input}
    \SetKwInOut{Output}{Output}
    \Input{HSI, GT, WS}
    r, c, b $\leftarrow$ dimensions of HSI\;
    margin $\leftarrow$ WS / 2\;
    Pad $\leftarrow$ pad HSI with zeros on all sides\;
    Cubes $\leftarrow$ create an array of size (r $\times$ c, WS, WS, b)\;
    Labels $\leftarrow$ create an array of size (r $\times$ c) \;
    patchIndex $\leftarrow$ 0\;
    \For{rr $\leftarrow$ margin \KwTo rr + margin}{
        \For{cc $\leftarrow$ margin \KwTo cc + margin}{
            cube $\leftarrow$ select a sub-array from Pad
            Cubes[patchIndex, :, :, :] $\leftarrow$ cube\;
            Labels[patchIndex] $\leftarrow$ GT[rr - margin, cc - margin]\;
            patchIndex $\leftarrow$ patchIndex + 1\;
        }
    }   
    \KwRet{Cubes, Labels}\;
\label{Algo1}
\end{algorithm}
%%%%%%%%%%%%%%%%%%%%%%%%%%

%%%%%%%%%%%%%%%%%%%%%%%%%%
\begin{figure*}[!hbt]
\centering
\includegraphics[width=0.98\textwidth]{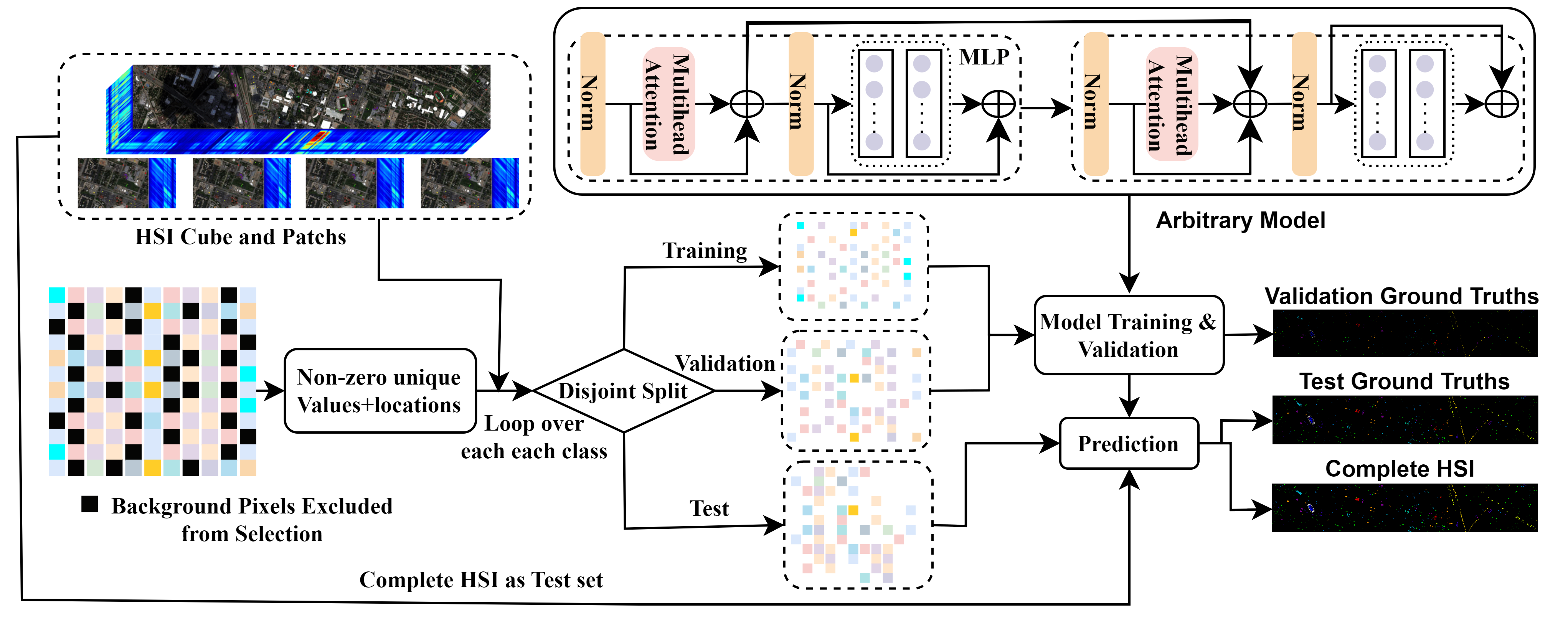}
\caption{The HSI cube is initially divided into overlapping 3D patches, as described in Algorithm 1. Each patch is centered at a spatial point and covers a $WS \times WS$ pixel extent across all spectral bands. These patches are then used in Algorithm 2 to create a disjoint train, validation, and test splits based on the geographical locations of the HSI samples. The selected samples are inputted into various models for feature learning and optimization. The processed features are subsequently passed through a fully connected layer for classification, and the softmax function is applied to generate class probability distributions. These distributions are used to generate the final ground truth maps for the disjoint validation, disjoint test, and full HSI test sets.}
\label{Fig1}
\end{figure*}
%%%%%%%%%%%%%%%%%%%%%%%%%%

The 3D patches extracted from the HSI are used to generate separate training, validation, and test sets using the proposed splitting algorithm. The key algorithm, titled "Disjoint Train, Validation, and Test Split", handles dividing the HSI data into the respective portions. It takes the ground truth labels (GT) and ratios for the test and validation sets (teRatio and vrRatio) as inputs. The unique values in the GT labels and their frequency counts are identified, excluding zeros. An iterative process is then used to create disjoint training, validation, and test sets based on these unique values and their indices. The resulting indices are utilized to extract and organize the corresponding hyperspectral cubes and labels for each set. This ensures the subsets are separate while maintaining the integrity of spectral classes during model training and evaluation. The algorithm outputs the training, validation, and test samples along with their matching labels. This partitioning approach contributes to the robustness and reliability of the subsequent analysis.

%%%%%%%%%%%%%%%%%%%%%%%%%%
\begin{algorithm}
\caption{Disjoint Train, Validation, and Test Split}
\SetKwInOut{Input}{Input}
\SetKwInOut{Output}{Output}
\Input{GT, p, m}
% \tcp{Labels obtained from Algorithm 1}
flattened $\gets$ GT.flatten()\;
unique $\gets$ np.unique(flattened)\;
nonzero\_indices $\gets$ np.where(unique $\neq$ 0)[0]\;
unique $\gets$ unique[nonzero\_indices]\;
TrInd, VaInd, TeInd $\gets$ [], [], []\;
\For{value $\leftarrow$ unique}{
C\_ind $\gets$ np.where(flattened == value)[0]\;
Tr\_ind, Te\_ind $\gets$ train\_test\_split(C\_ind, p)\;
Tr\_ind, V\_ind $\gets$ train\_test\_split(Tr\_ind, m)\;
TrInd.extend(Tr\_ind)\;
VaInd.extend(V\_ind)\;
TeInd.extend(Te\_ind)\;
}
% Train Labels $\gets$ Labels[TrInd] - 1\;
% Validation Labels $\gets$ Labels[VaInd] - 1\;
% Test Labels $\gets$ Labels[TeInd] - 1\;
% Train Samples = Cubes[TrInd]\;
% Validation Samples = Cubes[VaInd]\;
% Test Samples = Cubes[TeInd]\;
\end{algorithm}
%%%%%%%%%%%%%%%%%%%%%%%%%%

Let us consider that $n$, $m$, and $p$ represent the finite numbers of labeled training, validation, and test samples, respectively, selected from $X$ (HSI dataset) to form the training set $D_{TR} = {(x_i, y_i)}_{i=1}^n$ and the validation set $D_V = {(x_i, y_i)}_{i=1}^m$. The remaining samples constitute the test set $D_{TE} = {(x_i, y_i)}_{i=1}^p$. It is important to note that the intersection of the training set, validation set, and test set, denoted as $D_{TR} \cap D_V \cap D_{TE}$, is an empty set ($\phi$), ensuring the distinctiveness of the samples in each set as shown in Algorithm 2 and Figure \ref{Fig1}.

%%%%%%%%%%%%%%%%%%%%%%%%%%
\section{Experimental Results and Discussion}

%%%%%%%%%%%%%%%%%%%%%%%%%%
\subsection{Experimental Datasets}

\begin{figure*}[!hbt]
    \centering
    \includegraphics[width=0.98\textwidth]{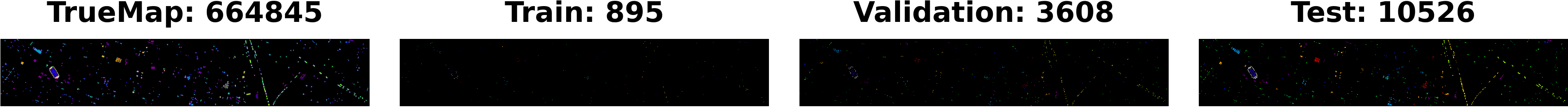}
    \caption{University of Houston Dataset: Geographical locations of the disjoint train, validation, and test samples presented in Table \ref{Tab2}. }
    \label{Fig3}
\end{figure*}

In order to further highlight the importance and the proposed procedure of disjoint sampling in HSIC, the following datasets are used. \textbf{The University of Houston:} The University of Houston HSI dataset consists of 144 spectral bands spanning wavelengths from 380 nm to 1050 nm, the dataset encompasses an imaged spatial region measuring 349 x 1905 pixels at a resolution of 2.5 meters per pixel. Additionally, the dataset annotates 15 labeled classes pertaining to urban land use and land cover types. The disjoint train, validation, and test samples are presented in Table \ref{Tab2} and Figure \ref{Fig3}.

%%%%%%%%%%%%%%%%%%%%%%%%%%
\begin{table}[!hbt]
    \centering
    \caption{University of Houston Dataset: Disjoint sets of training (Tr), validation (Va), and test (Te) samples were chosen, with their geographical locations (Excluding background samples) illustrated in Figure \ref{Fig3}, to train various SOTA models.}
    \resizebox{\columnwidth}{!}{\begin{tabular}{c|c|c|c|c|c|c|c} \hline
        \textbf{Class} & \textbf{Tr} & \textbf{Va} & \textbf{Te} & \textbf{Class} & \textbf{Tr} & \textbf{Va} & \textbf{Te}\\ \hline
        Healthy grass & 75 & 300 & 876 & Road & 75 & 300 & 877 \\
        Stressed grass & 75 & 301 & 878 & Highway & 73 & 295 & 859 \\
        Synthetic grass & 41 & 168 & 488 & Railway & 74 & 296 & 865 \\ 
        Trees & 74 & 299 & 871 & Parking Lot 1 & 73 & 296 & 864 \\ 
        Soil & 74 & 298 & 870 & Parking Lot 2 & 28 & 112 & 329 \\
        Water & 19 & 78 & 228 & Tennis Court & 25 & 103 & 300 \\ 
        Residential & 76 & 304 & 888 & Running Track & 39 & 159 & 462 \\ 
        Commercial & 74 & 299 & 871 & - & - & - & - \\ \hline 
    \end{tabular}}
    \label{Tab2}
\end{table}
%%%%%%%%%%%%%%%%%%%%%%%%%%
%%%%%%%%%%%%%%%%%%%%%%%%%%

\begin{figure}[!hbt]
    \centering
    \includegraphics[width=0.48\textwidth]{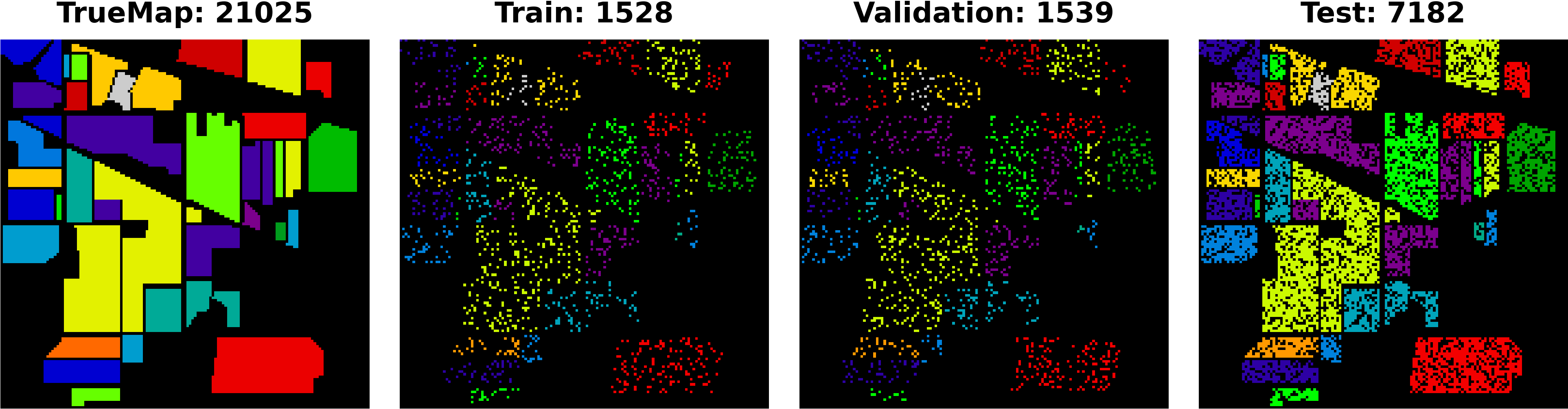}
    \caption{Indian Pines: Geographical locations of the disjoint train, validation, and test samples presented in Table \ref{Tab1}. }
    \label{Fig2}
\end{figure}

\textbf{Indian Pines:} The Indian Pines dataset was collected by the Airborne Visible/Infrared Imaging Spectrometer (AVIRIS) over an agricultural site in Northwestern Indiana. It consists of 145 x 145 pixels with spectral information across 224 narrow bands ranging from 0.4 to 2.5 micrometers. The major land cover classes in the dataset included agricultural land, forest, highways, rail lines, low-density housing, and built structures separated by smaller roads. Crops such as corn and soybeans covered less than 5\% of typical growing areas as the June image showed early stages of development. Ground truths designate 16 non-mutually exclusive classes. The number of bands was reduced to 200 by removing wavelengths associated with water absorption. The disjoint train, validation, and test samples are presented in Table \ref{Tab1} and Figure \ref{Fig2}. 

%%%%%%%%%%%%%%%%%%%%%%%%%%
\begin{table}[!hbt]
    \centering
    \caption{Indian Pines Dataset: Disjoint sets of training (Tr), validation (Va), and test (Te) samples were chosen, with their geographical locations (Excluding background samples) illustrated in Figure \ref{Fig2}, to train various SOTA models.}
    \resizebox{\columnwidth}{!}{\begin{tabular}{c|c|c|c|c|c|c|c} \hline 
        \textbf{Class} & \textbf{Tr} & \textbf{Va} & \textbf{Te} & \textbf{Class} & \textbf{Tr} & \textbf{Va} & \textbf{Te}\\ \hline

        Alfalfa & 6 & 7 & 33 & Oats & 3 & 3 & 14 \\ 
        Corn-notill & 214 & 214 & 1000 & Soybean-notill & 145 & 146 & 681 \\
        Corn-mintill & 124 & 125 & 581 & Soybean-mintill & 368 & 368 & 1719 \\
        Corn & 35 & 36 & 166 & Soybean-clean & 88 & 89 & 416 \\
        Grass-pasture & 72 & 72 & 339 & Wheat & 30 & 31 & 144 \\
        Grass-trees & 109 & 110 & 511 & Woods & 189 & 190 & 886 \\
        Grass-mowed & 4 & 4 & 20 & Buildings & 57 & 58 & 271 \\
        Hay-windrowed & 71 & 72 & 335 & Stone-Steel  & 13 & 14 & 66 \\ \hline 
    \end{tabular}}
    \label{Tab1}
\end{table}
%%%%%%%%%%%%%%%%%%%%%%%%%%
%%%%%%%%%%%%%%%%%%%%%%%%%%

\begin{figure}[!hbt]
    \centering
    \includegraphics[width=0.48\textwidth]{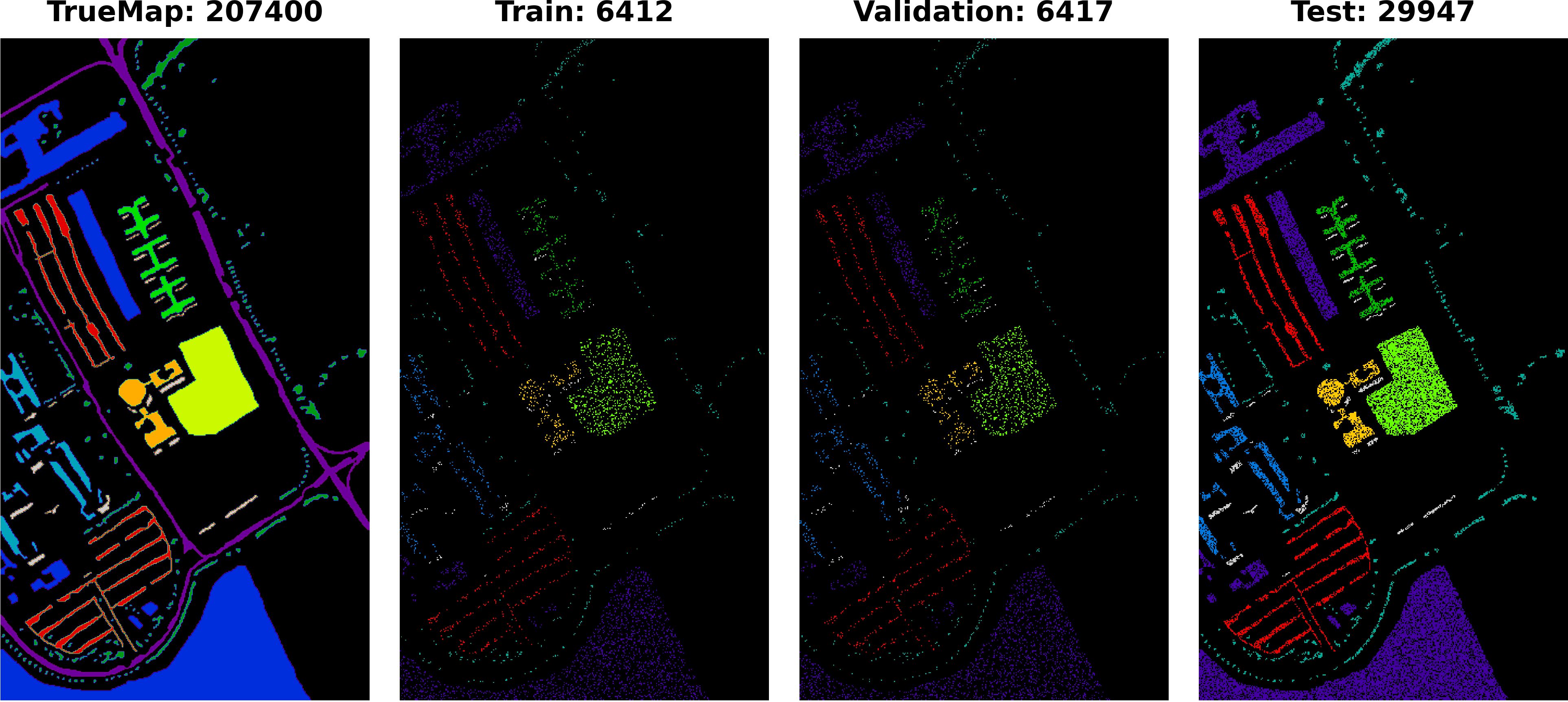}
    \caption{Pavia University Dataset: Geographical locations of the disjoint train, validation, and test samples presented in Table \ref{Tab3}.}
    \label{Fig4}
\end{figure}

\textbf{Pavia University:} The Pavia University HSI dataset was captured using the reflective optics system imaging spectrometer, this dataset consists of an image with 610×340 pixels and 115 spectral bands. It has 9 classes of urban materials - including asphalt, meadows, gravel, trees, metal sheets, bare soil, bitumen, brick, and shadows - comprising 42,776 labeled samples in total. The disjoint train, validation, and test samples are presented in Table \ref{Tab3} and Figure \ref{Fig4}.

%%%%%%%%%%%%%%%%%%%%%%%%%%
\begin{table}[!hbt]
    \centering
    \caption{Pavia University Dataset: Disjoint sets of training (Tr), validation (Va), and test (Te) samples were chosen, with their geographical locations (Excluding background samples) illustrated in Figure \ref{Fig4}, to train various SOTA models.}
    \begin{tabular}{c|c|c|c|c|c|c|c} \hline
        \textbf{Class} & \textbf{Tr} & \textbf{Va} & \textbf{Te} & \textbf{Class} & \textbf{Tr} & \textbf{Va} & \textbf{Te}\\ \hline
        Asphalt & 994 & 995 & 4642 & Soil & 754 & 754 & 3521 \\
        Meadows & 2797 & 2797 & 13055 & Bitumen & 199 & 200 & 931 \\
        Gravel & 314 & 315 & 1470 & Bricks & 552 & 552 & 2578 \\
        Trees & 459 & 460 & 2145 & Shadows& 142 & 142 & 663 \\
        Painted & 201 & 202 & 942 & - & - & - & - \\ \hline 
    \end{tabular}
    \label{Tab3}
\end{table}
%%%%%%%%%%%%%%%%%%%%%%%%%%
%%%%%%%%%%%%%%%%%%%%%%%%%%

\begin{figure}[!hbt]
    \centering
    \includegraphics[width=0.48\textwidth]{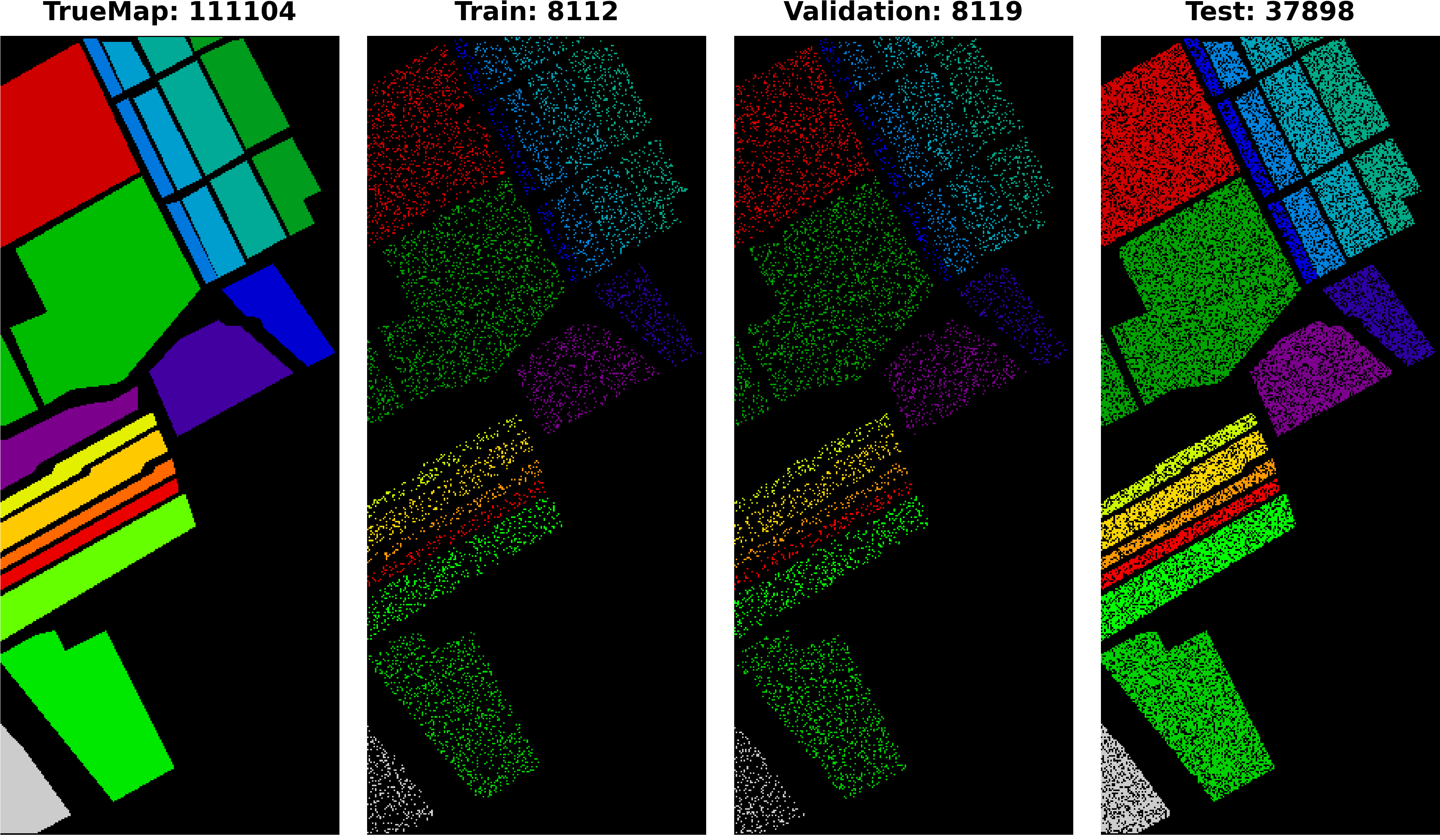}
    \caption{Salinas Dataset: Geographical locations of the disjoint train, validation, and test samples presented in Table \ref{Tab4}.}
    \label{Fig5}
\end{figure}

\textbf{Salinas:} The Salinas HSI dataset is collected using the 224-band AVIRIS sensor over Salinas Valley, California, this dataset is characterized by high spatial resolution at 3.7 meters per pixel. The study area encompasses 512 lines by 217 samples after removing 20 bands obscured by water absorption. Land cover types within the dataset include vegetables, bare soils, and vineyard fields. The Salinas ground truth annotates 16 classes. The disjoint train, validation, and test samples are presented in Table \ref{Tab4} and Figure \ref{Fig5}.

%%%%%%%%%%%%%%%%%%%%%%%%%%
\begin{table}[!hbt]
    \centering
    \caption{Salinas Dataset: Disjoint sets of training (Tr), validation (Va), and test (Te) samples were chosen, with their geographical locations (Excluding background samples) illustrated in Figure \ref{Fig5}, to train various SOTA models.}
    \resizebox{\columnwidth}{!}{\begin{tabular}{c|c|c|c|c|c|c|c} \hline 
        \textbf{Class} & \textbf{Tr} & \textbf{Va} & \textbf{Te} & \textbf{Class} & \textbf{Tr} & \textbf{Va} & \textbf{Te}\\ \hline
        Weeds 1 & 301 & 301 & 1407 & Soil vinyard develop & 930 & 930 & 4343 \\ 
        Weeds 2 & 558 & 559 & 2609 & Corn Weeds & 491 & 492 & 2295 \\ 
        Fallow & 296 & 296 & 1384 & Lettuce 4wk & 160 & 160 & 748 \\ 
        Fallow rough plow & 209 & 209 & 976 & Lettuce 5wk & 289 & 289 & 1349 \\
        Fallow smooth & 401 & 402 & 1875 & Lettuce 6wk & 137 & 137 & 642 \\
        Stubble & 593 & 594 & 2772 & Lettuce 7wk & 160 & 161 & 749 \\
        Celery & 536 & 537 & 2506 & Vinyard untrained & 1090 & 1090 & 5088 \\
        Grapes untrained & 1690 & 1691 & 7890 & Vinyard trellis & 271 & 271 & 1265 \\ \hline 
        \end{tabular}}
    \label{Tab4}
\end{table}
%%%%%%%%%%%%%%%%%%%%%%%%%%

\begin{figure}[!hbt]
	\centering
	\includegraphics[width=0.40\textwidth]{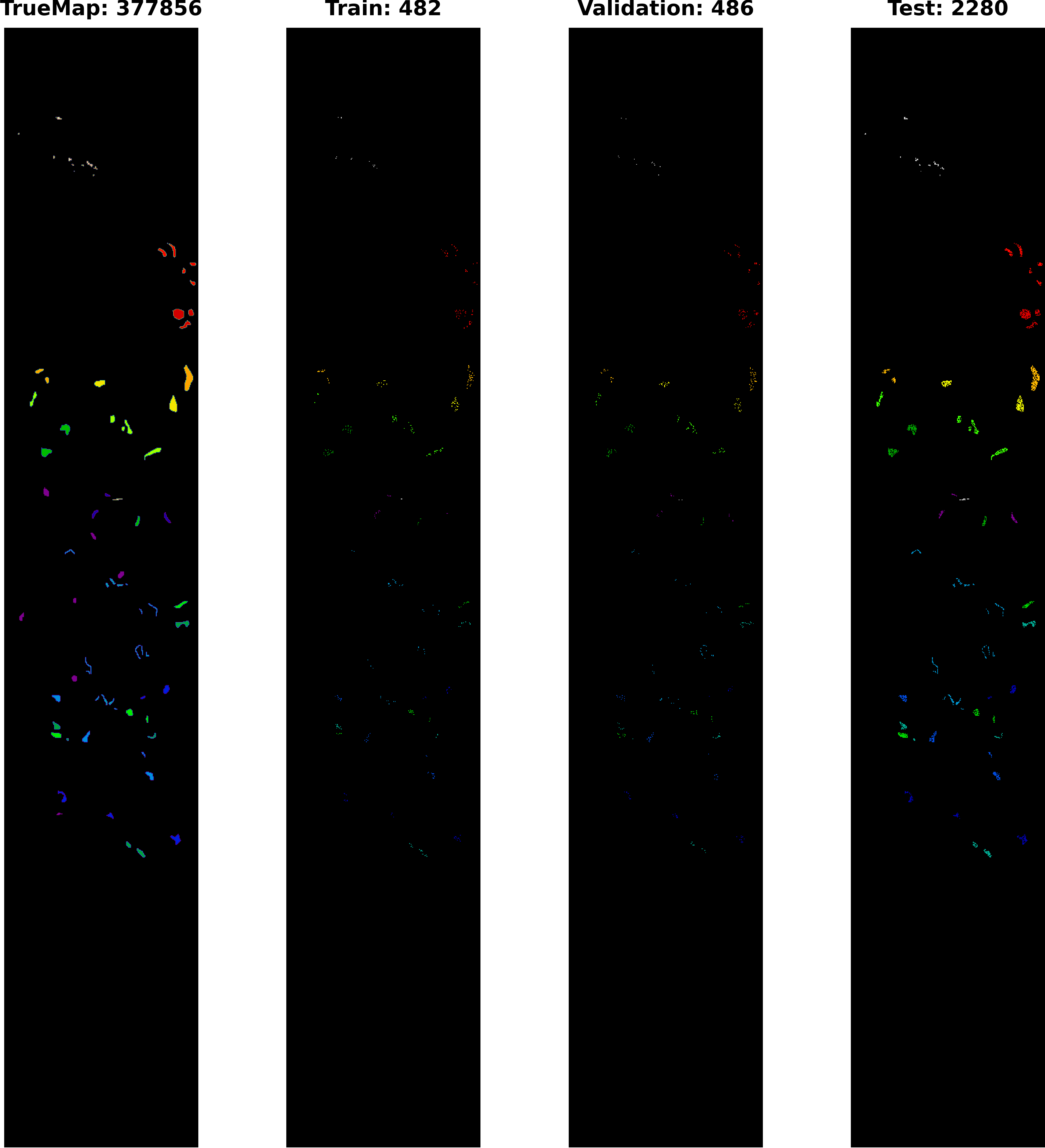}
	\caption{Botswana Dataset: Geographical locations of the disjoint train, validation, and test samples presented in Table \ref{Tab5}.}
	\label{Fig6}
\end{figure}

%%%%%%%%%%%%%%%%%%%%%%%%%%
\textbf{Botswana:} The NASA EO-1 satellite acquired hyperspectral imagery of the Okavango Delta region in Botswana from 2001-2004 using the Hyperion sensor to collect 30m resolution data across 242 bands from 400-2500nm over a 7.7km strip. The data analyzed from May 31, 2001, consisted of observations of 14 land cover classes representing seasonal swamps, occasional swamps, and drier woodlands in the distal delta region after preprocessing removed uncalibrated and noisy bands covering water absorption and retaining 145 bands. The disjoint train, validation, and test samples are presented in Table \ref{Tab5} and Figure \ref{Fig6}.

%%%%%%%%%%%%%%%%%%%%%%%%%%
\begin{table}[!hbt]
	\centering
	\caption{Botswana Dataset: Disjoint sets of training (Tr), validation (Va), and test (Te) samples were chosen, with their geographical locations (Excluding background samples) illustrated in Figure \ref{Fig6}, to train various SOTA models.}
	\resizebox{\columnwidth}{!}{\begin{tabular}{c|c|c|c|c|c|c|c} \hline 
			\textbf{Class} & \textbf{Tr} & \textbf{Va} & \textbf{Te} & \textbf{Class} & \textbf{Tr} & \textbf{Va} & \textbf{Te}\\ \hline
			Water & 40 & 41 & 189 & Island Interior & 30 & 30 & 143 \\
			Hippo Grass & 15 & 15 & 71 & Woodlands & 47 & 47 & 220 \\
			Floodplain Grasses 1 & 37 & 38 & 176 & Acacia Shrublands & 37 & 37 & 174 \\
			Floodplain Grasses 2 & 32 & 32 & 151 & Acacia Grasslands & 45 & 46 & 214 \\
			Reeds 1 & 40 & 40 & 189 & Short Mopane & 27 & 27 & 127 \\
			Riparian & 40 & 40 & 189 & Mixed Mopane & 40 & 40 & 188 \\ 
			Firescar 2 & 38 & 39 & 182 & Exposed Soils & 14 & 14 & 67 \\ \hline 
	\end{tabular}}
	\label{Tab5}
\end{table}
%%%%%%%%%%%%%%%%%%%%%%%%%%

%%%%%%%%%%%%%%%%%%%%%%%%%%
\subsection{Experimental Settings}

\begin{figure}[!hbt]
    \centering
    \includegraphics[width=0.48\textwidth]{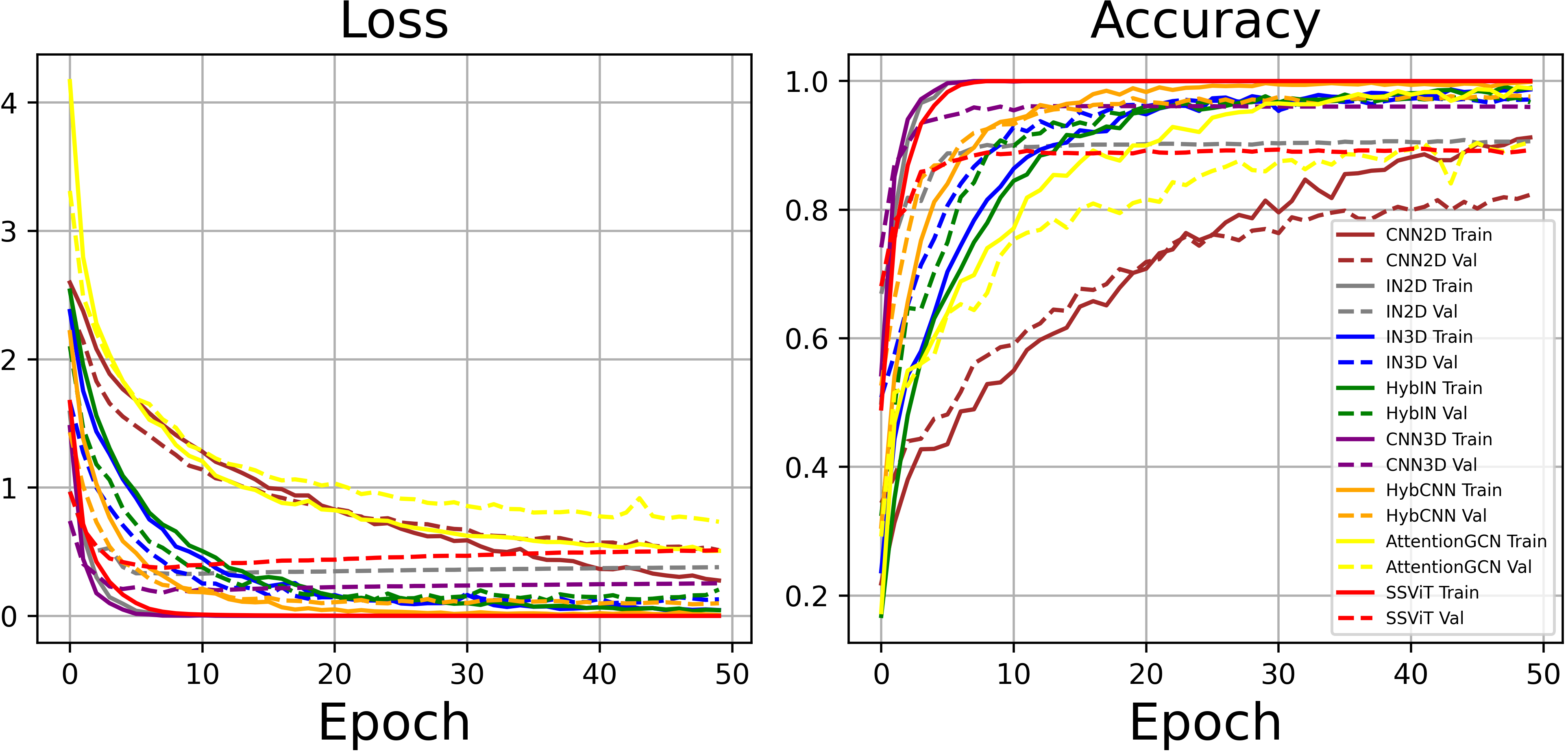}
    \caption{Loss and Accuracy trends for all the competing methods.}
    \label{fig:enter-label}
\end{figure}

In this section, we present comprehensive experimental settings for various deep learning models, including 3D CNN \cite{ahmad2020fast}, Hybrid Inception Net \cite{firat2023hybrid}, 3D Inception Net \cite{zhang2023improved}, 2D Inception Net \cite{xiong2018ai}, 2D CNN \cite{wu2022convolutional}, Hybrid CNN \cite{ghaderizadeh2021hyperspectral}, Attention Graph CNN \cite{10409250}, and Spatial-Spectral Transformer \cite{ahmad2024waveformer}. Prior to training, 3D overlapped patches are extracted using an $8 \times 8$ window size, as outlined in Algorithm 1. All models in this study are trained using the Adam optimizer with a learning rate of 0.0001, a decay rate of 1e-06, and a batch size of 56 for 50 epochs. The loss and accuracy trend is presented in Figure \ref{fig:enter-label} for all the competing methods.

%%%%%%%%%%%%%%%%%%%%%%%%%%
%%%%%%%%%%%%%%%%%%%%%%%%%%

The 2D CNN model is trained using four convolutional layers with kernel sizes of $(3 \times 3 \times 8, 16, 32, 64)$ and the same padding with $(8, 16, 32, 64)$ number of filters, respectively. Following the convolutional layers, two dense layers are utilized with a dropout rate of 0.4\%. Finally, a classification layer is added with the number of output units corresponding to the number of classes in the HSI dataset. The 3D CNN model is trained using four convolutional layers with kernel sizes of $(3 \times 3 \times 7, 5, 3, 3)$ with $(8, 16, 32, 64)$ number of filters, respectively. Following the convolutional layers, two dense layers are utilized and finally, a classification layer is added with the number of output units corresponding to the number of classes in the HSI dataset. The Hybrid CNN model is trained using three 3D convolutional layers with kernel sizes of $(3 \times 3 \times 7, 5, 3)$, followed by a reshaped layer to transform the features into 2D to learn spatial features using $3\times 3)$ kernel with 64 filters. Following the convolutional layers, two dense layers are utilized with a dropout rate of 0.4\%. Finally, a classification layer is added with the number of output units corresponding to the number of classes in the HSI dataset. 

The 2D Inception Net architecture consists of three blocks with the following configurations. In the first block, three 2D convolutional layers are used. The first layer employs a $(1 \times 1)$ kernel with 30 filters, the second layer uses a $(3 \times 3)$ kernel with 20 filters, and the third layer utilizes a $(1 \times 1)$ kernel with 10 filters. In the second block, three 2D convolutional layers are utilized. The first layer has a $(1 \times 1)$ kernel with 40 filters, the second layer employs a $(5 \times 5)$ kernel with 20 filters, and the third layer uses a $(1 \times 1)$ kernel with 10 filters. The third block begins with a 2D max pooling operation using a $(3 \times 3)$ kernel and the same padding. This is followed by two 2D convolutional layers with $(1 \times 1)$ kernels and the same padding. The filters for these layers are set to 20 and 10, respectively. Afterward, the outputs from all three blocks are concatenated, and a convolutional layer with a $(1 \times 1)$ kernel and 128 filters is applied. Following the convolutional layer, two dense layers are deployed. Finally, a classification layer is added with the number of output units corresponding to the number of classes in the HSI dataset.

The 3D Inception Net architecture consists of three blocks with the following configurations. In the first block, three 3D convolutional layers are used. The first layer employs a $(5 \times 5 \times 7)$ kernel with 30 filters, the second layer uses a $(3 \times 3 \times 5)$ kernel with 20 filters, and the third layer utilizes a $(3 \times 3 \times 3)$ kernel with 10 filters and the same padding in all three layers. In the second block, three 3D convolutional layers are utilized. The first layer has a $(5 \times 5 \times 7)$ kernel with 40 filters, the second layer employs a $(3 \times 3 \times 5)$ kernel with 20 filters, and the third layer uses a $(3 \times 3 \times 3)$ kernel with 10 filters and the same padding in all three layers. The third block begins with three 3D convolutional layers with $(5 \times 5 \times 7)$ kernel with 60 filters, the second layer uses a $(3 \times 3 \times 5)$ kernel with 30 filters, and the third layer utilizes a $(3 \times 3 \times 3)$ kernel with 10 filters and the same padding in all three layers. Afterward, the outputs from all three blocks are concatenated, and a convolutional layer with a $(1 \times 1 \times 1)$ kernel and 128 filters is applied. Following the convolutional layer, two dense layers are deployed with a 0.4\% dropout rate. Finally, a classification layer is added with the number of output units corresponding to the number of classes in the HSI dataset.

%%%%%%%%%%%%%%%%%%%%%%%%%%
\begin{figure*}[!hbt]
    \centering
	\begin{subfigure}{0.49\textwidth}
		\includegraphics[width=0.99\textwidth]{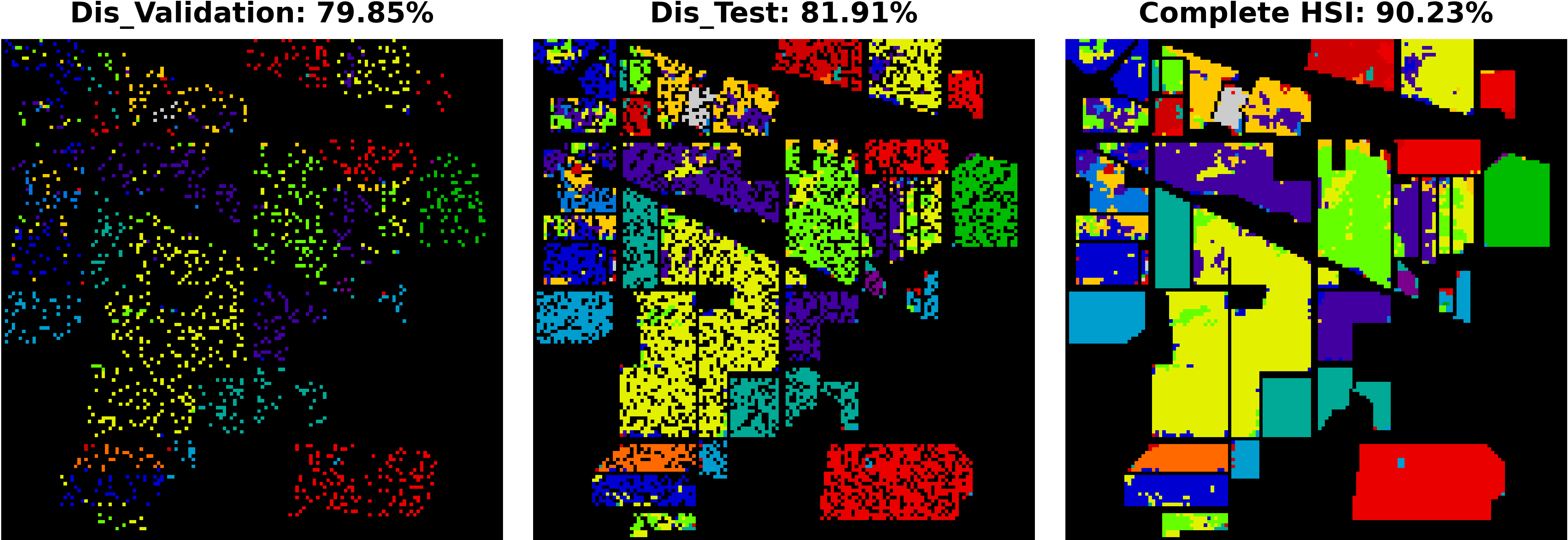}
		\caption{2D CNN} 
		\label{Fig7A}
	\end{subfigure}
	\begin{subfigure}{0.49\textwidth}
		\includegraphics[width=0.99\textwidth]{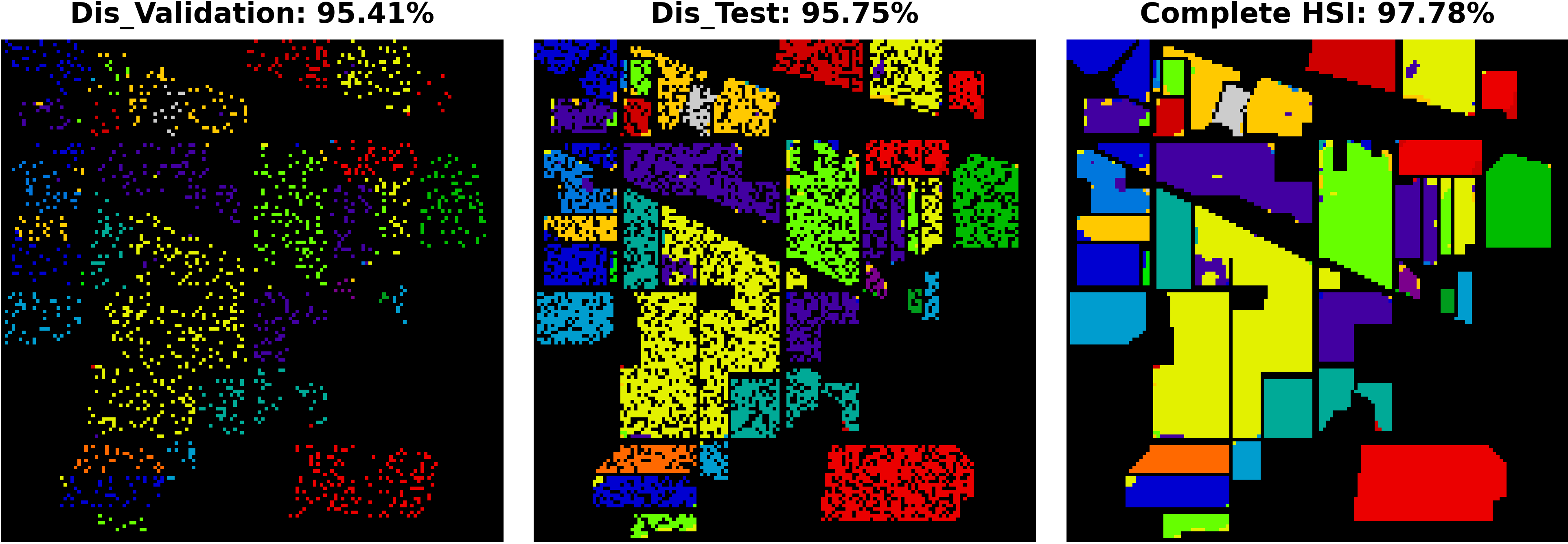}
		\caption{3D CNN}
		\label{Fig7B}
	\end{subfigure} 
	\begin{subfigure}{0.49\textwidth}
		\includegraphics[width=0.99\textwidth]{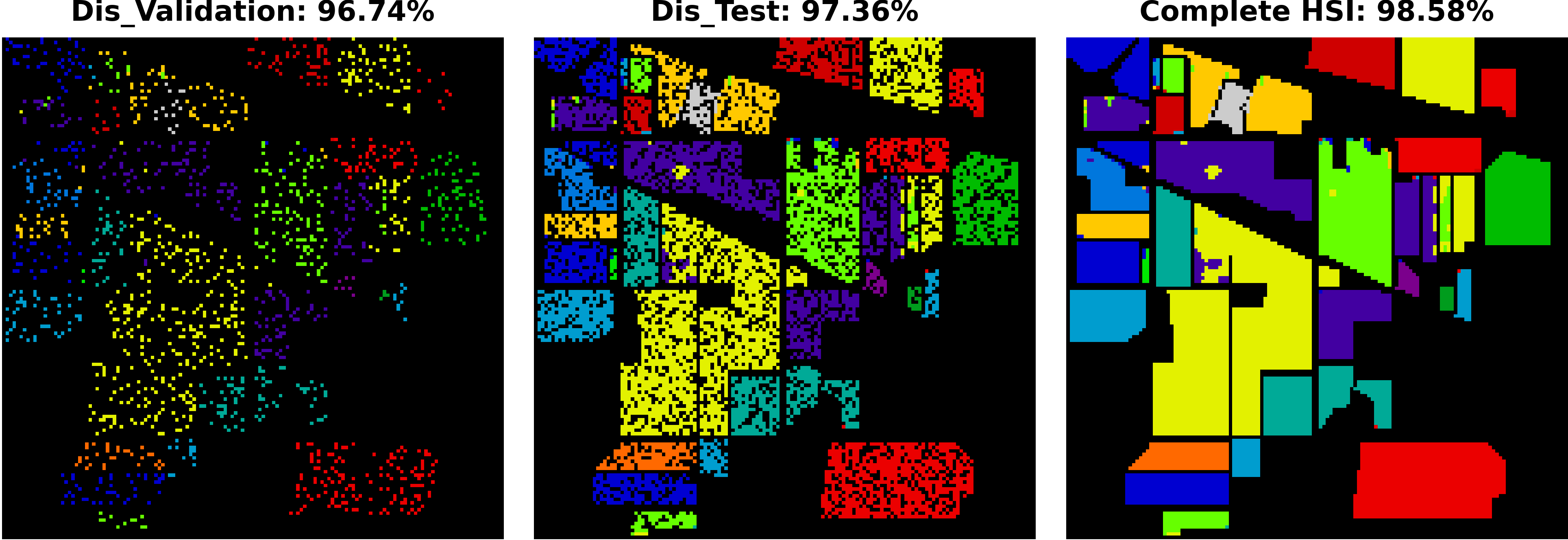}
		\caption{Hybrid CNN}
		\label{Fig7C}
	\end{subfigure} 
	\begin{subfigure}{0.49\textwidth}
		\includegraphics[width=0.99\textwidth]{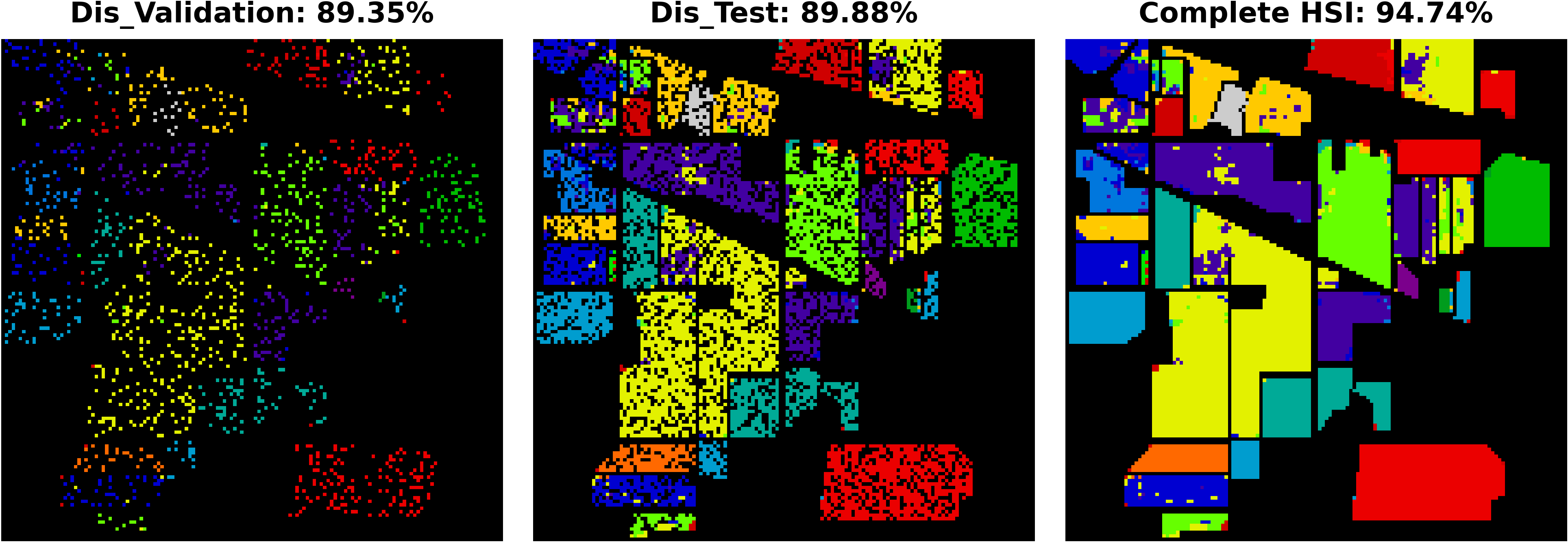}
		\caption{2D IN}
		\label{Fig7D}
	\end{subfigure} 
	\begin{subfigure}{0.49\textwidth}
		\includegraphics[width=0.99\textwidth]{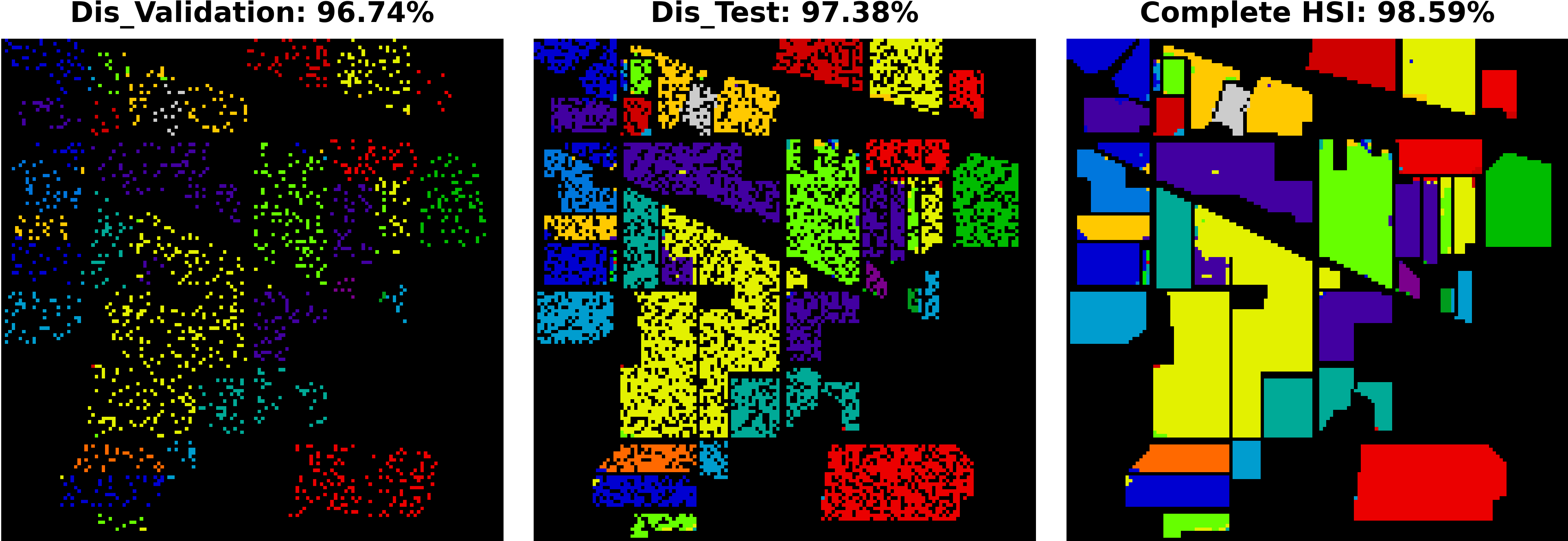}
		\caption{3D IN}
		\label{Fig7E}
	\end{subfigure} 
	\begin{subfigure}{0.49\textwidth}
		\includegraphics[width=0.99\textwidth]{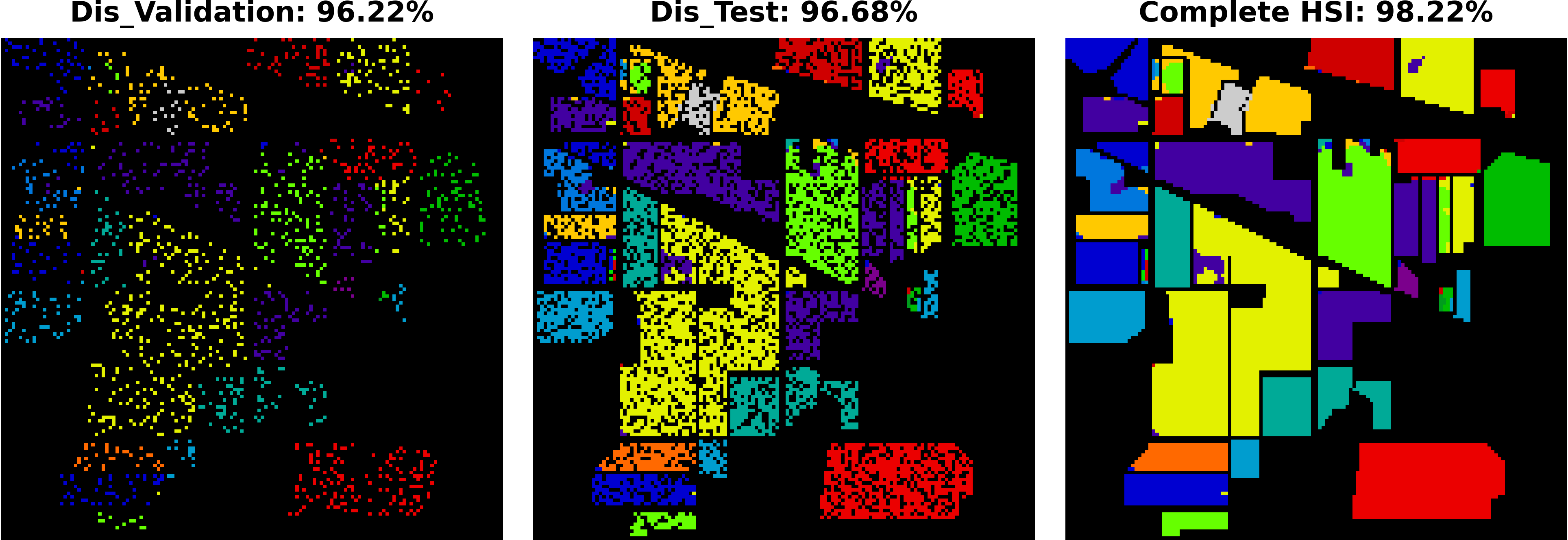}
		\caption{Hybrid IN}
		\label{Fig7F}
	\end{subfigure} 
	\begin{subfigure}{0.49\textwidth}
		\includegraphics[width=0.99\textwidth]{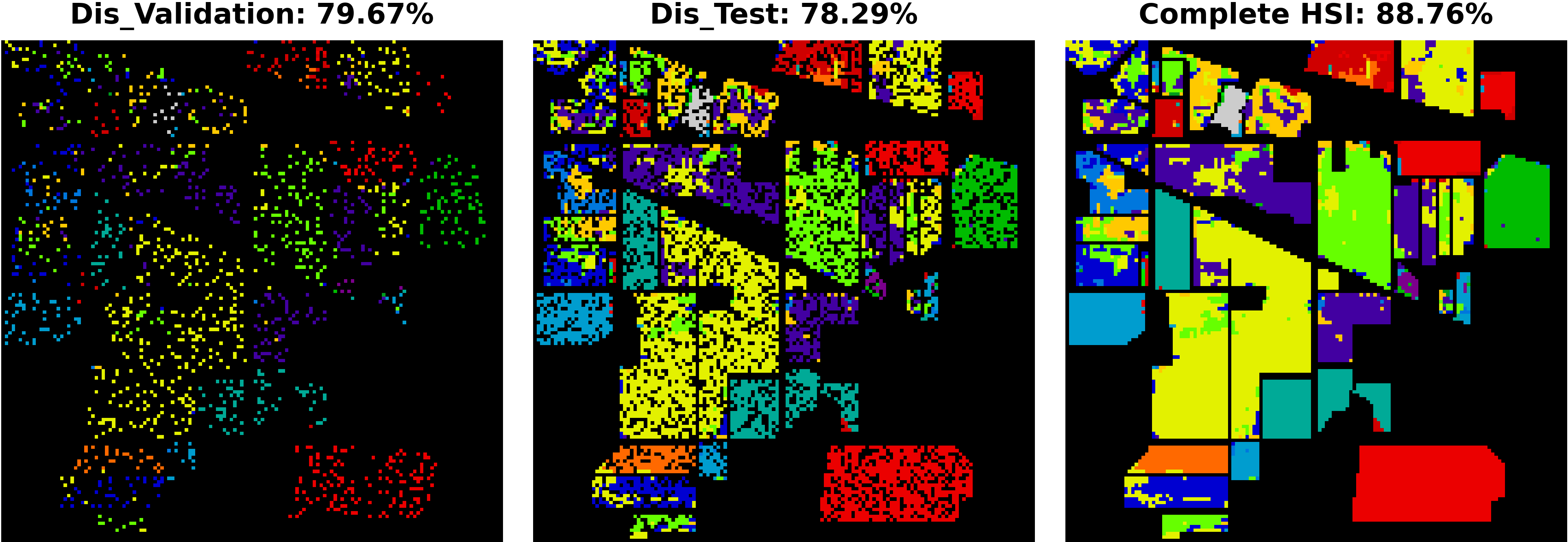}
		\caption{GCN}
		\label{Fig7G}
	\end{subfigure} 
	\begin{subfigure}{0.49\textwidth}
		\includegraphics[width=0.99\textwidth]{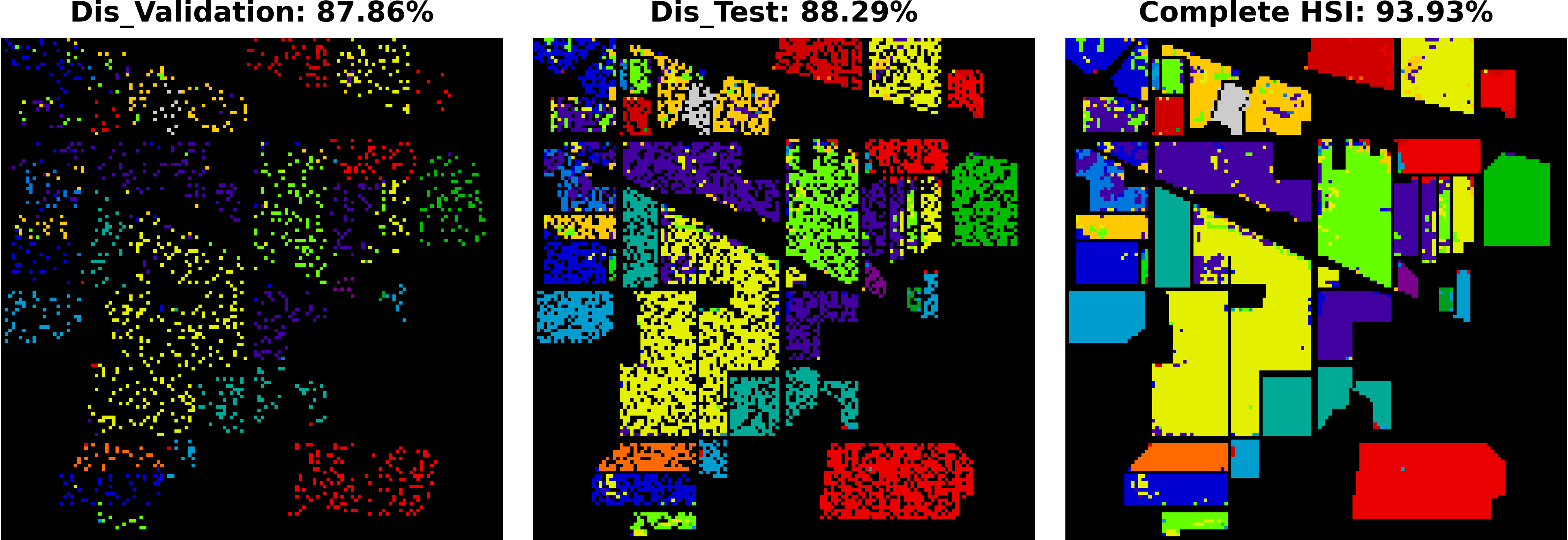}
		\caption{Transformer}
		\label{Fig7H}
	\end{subfigure} 
\caption{\textbf{Indian Pines Dataset:} Land cover maps for disjoint validation, test, and the entire HSI used as a test set are provided. Comprehensive class-wise results can be found in Table \ref{Tab6}.}
\label{Fig7}
\end{figure*}
%%%%%%%%%%%%%%%%%%%%%%%%%%
\begin{table*}[!hbt]
    \centering
    \caption{Indian Pines Dataset: Per class comparative results of various SOTA models are showcased on disjoint validation and test sets. Additionally, results on the entire HSI dataset serving as the test set are also presented. The comparative methods include 3D CNN \cite{ahmad2020fast}, Hybrid Inception Net (Hybrid IN) \cite{firat2023hybrid}, 3D Inception Net (3D IN) \cite{zhang2023improved}, 2D Inception Net (2D IN) \cite{xiong2018ai}, 2D CNN \cite{wu2022convolutional}, Hybrid CNN \cite{ghaderizadeh2021hyperspectral}, Attention Graph CNN (Attention GCN) \cite{10409250}, and Spatial-Spectral Transformer \cite{ahmad2024waveformer}. The geographical maps for each model for disjoint validation, test, and complete test are presented in Figure \ref{Fig7}.}
    
    \resizebox{\textwidth}{!}{\begin{tabular}{c|ccc|ccc|ccc|ccc|ccc|ccc|ccc|ccc} \hline 
    
    \multirow{2}{*}{\textbf{Class}} & \multicolumn{3}{c|}{\textbf{2D CNN}} & \multicolumn{3}{c|}{\textbf{3D CNN}} & \multicolumn{3}{c|}{\textbf{Hybrid CNN}} & \multicolumn{3}{c|}{\textbf{2D IN}} & \multicolumn{3}{c|}{\textbf{3D IN}} & \multicolumn{3}{c|}{\textbf{Hybrid IN}} &  \multicolumn{3}{c|}{\textbf{Attention GCN}} & \multicolumn{3}{c}{\textbf{SSViT}} \\ \cline{2-25}

    & Va & Te & HSI & Va & Te & HSI & Va & Te & HSI & Va & Te & HSI & Va & Te & HSI & Va & Te & HSI & Va & Te & HSI & Va & Te & HSI \\ \hline

Alfalfa & 57.14 & 69.69 & 99.87 & 85.71 & 75.75 & 99.92 & 100.00 & 96.97 & 99.99 & 100.00 & 93.94 & 99.98 & 100.00 & 90.91 & 99.97 & 100.00 & 93.94 & 99.98 & 71.43 & 60.61 & 99.86 & 100.00 & 90.91 & 99.97\\

Corn-notill & 74.76 & 75.4 & 78.36 & 92.52 & 94.10 & 94.75 & 92.53 & 92.80 & 93.77 & 84.58 & 84.20 & 86.62 & 97.66 & 97.30 & 97.76 & 95.33 & 96.00 & 96.50 & 76.64 & 71.30 & 75.70 & 85.51 & 84.70 & 87.11\\

Corn-mintill & 76.8 & 81.06 & 82.77 & 96.00 & 97.76 & 97.83 & 97.60 & 99.83 & 99.52 & 84.80 & 86.40 & 88.19 & 96.80 & 98.45 & 98.43 & 97.60 & 99.66 & 99.40 & 60.80 & 61.96 & 66.14 & 82.4 & 84.34 & 86.39\\ 

Corn & 58.33 & 45.78 & 54.85 & 94.44 & 89.16 & 91.56 & 94.44  & 95.18 & 95.78 & 88.89 & 79.52 & 83.97 & 100.00 & 98.80 & 99.16 & 88.89 & 88.55 & 90.30 & 63.89 & 53.61 & 62.03 & 58.33 & 48.80 & 57.81\\
Grass-pasture & 94.44 & 94.10 & 95.03 & 97.22 & 96.76 & 97.31 & 98.61 & 97.94 & 98.34 & 95.83 & 95.28 & 96.07 & 98.61 & 97.05 & 97.72 & 95.83 & 96.76 & 97.10 & 90.28 & 89.97 & 91.30 & 94.44 & 96.17 & 96.48\\
Grass-trees & 100. & 99.41 & 99.58 & 99.09 & 99.41 & 99.45 & 100.00 & 99.80 & 99.86 & 98.18 & 99.41 & 99.32 & 100.00 & 99.80 & 99.86 & 100.00 & 100.00 & 100.00 & 98.18 & 98.63 & 98.77 & 99.09 & 99.22 & 99.32\\   
Grass-mowed & 0. & 0. & 7.14 & 100.00 & 100.00 & 100.00 & 100.00 & 100.00 & 100.00 & 75.00 & 70.00 & 75.00 & 75.00 & 70.00 & 75.00 & 25.00 & 20.00 & 25.00 & 25.00 & 5.00 & 14.29 & 75.00 & 75.00 & 78.57\\   
Hay-windrowed & 98.611 & 98.50 & 98.74 & 100 & 99.70 & 99.79 & 100.00 & 100.00 & 100.00 & 100.00 & 98.21 & 98.74 & 100.00 & 100.00 & 100.00 & 100.00 & 100.00 & 100.00 & 98.61 & 95.52 & 96.65 & 98.61 & 99.40 & 99.37\\
Oats & 0. & 0. & 5.0 & 66.66 & 71.43 & 75.00 & 66.67 & 85.71 & 85.00 & 33.33 & 71.43 & 70.00 & 0.00 & 28.57 & 35.00 & 0.00 & 28.57 & 30.00 & 0.00 & 21.43 & 30.00 & 0.00 & 78.57 & 70.00\\
Soybean-notill & 75.34 & 74.44 & 77.77 & 92.47 & 91.34 & 92.80 & 91.78 & 93.10 & 93.93 & 86.30 & 81.64 & 85.08 & 91.78 & 94.42 & 94.86 & 88.36 & 91.48 & 92.18 & 81.51 & 79.00 & 82.41 & 82.19 & 82.97 & 85.39\\
Soybean-mintill & 85.05 & 88.24 & 89.16 & 97.83 & 97.27 & 97.76 & 99.73 & 99.71 & 99.76 & 89.40 & 92.15 & 92.91 & 98.10 & 98.49 & 98.66 & 98.91 & 98.72 & 98.94 & 86.14 & 87.73 & 89.33 & 91.58 & 92.03 & 93.16\\
Soybean-clean & 58.42 & 63.22 & 67.11 & 92.13 & 96.88 & 96.63 & 93.26 & 96.88 & 96.80 & 91.01 & 90.87 & 92.24 & 89.89 & 97.12 & 96.46 & 97.75 & 96.88 & 97.47 & 50.56 & 49.76 & 56.16 & 79.78 & 80.53 & 83.31\\
Wheat & 83.87 & 95.13 & 94.14 & 100.00 & 100.00 & 100.00 & 100.00 & 100.00 & 100.00 & 96.77 & 99.31 & 99.02 & 100.00 & 97.92 & 98.54 & 100.00 & 98.61 & 99.02 & 96.77 & 92.36 & 94.15 & 93.55 & 99.31 & 98.54\\
Woods & 97.36 & 97.40 & 97.78 & 96.32 & 97.29 & 97.55 & 98.95 & 99.32 & 99.37 & 98.42 & 99.10 & 99.13 & 98.95 & 98.53 & 98.81 & 98.95 & 99.21 & 99.30 & 97.37 & 97.18 & 97.63 & 97.89 & 96.73 & 97.39\\
Buildings & 75.86 & 81.91 & 83.67 & 100 & 97.79 & 98.45 & 100.00 & 98.89 & 99.22 & 86.21 & 92.25 & 92.49 & 100.00 & 98.15 & 98.70 & 100.00 & 97.42 &98.19 & 77.58 & 75.65 & 79.53 & 93.10 & 94.83 & 95.34\\
Stone-Steel & 50. & 83.33 & 79.56 & 92,86 & 96.97 & 96.77 & 100.00 & 100.00 & 100.00 & 92.86 & 98.48 & 97.85 & 100.00 & 100.00 & 100.00 & 100.00 & 100.00 & 100.00 & 78.57 & 83.33 & 84.95 & 92.86 & 100.00 & 98.92\\ \hline 

\textbf{Kappa} & 79.85 & 81.91 & 90.23 & 95.41 & 95.75 & 97.78 & 96.74 & 97.36 & 98.58 & 89.35 & 89.88 & 94.74 & 96.74 & 97.38 & 98.59 & 96.22 & 96.68 & 98.22 & 79.67 & 78.29 & 88.76 & 87.86 & 88.29 & 93.93\\
\textbf{OA} & 82.33 & 84.17 & 93.10 & 95.97 & 96.27 & 98.43 & 97.14 & 97.69 & 99.00 & 90.64 & 91.13 & 96.29 & 97.14 & 97.70 & 99.00 & 96.69 & 97.09 &  98.74 & 82.20 & 81.06 & 92.08 & 89.34 & 89.74 & 95.71\\
\textbf{AA} & 67.88 & 71.73 & 75.66 &  93.95 & 93.85 & 95.97 & 95.85 & 97.26 & 97.58 & 87.60 & 89.51 & 91.04 & 90.42 & 91.59 & 93.06 & 87.86 & 87.86 & 88.96 & 72.08 & 70.19 & 76.18 & 87.72 & 87.72 & 89.19 \\
\textbf{Time (S)} & 1.54 & 1.68 & 8.74 &  0.45 & 1.39 & 9.80 & 0.53 & 0.81 & 8.64 & 0.85 & 1.39 & 10.96 & 1.48 & 3.14 & 14.84 & 0.89 & 2.68 & 12.75 & 1.18 & 1.38 & 10.27 & 1.28 & 2.24 & 14.72\\ \hline 
    \end{tabular}
    \label{Tab6}}
\end{table*}
%%%%%%%%%%%%%%%%%%%%%%%%%%

%%%%%%%%%%%%%%%%%%%%%%%%%%
\begin{figure*}[!hbt]
    \centering
	\begin{subfigure}{0.32\textwidth}
		\includegraphics[width=0.99\textwidth]{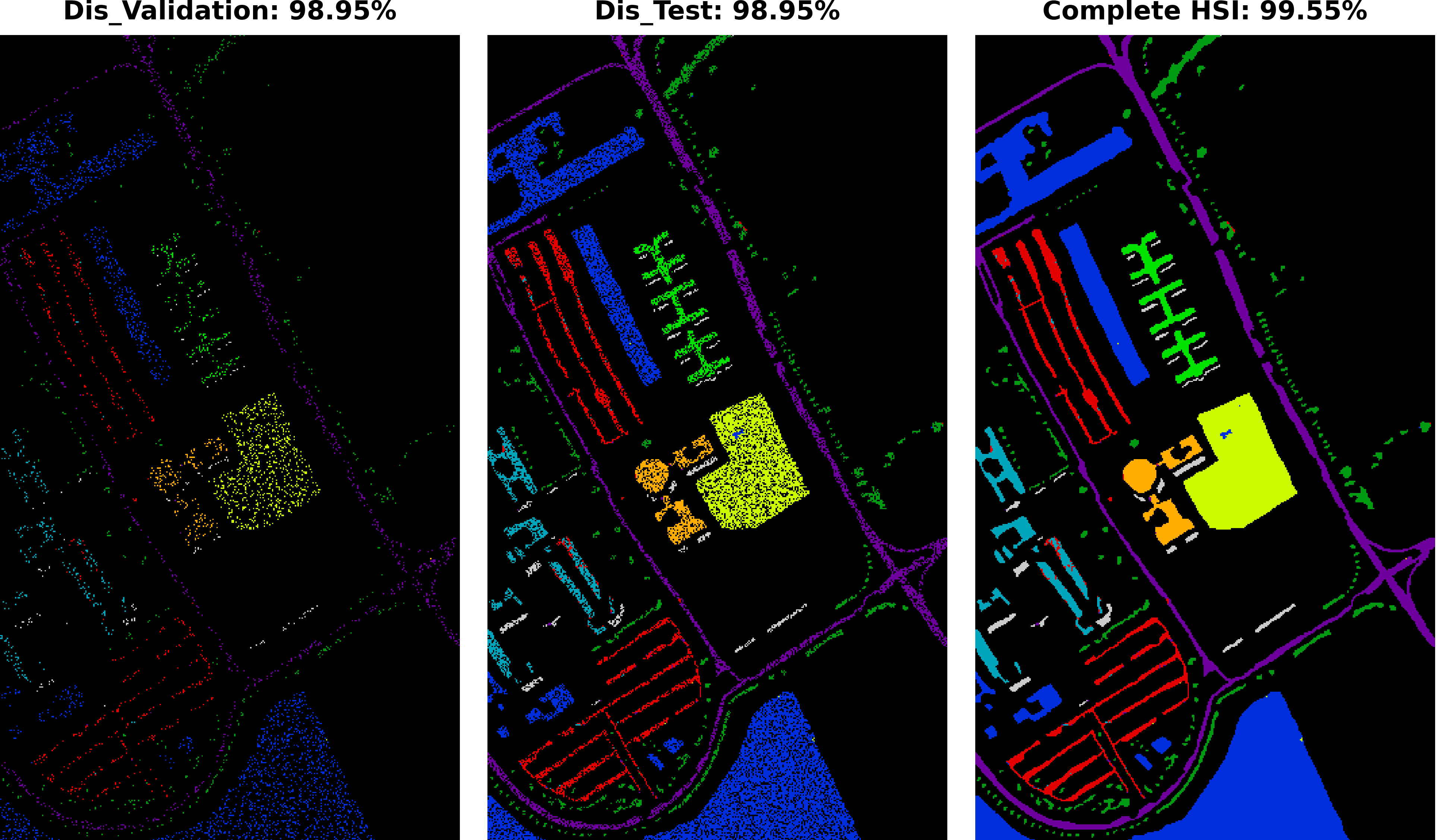}
		\caption{2D CNN} 
		\label{Fig8A}
	\end{subfigure}
	\begin{subfigure}{0.32\textwidth}
		\includegraphics[width=0.99\textwidth]{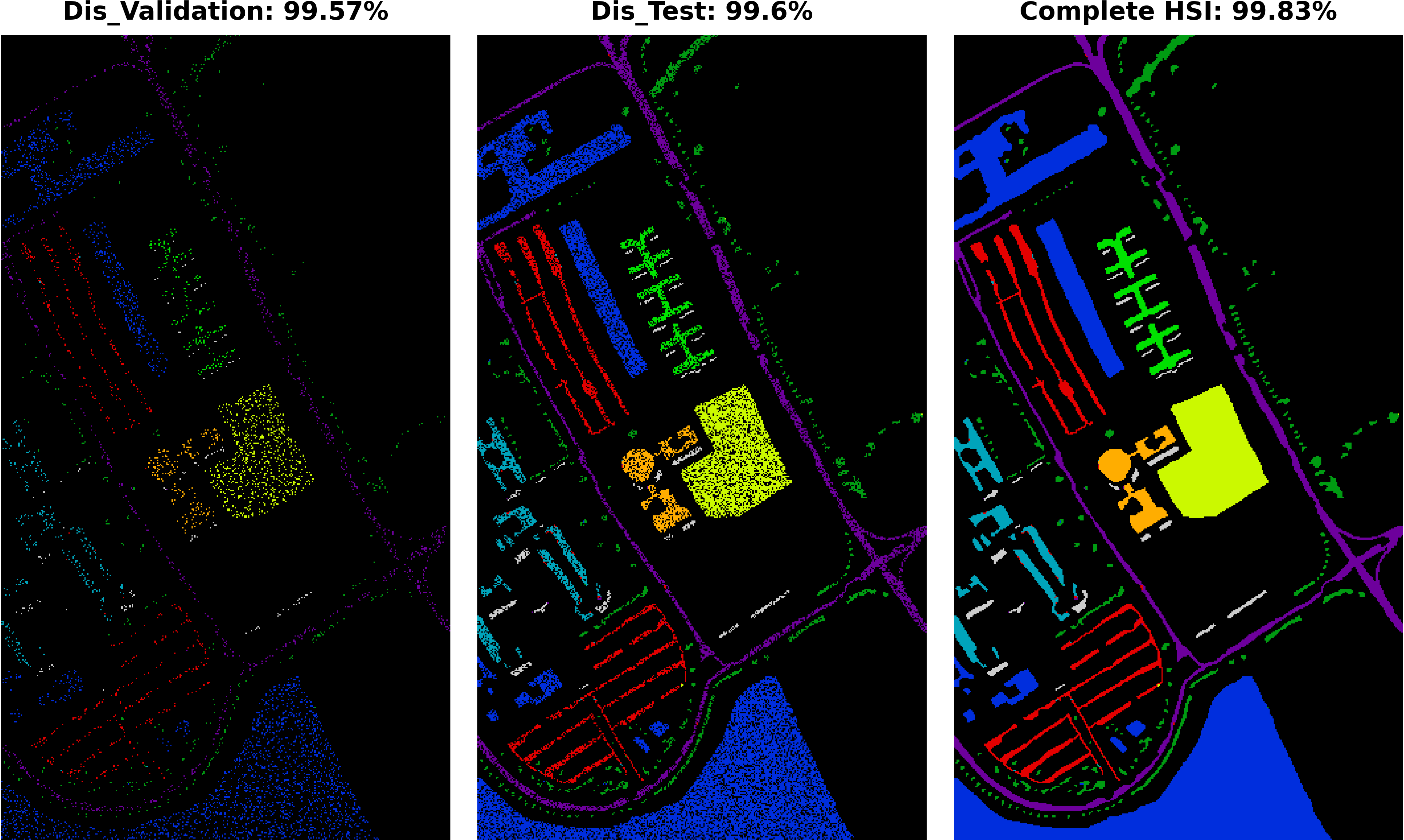}
		\caption{3D CNN}
		\label{Fig8B}
	\end{subfigure} 
	\begin{subfigure}{0.32\textwidth}
		\includegraphics[width=0.99\textwidth]{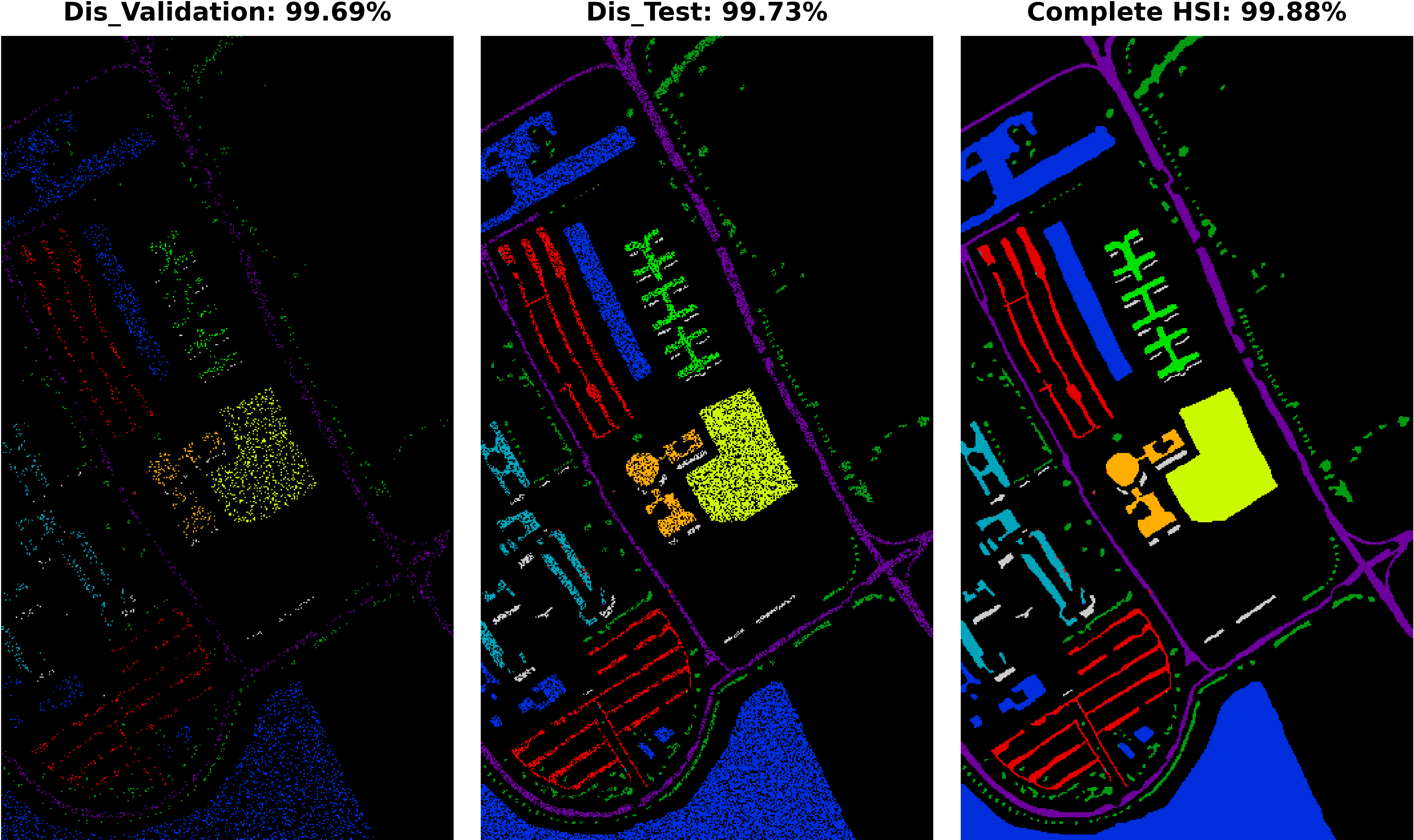}
		\caption{Hybrid CNN}
		\label{Fig8C}
	\end{subfigure} 
	\begin{subfigure}{0.32\textwidth}
		\includegraphics[width=0.99\textwidth]{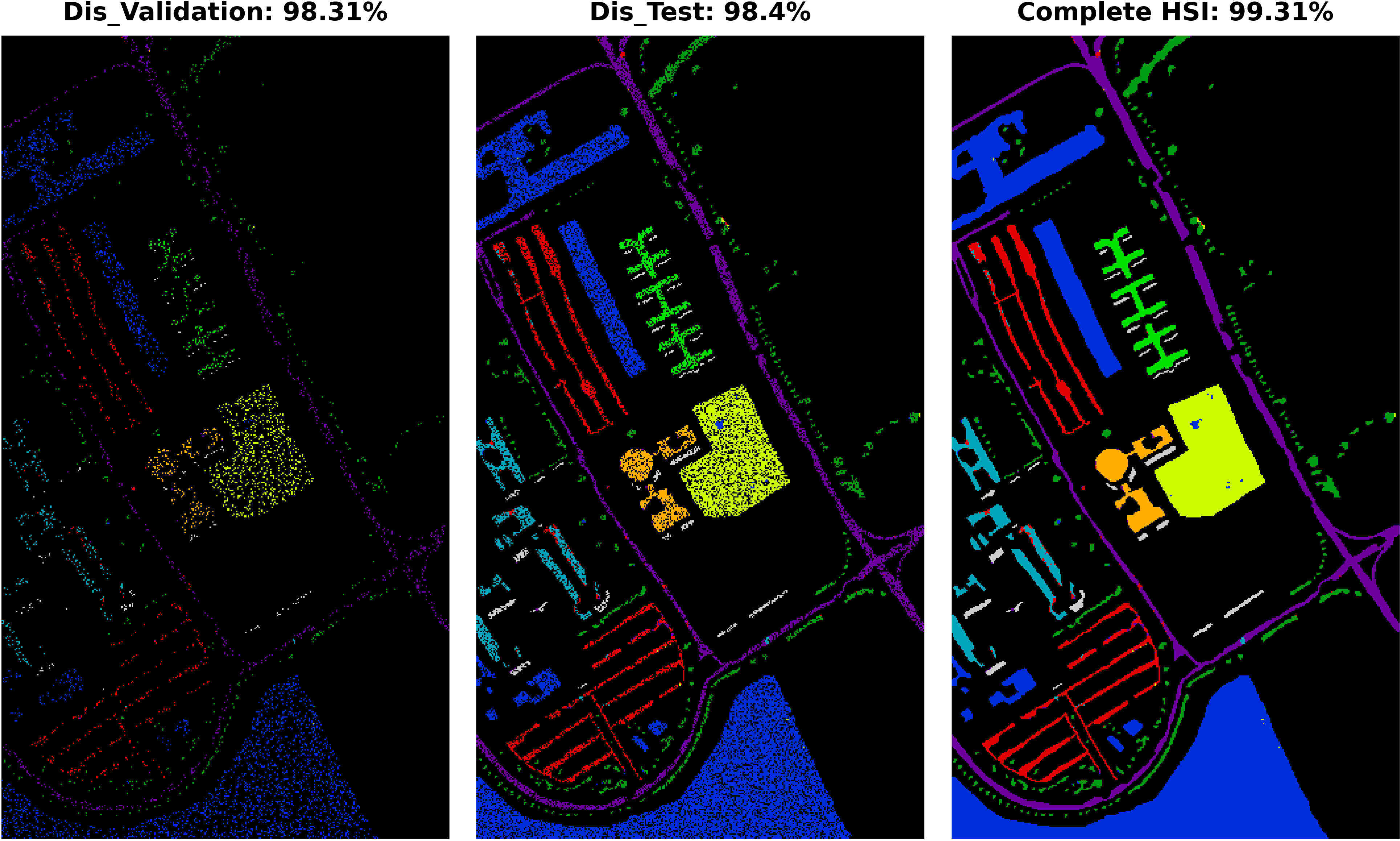}
		\caption{2D IN}
		\label{Fig8D}
	\end{subfigure} 
	\begin{subfigure}{0.32\textwidth}
		\includegraphics[width=0.99\textwidth]{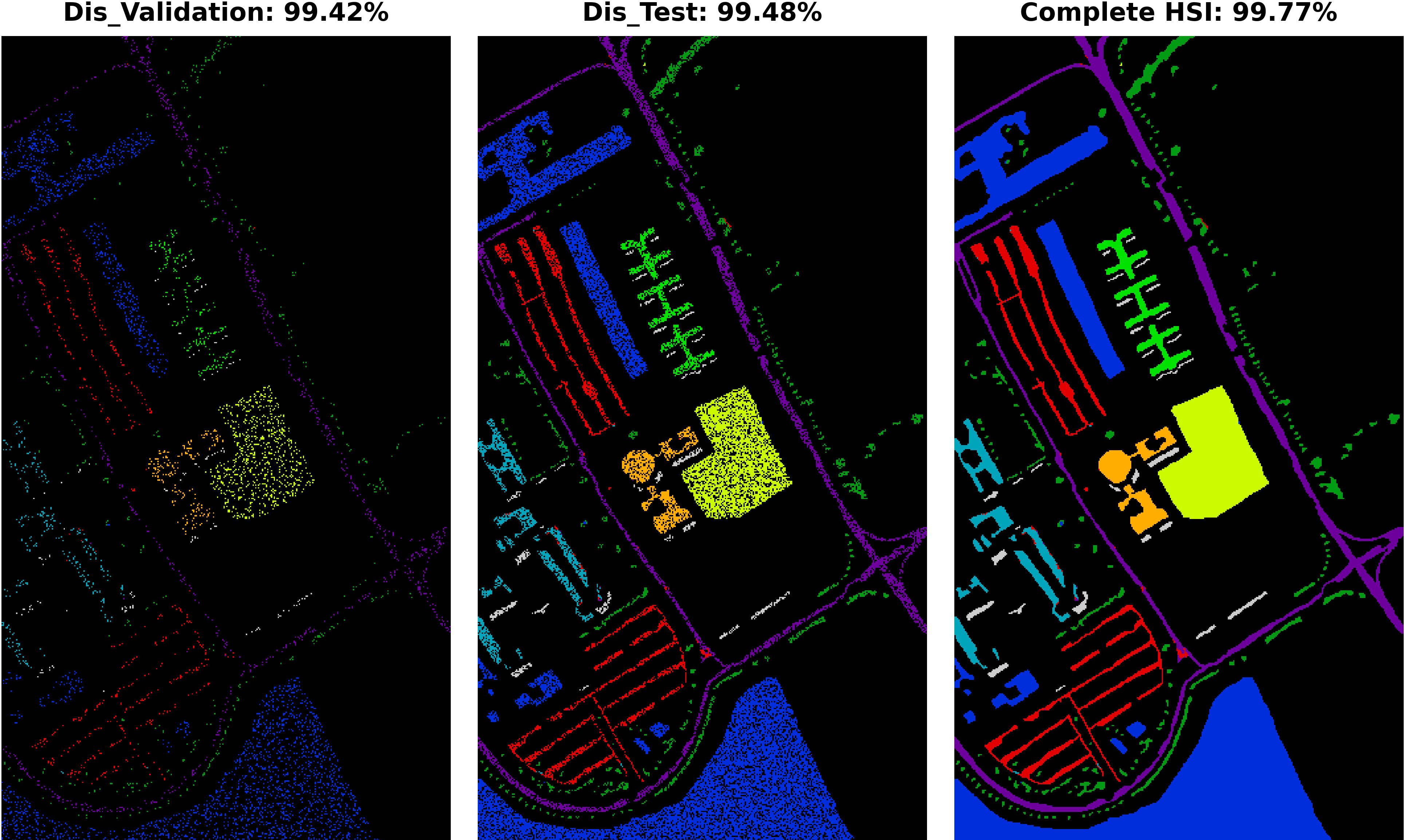}
		\caption{3D IN}
		\label{Fig8E}
	\end{subfigure} 
	\begin{subfigure}{0.32\textwidth}
		\includegraphics[width=0.99\textwidth]{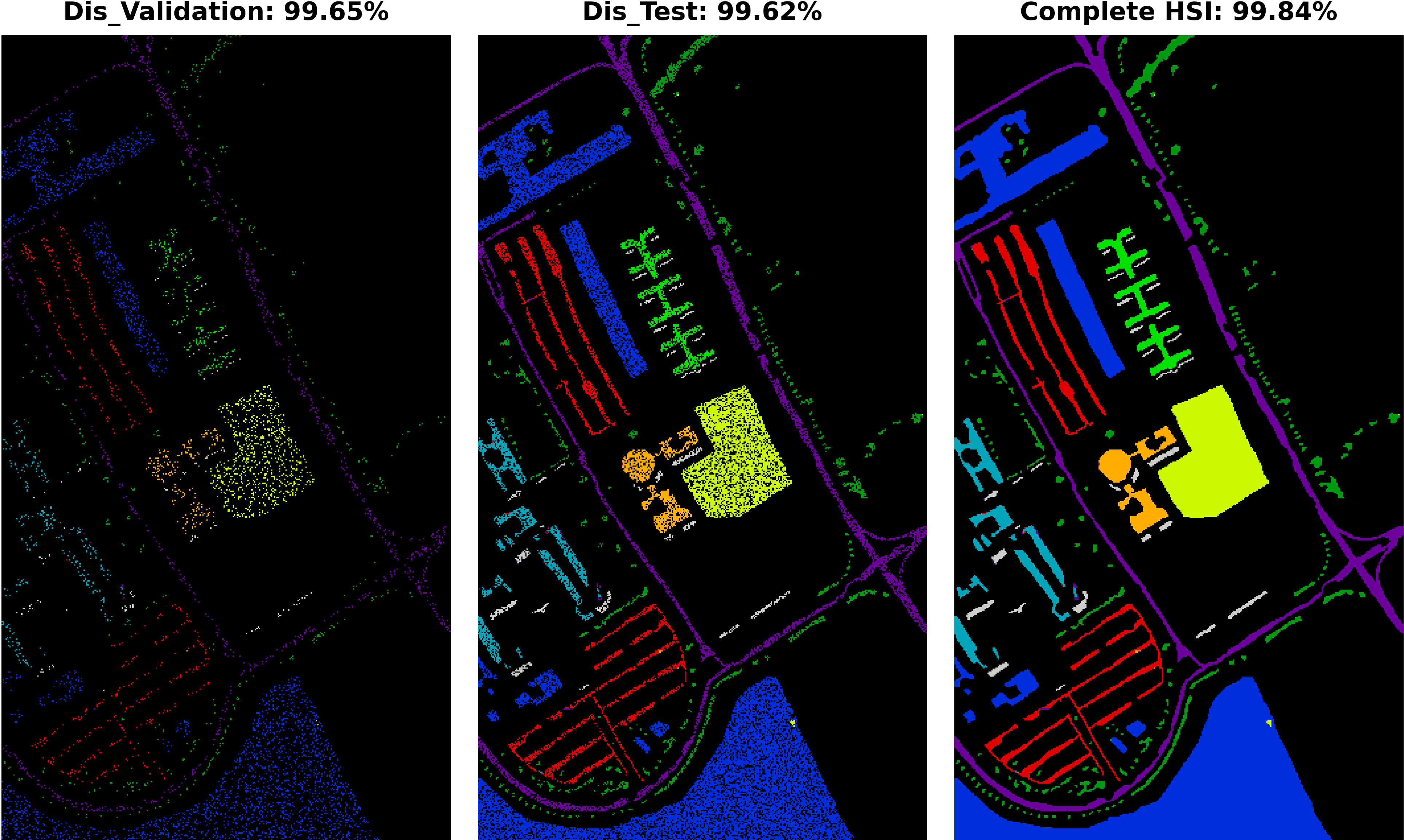}
		\caption{Hybrid IN}
		\label{Fig8F}
	\end{subfigure} 
	\begin{subfigure}{0.32\textwidth}
		\includegraphics[width=0.99\textwidth]{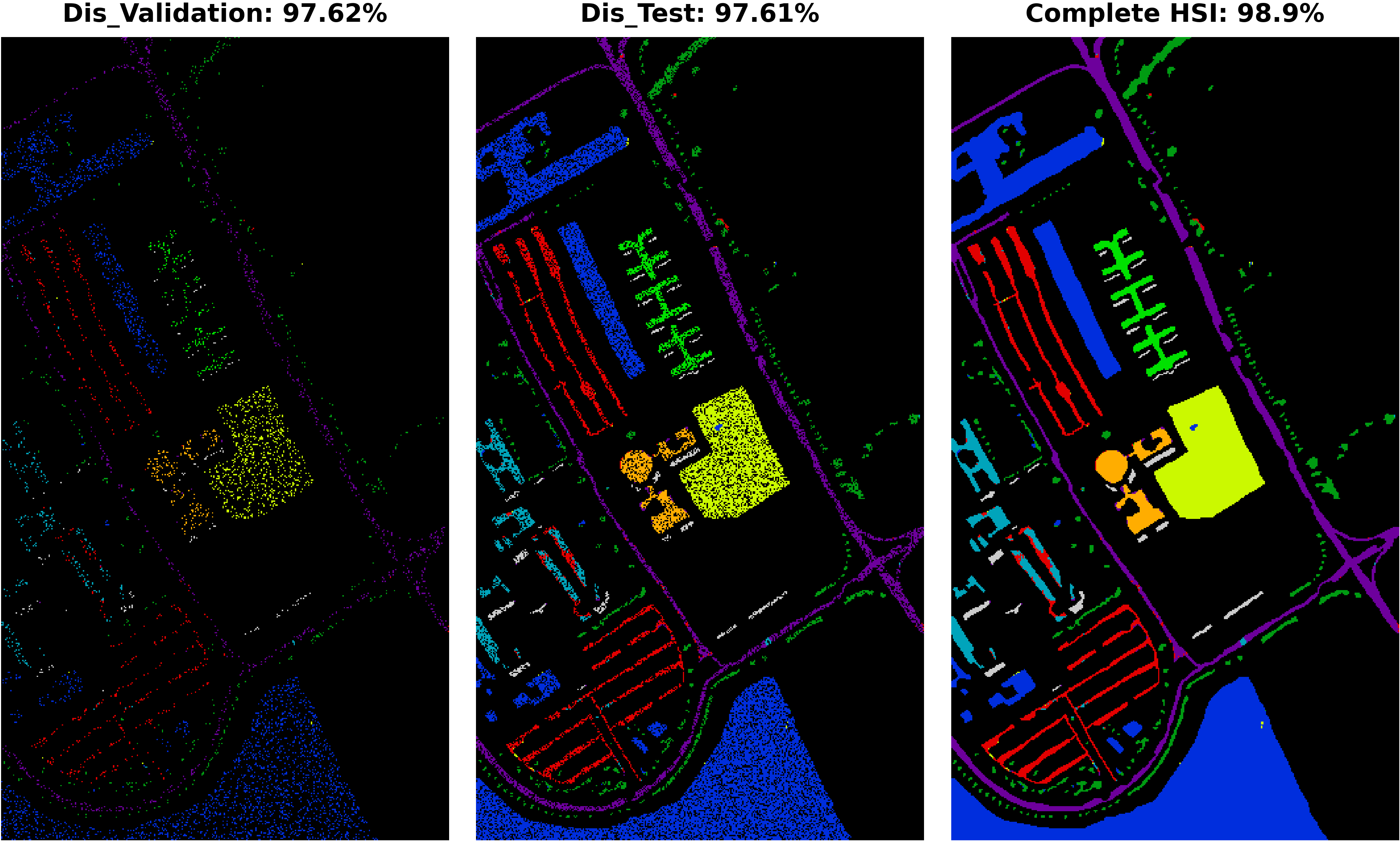}
		\caption{GCN}
		\label{Fig8G}
	\end{subfigure} 
	\begin{subfigure}{0.32\textwidth}
		\includegraphics[width=0.99\textwidth]{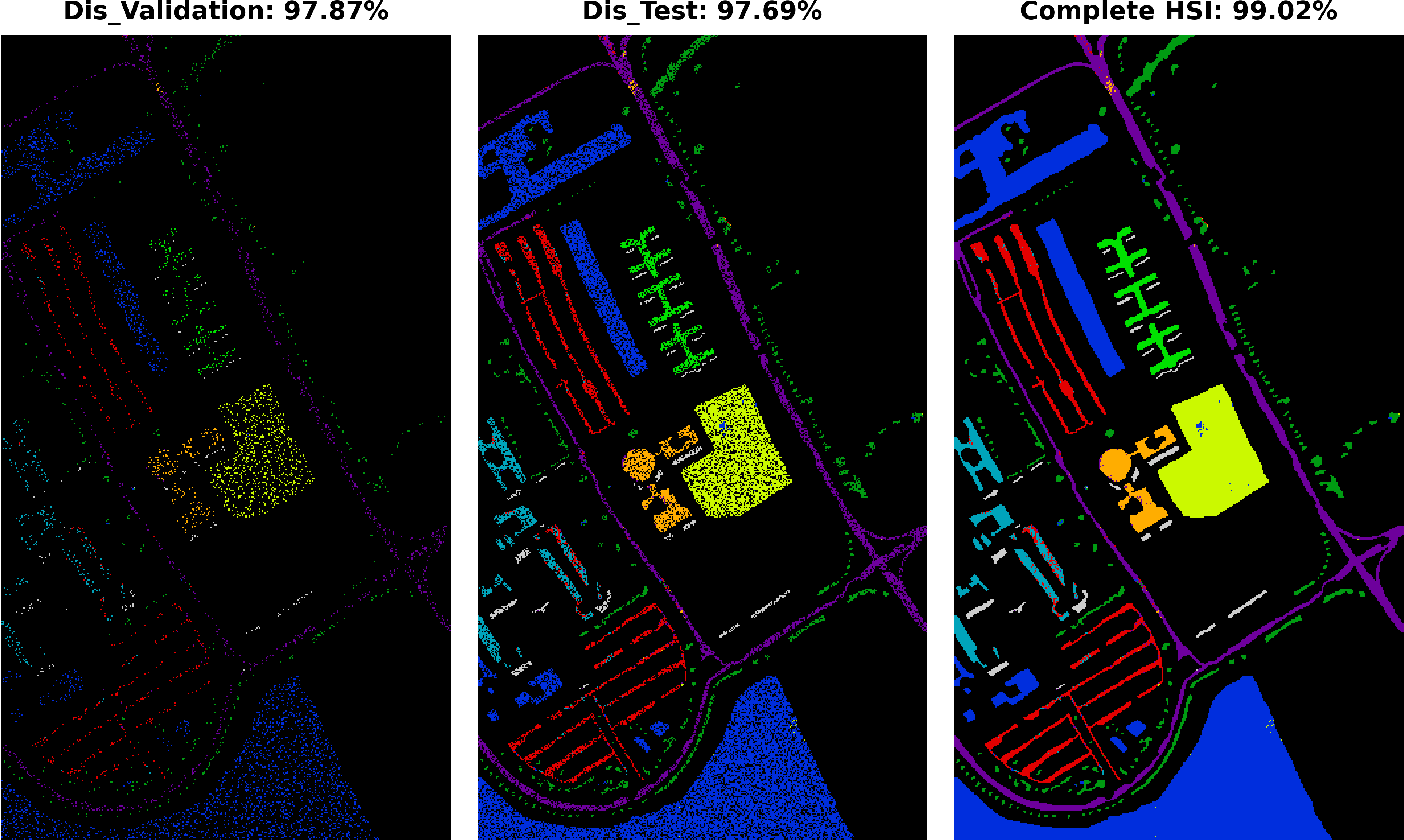}
		\caption{Transformer}
		\label{Fig8H}
	\end{subfigure} 
\caption{\textbf{Pavia University Dataset:} Land cover maps for disjoint validation, test, and the entire HSI used as a test set are provided. Comprehensive class-wise results can be found in Table \ref{Tab7}.}
\label{Fig8}
\end{figure*}
%%%%%%%%%%%%%%%%%%%%%%%%%%
\begin{table*}[!t]
    \centering
    \caption{Pavia University Dataset: Per class comparative results of various SOTA models are showcased on disjoint validation and test sets. Additionally, results on the entire HSI dataset serving as the test set are also presented. The comparative methods include 3D CNN \cite{ahmad2020fast}, Hybrid Inception Net (Hybrid IN) \cite{firat2023hybrid}, 3D Inception Net (3D IN) \cite{zhang2023improved}, 2D Inception Net (2D IN) \cite{xiong2018ai}, 2D CNN \cite{wu2022convolutional}, Hybrid CNN \cite{ghaderizadeh2021hyperspectral}, Attention Graph CNN (Attention GCN) \cite{10409250}, and Spatial-Spectral Transformer \cite{ahmad2024waveformer}. The geographical maps for each model for disjoint validation, test, and complete test are presented in Figure \ref{Fig8}.}
    \resizebox{\textwidth}{!}{\begin{tabular}{c|ccc|ccc|ccc|ccc|ccc|ccc|ccc|ccc} \hline 
    
    \multirow{2}{*}{\textbf{Class}} & \multicolumn{3}{c|}{\textbf{2D CNN}} & \multicolumn{3}{c|}{\textbf{3D CNN}} & \multicolumn{3}{c|}{\textbf{Hybrid CNN}} & \multicolumn{3}{c|}{\textbf{2D IN}} & \multicolumn{3}{c|}{\textbf{3D IN}} & \multicolumn{3}{c|}{\textbf{Hybrid IN}} &  \multicolumn{3}{c|}{\textbf{Attention GCN}} & \multicolumn{3}{c}{\textbf{SSViT}} \\ \cline{2-25}

    & Va & Te & HSI & Va & Te & HSI & Va & Te & HSI & Va & Te & HSI & Va & Te & HSI & Va & Te & HSI & Va & Te & HSI & Va & Te & HSI \\ \hline 

Asphalt & 99.70 & 99.87 & 99.99 & 99.80 & 99.78 & 99.99 & 100.00 & 99.83 & 99.99 & 98.59 & 98.97 & 99.96 & 99.40 & 99.05 & 99.97 & 100.00 & 100.00 & 100.00 & 98.29 & 98.32 & 99.94 & 98.29 & 98.15 & 99.94 \\ 
Meadows & 99.96 & 99.95 & 99.96 & 99.96 & 100.00 & 99.99 & 99.96 & 99.99 & 99.99 & 99.89 & 99.96 & 99.96 & 99.96 & 100.00 & 99.99 & 99.89 & 99.91 & 99.92 & 99.82 & 99.90 & 99.89 & 100.00 & 98.91 & 99.94 \\ 
Gravel &  94.29 & 94.35 & 95.19 & 98.41 & 97.55 & 98.04 & 98.41 & 98.50 & 98.71 & 93.65 & 93.67 & 94.62 & 97.46 & 97.76 & 98.05 & 97.14 & 97.69 & 97.95 & 87.30 & 85.17 & 86.04 & 89.21 & 88.76 & 90.52\\ 
Trees &  98.91 & 99.07 & 99.18 & 99.13 & 99.39 & 99.25 & 99.13 & 99.49 & 99.51 & 96.96 & 97.86 & 98.04 & 98.26 & 98.97 & 99.02 & 99.78 & 99.86 & 99.87 & 95.43 & 96.41 & 96.57 & 97.39 & 98.60 & 98.63\\ 
Painted & 100.00 & 100.00 & 100.00 & 100.00 & 100.00 & 100.00 & 100.00 & 100.00 & 100.00 & 100.00 & 99.89 & 99.93 & 100.00 & 100.00 & 100.00 & 100.00 & 100.00 & 100.00 & 100.00 & 100.00 & 100.00 & 100.00 & 100.00 & 100.00\\ 
Soil & 99.87 & 99.20 & 99.42 & 100.00 & 100.00 & 100.00 & 100.00 & 100.00 & 100.00 & 99.34 & 98.64 & 98.95 & 100.00 & 99.94 & 99.96 & 100.00 & 100.00 & 100.00 & 100.00 & 99.43 & 99.60 & 99.73 & 99.03 & 99.28\\ 
Bitumen & 98.00 & 98.82 & 98.87 & 99.50 & 99.36 & 99.47 & 99.50 & 99.79 & 99.77 & 96.50 & 97.96 & 98.05 & 99.50 & 99.79 & 99.77 & 100.00 & 99.68 & 99.77 & 94.00 & 93.34 & 94.29 & 97.00 & 96.24 & 96.92\\ 
Bricks & 96.56 & 97.05 & 97.42 & 98.73 & 99.19 & 99.24 & 99.46 & 99.34 & 99.46 & 96.56 & 96.47 & 97.01 & 99.28 & 99.50 & 99.54 & 99.28 & 98.80 & 99.02 & 96.56 & 97.40 & 97.64 & 94.38 & 93.91 & 94.89\\ 
Shadows & 100.00 & 98.79 & 99.16 & 99.30 & 99.40 & 99.47 & 99.30 & 99.85 & 99.79 & 100.00 & 98.49 & 98.94 & 100.00 & 99.85 & 99.94 & 100.00 & 99.70 & 99.79 & 99.30 & 99.25 & 99.37 & 99.30 & 99.55 &99.58\\ \hline 

\textbf{Kappa} & 98.95 & 98.95 & 99.55 & 99.57 & 99.60 & 99.83 & 99.69 & 99.73 & 99.88 & 98.31 & 98.40 & 99.31 & 99.42 & 99.48 & 99.77 & 99.65 & 99.62 & 99.84 & 97.62 & 97.61 & 98.90 & 97.81 & 97.69 & 99.02\\
\textbf{OA} & 99.21 & 99.21 & 99.86 & 99.67 & 99.70 & 99.95 & 99.77 & 99.79 & 99.96 & 98.72 & 98.79 & 99.79 & 99.56 & 99.61 & 99.93 & 99.74 & 99.72 & 99.95 & 98.21 & 98.20 & 99.66 & 98.39 & 98.26 & 99.70\\
\textbf{AA} & 98.59 & 98.57 & 98.80 & 99.43 & 99.41 & 99.52 & 99.53 & 99.64 & 99.69 & 97.94 & 97.99 & 98.38 & 99.32 & 99.43 & 99.58 & 99.57 & 99.51 & 99.59 & 96.74 & 96.58 & 97.04 & 97.26 & 97.13 & 97.74\\
\textbf{Time (S)} & 1.05 & 2.91 & 85.48 & 1.42 & 3.52 & 90.00 & 0.80 & 5.47 & 83.94 & 1.47 & 3.81 & 85.17 & 2.98 & 11.53 & 139.04 & 2.90 & 7.13 & 105.40 & 1.80 & 5.01 & 100.10 & 2.49 & 10.61 & 144.50\\ \hline 
    \end{tabular}
    \label{Tab7}}
\end{table*}
%%%%%%%%%%%%%%%%%%%%%%%%%%

%%%%%%%%%%%%%%%%%%%%%%%%%%
\begin{figure*}[!hbt]
    \centering
	\begin{subfigure}{0.49\textwidth}
		\includegraphics[width=0.99\textwidth]{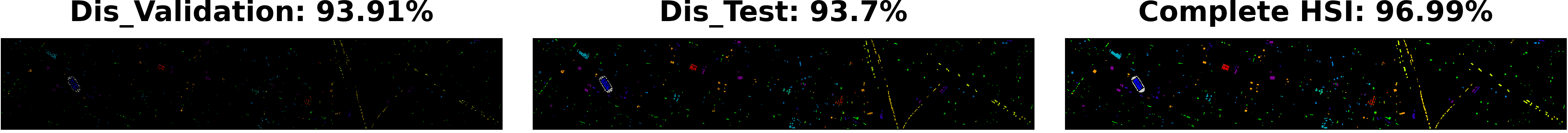}
		\caption{2D CNN} 
		\label{Fig9A}
	\end{subfigure}
	\begin{subfigure}{0.49\textwidth}
		\includegraphics[width=0.99\textwidth]{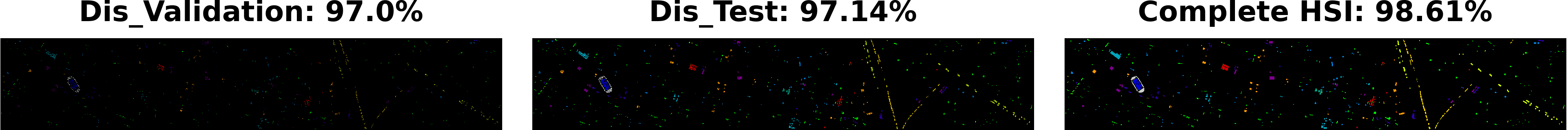}
		\caption{3D CNN}
		\label{Fig9B}
	\end{subfigure} 
	\begin{subfigure}{0.49\textwidth}
		\includegraphics[width=0.99\textwidth]{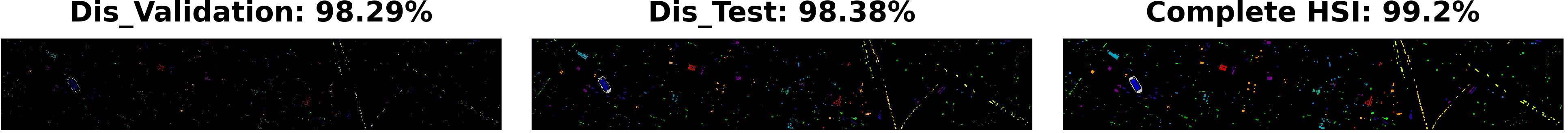}
		\caption{Hybrid CNN}
		\label{Fig9C}
	\end{subfigure} 
	\begin{subfigure}{0.49\textwidth}
		\includegraphics[width=0.99\textwidth]{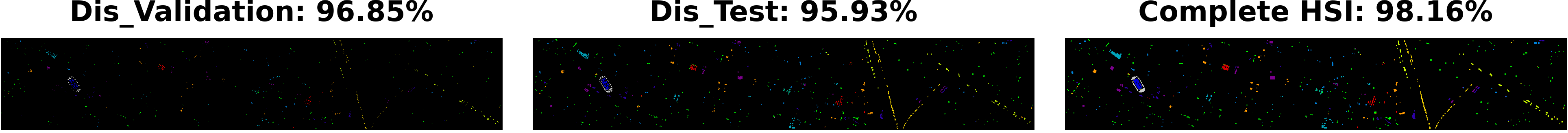}
		\caption{2D IN}
		\label{Fig9D}
	\end{subfigure} 
	\begin{subfigure}{0.49\textwidth}
		\includegraphics[width=0.99\textwidth]{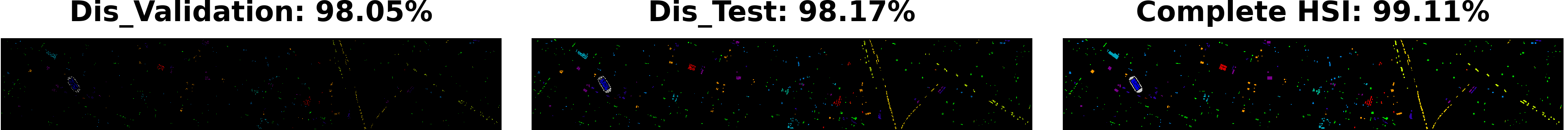}
		\caption{3D IN}
		\label{Fig9E}
	\end{subfigure} 
	\begin{subfigure}{0.49\textwidth}
		\includegraphics[width=0.99\textwidth]{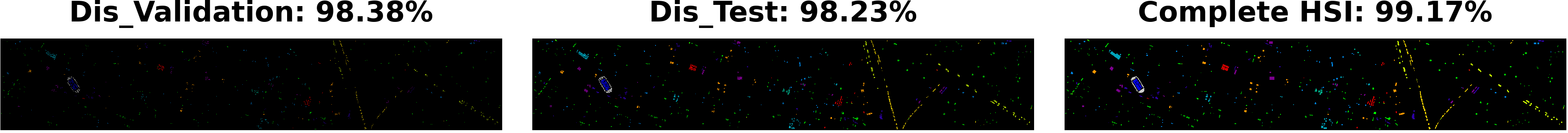}
		\caption{Hybrid IN}
		\label{Fig9F}
	\end{subfigure} 
	\begin{subfigure}{0.49\textwidth}
		\includegraphics[width=0.99\textwidth]{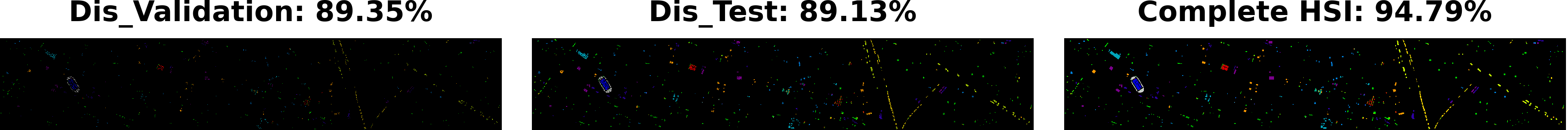}
		\caption{GCN}
		\label{Fig9G}
	\end{subfigure} 
	\begin{subfigure}{0.49\textwidth}
		\includegraphics[width=0.99\textwidth]{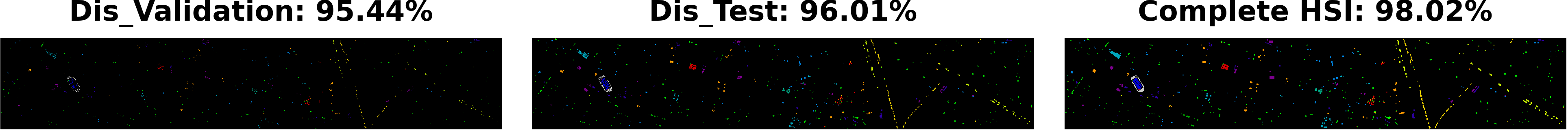}
		\caption{Transformer}
		\label{Fig9H}
	\end{subfigure} 
\caption{\textbf{University Houston Dataset:} Land cover maps for disjoint validation, test, and the entire HSI used as a test set are provided. Comprehensive class-wise results can be found in Table \ref{Tab8}.}
\label{Fig9}
\end{figure*}
%%%%%%%%%%%%%%%%%%%%%%%%%%
\begin{table*}[!t]
    \centering
    \caption{University of Houston Dataset: Per class comparative results of various SOTA models are showcased on disjoint validation and test sets. Additionally, results on the entire HSI dataset serving as the test set are also presented. The comparative methods include 3D CNN \cite{ahmad2020fast}, Hybrid Inception Net (Hybrid IN) \cite{firat2023hybrid}, 3D Inception Net (3D IN) \cite{zhang2023improved}, 2D Inception Net (2D IN) \cite{xiong2018ai}, 2D CNN \cite{wu2022convolutional}, Hybrid CNN \cite{ghaderizadeh2021hyperspectral}, Attention Graph CNN (Attention GCN) \cite{10409250}, and Spatial-Spectral Transformer \cite{ahmad2024waveformer}. The geographical maps for each model for disjoint validation, test, and complete test are presented in Figure \ref{Fig9}.}
    \resizebox{\textwidth}{!}{\begin{tabular}{c|ccc|ccc|ccc|ccc|ccc|ccc|ccc|ccc} \hline 
    
    \multirow{2}{*}{\textbf{Class}} & \multicolumn{3}{c|}{\textbf{2D CNN}} & \multicolumn{3}{c|}{\textbf{3D CNN}} & \multicolumn{3}{c|}{\textbf{Hybrid CNN}} & \multicolumn{3}{c|}{\textbf{2D IN}} & \multicolumn{3}{c|}{\textbf{3D IN}} & \multicolumn{3}{c|}{\textbf{Hybrid IN}} &  \multicolumn{3}{c|}{\textbf{Attention GCN}} & \multicolumn{3}{c}{\textbf{SSViT}} \\ \cline{2-25}

    & Va & Te & HSI & Va & Te & HSI & Va & Te & HSI & Va & Te & HSI & Va & Te & HSI & Va & Te & HSI & Va & Te & HSI & Va & Te & HSI \\ \hline 

Healthy grass &  97.33 & 96.69 & 99.99 & 97.00 & 97.95 & 99.99 & 99.33 & 99.09 & 99.99 & 98.67 & 98.63 & 99.99 & 99.33 & 98.86 & 99.99 & 100.00 & 99.54 & 99.99 & 94.33 & 94.18 & 99.99 & 98.00 & 98.17 & 99.99\\
Stressed grass &  99.67 & 99.20 & 99.36 & 99.67 & 98.86 & 99.12 & 99.67 & 98.97 & 99.20 & 99.67 & 98.97 & 99.20 & 99.67 & 99.32 & 99.44 & 99.34 & 98.41 & 98.72 & 98.67 & 98.41 & 98.56 & 99.34 & 98.75 & 98.96\\
Synthetic grass & 98.81 & 99.39 & 99.28 &95.83 & 96.31 & 96.41 & 97.02 & 97.95 & 97.85 & 95.24 & 92.83 & 93.83 & 98.21 & 98.16 & 98.28 & 98.81 & 98.57 & 98.71 & 97.62 & 97.54 & 97.70 & 97.02 & 96.11 & 96.56\\
Trees & 100.00 & 99.77 & 99.84 & 98.33 & 98.85 & 98.79 & 98.99 & 99.54 & 99.44 & 99.33 & 98.74 & 98.95 & 99.67 & 99.89 & 99.84 & 99.33 & 99.66 & 99.60 & 90.30 & 89.44 & 90.03 & 97.32 & 97.93 & 97.91\\
Soil & 100.00 & 100.00 & 100.00 & 99.66 & 100.00 & 99.92 & 100.00 & 100.00 & 100.00 & 100.00 & 99.89 & 99.92 & 100.00 & 100.00 & 100.00 & 100.00 & 100.00 & 100.00 & 99.66 & 99.77 & 99.76 & 100.00 & 99.89 & 99.92\\
Water & 78.21 & 82.46 & 82.46 & 92.31 & 95.18 & 94.77 & 97.44 & 99.12 & 98.77 & 79.49 & 85.09 & 84.62 & 92.31 & 91.23 & 92.00 & 97.44 & 97.81 & 97.85 & 75.64 & 75.44 & 76.92 & 94.87 & 95.61 & 95.69\\
Residential & 95.07 & 94.14 & 94.72 & 96.05 & 96.28 & 96.45 & 98.36 & 98.54 & 98.58 & 95.39 & 94.26 & 94.87 & 98.36 & 97.75 & 98.03 & 98.68 & 98.09 & 98.34 & 84.54 & 86.26 & 86.67 & 92.76 & 94.37 & 94.32\\
Commercial & 85.28 & 85.19 & 85.85 & 97.66 & 97.70 & 97.83 & 97.66 & 97.82 & 97.91 & 99.33 & 95.64 & 96.78 & 94.98 & 97.82 & 97.27 & 96.32 & 97.36 & 97.27 & 81.94 & 84.27 & 84.65 & 93.98 & 96.90 & 96.38\\
Road & 91.67 & 89.28 & 90.50 & 94.33 & 92.82 & 93.61 & 97.33 & 95.78 & 96.41 & 94.33 & 91.33 & 92.57 & 94.00 & 92.93 & 93.61 & 94.33 & 92.93 & 93.69 & 83.00 & 80.62 & 82.35 & 91.66 & 91.33 & 91.93\\
Highway & 93.90 & 93.83 & 94.21 & 98.98 & 98.72 & 98.86 & 95.93 & 95.81 & 95.93 & 97.63 & 96.97 & 97.31 & 99.32 & 99.30 & 99.35 & 99.32 & 99.30 & 99.35 & 91.53 & 93.95 & 93.72 & 97.63 & 97.21 & 97.47\\
Railway & 95.27 & 96.30 & 96.28 & 98.31 & 98.15 & 98.30 & 100.00 & 99.77 & 99.84 & 99.66 & 99.65 & 99.68 & 99.32 & 99.31 & 99.35 & 99.66 & 98.03 & 98.54 & 93.24 & 91.56 & 92.47 & 98.99 & 98.38 & 98.62\\
Parking Lot 1 & 94.93 & 95.49 & 95.62 & 98.99 & 99.88 & 99.68 & 98.65 & 99.54 & 99.35 & 97.64 & 98.03 & 98.05 & 98.99 & 99.77 & 99.59 & 98.99 & 99..88 & 99.68 & 95.27 & 93.52 & 94.32 & 98.65 & 99.19 & 99.10\\
Parking Lot 2 & 63.39 & 62.92 & 65.25 & 79.46 & 81.46 & 82.09 & 93.75 & 96.96 & 96.38 & 85.71 & 88.15 & 88.27 & 94.64 & 94.83 & 95.10 & 96.43 & 96.05 & 96.38 & 50.00 & 52.88 & 55.01 & 59.82 & 68.99 & 68.65\\
Tennis Court & 97.09 & 96.67 & 96.96 & 99.03 & 98.00 & 98.36 & 99.03 & 98.67 & 98.83 & 90.29 & 87.00 & 88.55 & 99.03 & 100.00 & 99.77 & 96.12 & 100.00 & 99.07 & 90.29 & 90.00 & 90.65 & 97.09 & 95.67 & 96.26\\
Running Track & 100.00 & 99.78 & 99.85 & 100.00 & 100.00 & 100.00 & 100.00 & 100.00 & 100.00 & 100.00 & 98.92 & 99.24 & 100.00 & 100.00 & 100.00 & 100.00 & 100.00 & 100.00 & 95.27 & 92.42 & 94.09 & 100.00 & 100.00 & 100.00\\ \hline 

\textbf{Kappa} & 93.91 & 93.7 & 96.99 & 97.00 & 97.14 & 98.61 & 98.29 & 98.38 & 99.20 & 96.85 & 95.93 & 98.16 & 98.05 & 98.17 & 99.11 & 98.38 & 98.23 & 99.17 & 89.35 & 89.13 & 94.79 & 95.44 & 96.01 & 98.02\\
\textbf{OA} & 94.37 & 94.18 & 99.88 & 97.23 & 97.36 & 99.94 & 98.42 & 98.50 & 99.97 & 97.09 & 96.24 & 99.92 & 98.20 & 98.31 & 99.96 & 98.50 & 98.37 & 99.97 & 90.16 & 89.96 & 99.79 & 95.79 & 96.31 & 99.92\\
\textbf{AA} & 92.71 & 92.74 & 93.34 & 96.37 & 96.68 & 96.95 & 98.21 & 98.50 & 98.56 & 95.49 & 94.94 & 95.46 & 97.86 & 97.94 & 98.11 & 98.32 & 98.37 & 99.48 & 88.23 & 88.02 & 89.13 & 94.48 & 95.23 & 95.45\\
\textbf{Time (S)} & 1.63 & 2.54 &320.98 & 0.76 & 1.58 & 323.42 & 0.47 & 1.12 & 309.30 & 0.63 & 1.38 & 362.58 & 1.78 & 4.55 & 542.21 & 1.24 & 2.85 & 449.60 & 2.07 & 2.73 & 342.90 & 2.04 & 3.76 & 533.75\\ \hline 

    \end{tabular}
    \label{Tab8}}
\end{table*}
%%%%%%%%%%%%%%%%%%%%%%%%%%

%%%%%%%%%%%%%%%%%%%%%%%%%%
\begin{table*}[!hbt]
	\centering
	\caption{Botswana Dataset: Per class comparative results of various SOTA models are showcased on disjoint validation and test sets. Additionally, results on the entire HSI dataset serving as the test set are also presented. The comparative methods include 3D CNN \cite{ahmad2020fast}, Hybrid Inception Net (Hybrid IN) \cite{firat2023hybrid}, 3D Inception Net (3D IN) \cite{zhang2023improved}, 2D Inception Net (2D IN) \cite{xiong2018ai}, 2D CNN \cite{yang2018hyperspectral}, Hybrid CNN \cite{ghaderizadeh2021hyperspectral}, Attention Graph CNN (Attention GCN) \cite{10409250}, and Spatial-Spectral Transformer \cite{ahmad2024waveformer}. The geographical maps for each model for disjoint validation, test, and complete test are presented in Figure \ref{Fig10}.}
	\resizebox{\textwidth}{!}{\begin{tabular}{c|ccc|ccc|ccc|ccc|ccc|ccc|ccc|ccc} \hline 
			
			\multirow{2}{*}{\textbf{Class}} & \multicolumn{3}{c|}{\textbf{2D CNN}} & \multicolumn{3}{c|}{\textbf{3D CNN}} & \multicolumn{3}{c|}{\textbf{Hybrid CNN}} & \multicolumn{3}{c|}{\textbf{2D IN}} & \multicolumn{3}{c|}{\textbf{3D IN}} & \multicolumn{3}{c|}{\textbf{Hybrid IN}} &  \multicolumn{3}{c|}{\textbf{Attention GCN}} & \multicolumn{3}{c}{\textbf{SSViT}} \\ \cline{2-25}
			
			& Va & Te & HSI & Va & Te & HSI & Va & Te & HSI & Va & Te & HSI & Va & Te & HSI & Va & Te & HSI & Va & Te & HSI & Va & Te & HSI \\ \hline 
			
			Water & 100.00 & 100.00 & 100.00 & 100.00 & 98.94 & 99.99 & 97.56 & 100.00 & 99.99 & 100.00 & 100.00 & 100.00 & 100.00 & 100.00 & 100.00 & 100.00 & 100.00 & 100.00 & 97.56 & 99.47 & 99.99 & 100.00 & 100.00 & 100.00\\
			Hippo Grass & 53.33 & 53.52 & 60.40 & 100.00 & 100.00 & 100.00 & 100.00 & 100.00 & 100.00 & 100.00 & 100.00 & 100.00 & 100.00 & 100.00 & 100.00 & 100.00 & 100.00 & 100.00 & 80.00 & 50.70 & 62.38 & 100.00 & 100.00 & 100.00\\
			Floodplain Grasses 1 & 100.00 & 100.00 & 100.00 & 100.00 & 100.00 & 100.00 & 100.00 & 100.00 & 100.00 & 100.00 & 100.00 & 100.00 & 100.00 & 100.00 & 100.00 & 100.00 & 100.00 & 100.00 & 89.47 & 87.50 & 89.64 & 100.00 & 100.00 & 100.00\\
			Floodplain Grasses 2 & 96.88 & 99.34 & 99.07 & 100.00 & 100.00 & 100.00 & 100.00 & 100.00 & 100.00 & 100.00 & 100.00 & 100.00 & 100.00 & 100.00 & 100.00 & 100.00 & 100.00 & 100.00 & 93.75 & 99.34 & 98.60 & 100.00 & 100.00 & 100.00\\
			Reeds 1 & 87.5 & 92.06 & 92.19 & 100.00 & 99.47 & 99.63 & 95.00 & 97.35 & 97.40 & 87.5 & 90.48 & 91.45 & 90.00 & 94.18 & 94.42 & 97.50 & 100.00 & 99.63 & 92.50 & 95.24 & 95.54 & 87.50 & 93.12 & 93.31\\
			Riparian & 82.50 & 82.54 & 85.13 & 95.00 & 94.71 & 95.54 & 87.50 & 92.06 & 92.57 & 90.00 & 95.24 & 95.17 & 95.00 & 95.24 & 95.91 & 90.00 & 94.71 & 94.80 & 87.50 & 86.77 & 88.85 & 92.50 & 94.71 & 95.17\\
			Firescar 2 & 100.00 & 98.35 & 98.84 & 100.00 & 98.90 & 99.23 & 100.00 & 100.00 & 100.00 & 100.00 & 100.00 & 100.00 & 100.00 & 100.00 & 100.00 & 100.00 & 100.00 & 100.00 & 100.00 & 100.00 & 100.00 & 100.00 & 100.00 & 100.00\\
			Island Interior & 96.67 & 98.60 & 98.52 & 100.00 & 95.10 & 96.55 & 100.00 & 96.50 & 97.54 & 100.00 & 95.80 & 97.04 & 100.00 & 100.00 & 100.00 & 100.00 & 100.00 & 100.00 & 86.67 & 93.01 & 93.10 & 100.00 & 100.00 & 100.00\\
			Woodlands & 82.98 & 85.00 & 86.94 & 97.87 & 96.36 & 97.13 & 100.00 & 100.00 & 100.00 & 100.00 & 95.45 & 96.82 & 100.00 & 100.00 & 100.00 & 100.00 & 100.00 & 100.00 & 100.00 & 98.64 & 99.04 & 100.00 & 97.73 & 98.41\\
			Acacia Shrublands & 100.00 & 100.00 & 100.00 & 100.00 & 100.00 & 100.00 & 100.00 & 100.00 & 100.00 & 100.00 & 100.00 & 100.00 & 100.00 & 100.00 & 100.00 & 100.00 & 100.00 & 100.00 & 97.30 & 88.51 & 91.53 & 100.00 & 100.00 & 100.00\\
			Acacia Grasslands & 93.48 & 97.66 & 97.38 & 100.00 & 100.00 & 100.00 & 100.00 & 100.00 & 100.00 & 93.48 & 95.79 & 96.07 & 97.83 & 99.07 & 99.02 & 100.00 & 100.00 & 100.00 & 93.48 & 85.51 & 88.85 & 97.83 & 99.07 & 99.02\\
			Short Mopane & 88.89 & 87.40 & 89.50 & 100.00 & 100.00 & 100.00 & 100.00 & 100.00 & 100.00 & 92.59 & 97.64 & 97.24 & 100.00 & 100.00 & 100.00 & 100.00 & 100.00 & 100.00 & 85.19 & 90.55 & 99.16 & 100.00 & 99.21 & 99.44\\
			Mixed Mopane & 100.00 & 100.00 & 100.00 & 100.00 & 100.00 & 100.00 & 100.00 & 100.00 & 100.00 & 100.00 & 99.47 & 99.63 & 100.00 & 100.00 & 100.00 & 100.00 & 100.00 & 100.00 & 95.00 & 94.15 & 95.15 & 100.00 & 99.47 & 99.62\\
			Exposed Soils & 92.86 & 89.55 & 91.58 & 92.86 & 97.01 & 96.84 & 100.00 & 100.00 & 100.00 & 92.85 & 92.54 & 93.68 & 92.86 & 95.52 & 95.79 & 92.86 & 97.01 & 96.84 & 71.43 & 70.15 & 74.74 & 92.86 & 97.01 & 96.84\\ \hline 
			
			\textbf{Kappa} & 91.97 & 92.96 & 96.88 & 99.11 & 98.48 & 99.39 & 98.22 & 98.81 & 99.44 & 96.66 & 97.10 & 98.72 & 98.22 & 98.81 & 99.44 & 98.66 & 99.43 & 99.7 & 91.97 & 90.48 & 96.02 & 97.77 & 98.38 & 99.26\\
			\textbf{OA} & 92.59 & 93.51 & 99.95 & 99.18 & 98.60 & 99.99 & 98.35 & 98.90 & 99.99 & 96.91 & 97.32 & 99.98 & 98.35 & 98.90 & 99.99 & 98.77 & 99.47 & 99.99 & 92.59 & 91.23 & 99.94 & 97.94 &  98.51 &99.98\\
			\textbf{AA} & 91.08 & 91.72 & 92.83 & 98.98 & 98.61 & 98.92 & 98.58 & 98.99 & 99.11 & 96.89 & 97.32 & 97.65 & 98.26 & 98.86 & 98.94 & 98.60 & 99.41 & 99.38 & 90.70 & 88.54 & 90.61 & 97.91 & 98.59 & 98.70\\
			\textbf{Time (S)} & 1.67 & 0.71 & 91.08 & 0.50 & 0.73 & 173.25 & 0.20 & 0.33 & 169.01 & 0.81 & 0.52 & 153.69 & 0.78 & 1.34 & 269.21 & 0.61 & 1.34 & 231.29 & 0.71 & 0.70 & 185.74 & 1.01 & 1.35 & 276.85\\ \hline 
			
		\end{tabular}
		\label{Tab9}}
\end{table*}
%%%%%%%%%%%%%%%%%%%%%%%%%%
\begin{figure*}[!hbt]
	\centering
	\begin{subfigure}{0.24\textwidth}
		\includegraphics[width=0.82\textwidth]{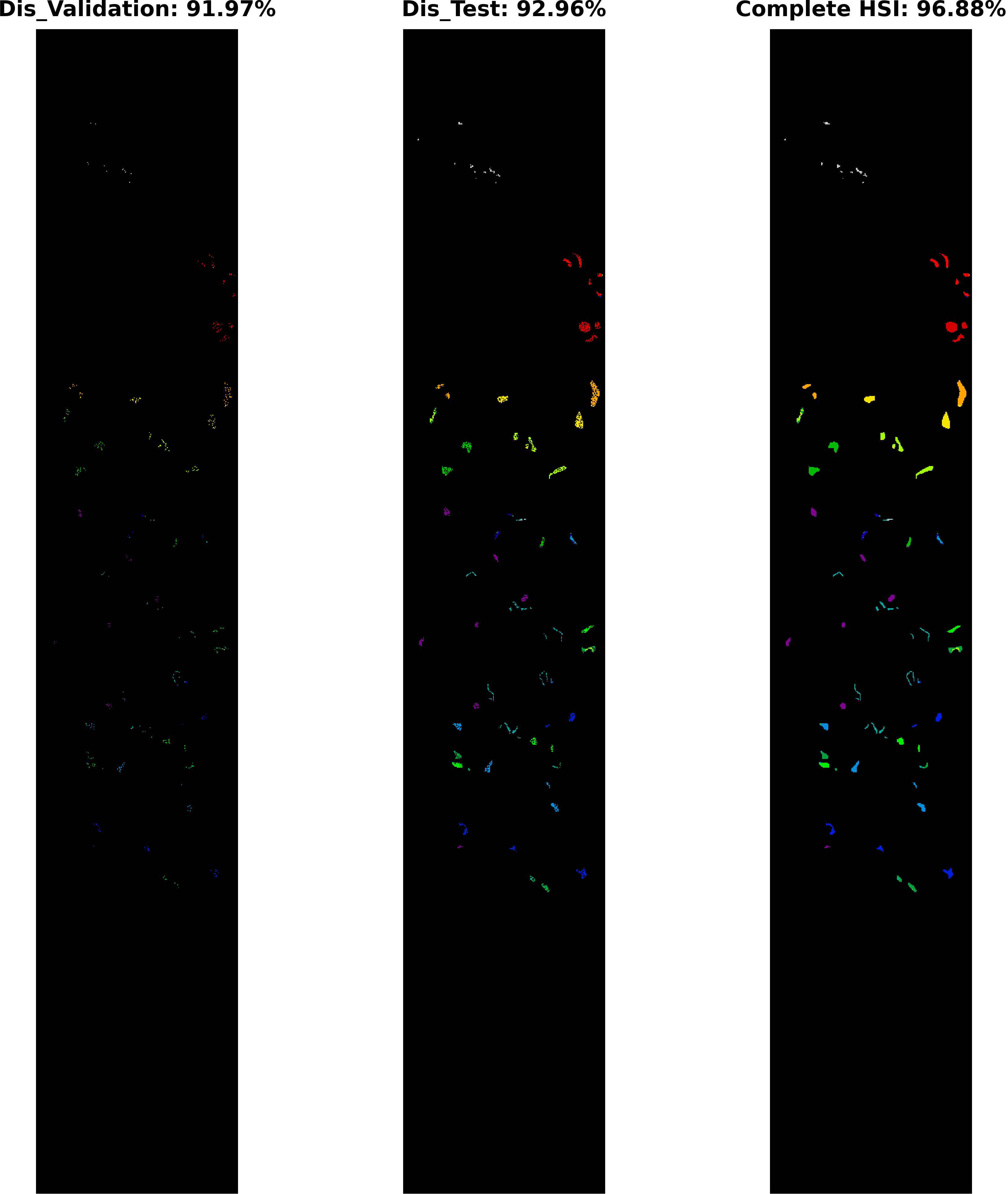}
		\caption{2D CNN} 
		\label{Fig10A}
	\end{subfigure}
	\begin{subfigure}{0.24\textwidth}
		\includegraphics[width=0.99\textwidth]{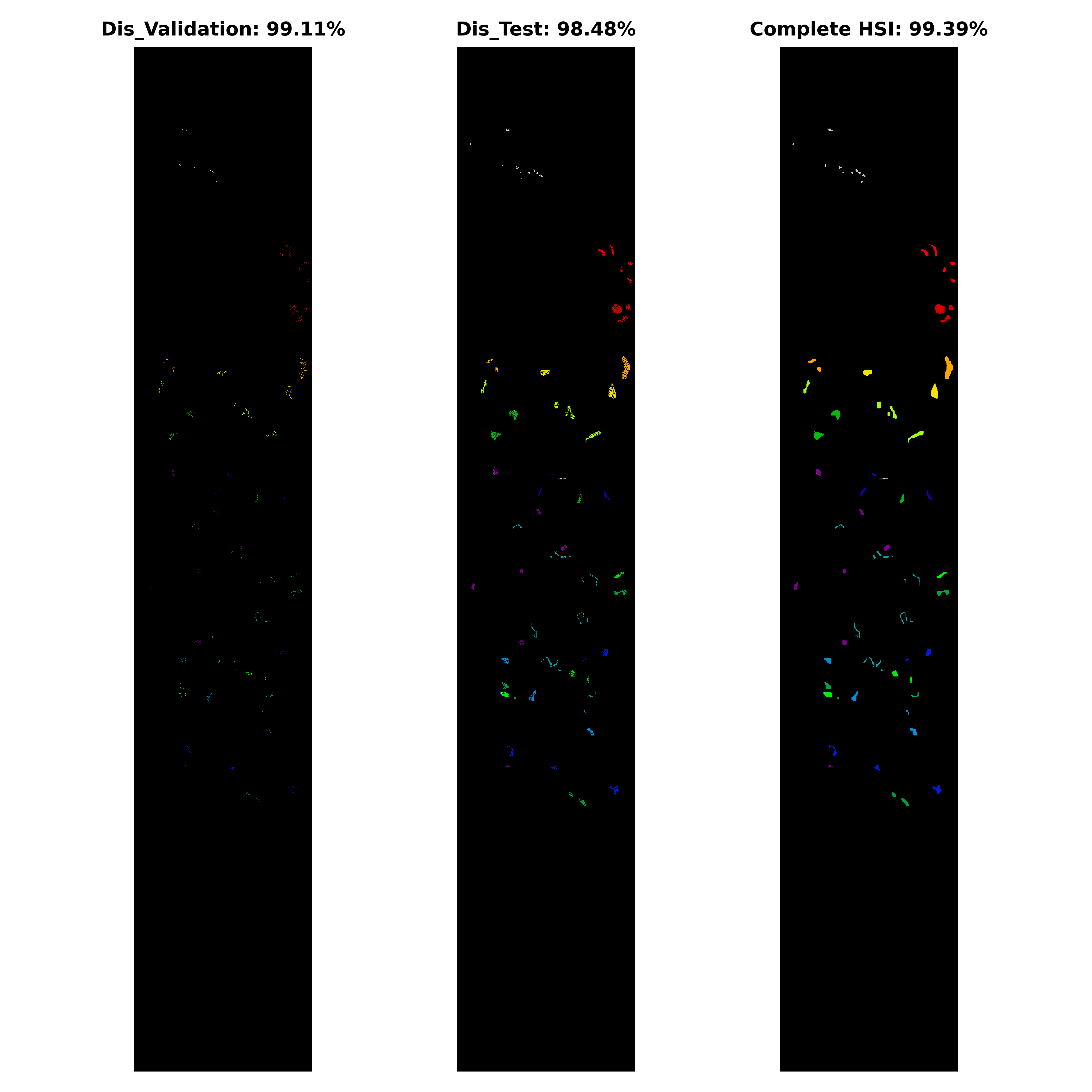}
		\caption{3D CNN}
		\label{Fig10B}
	\end{subfigure} 
	\begin{subfigure}{0.24\textwidth}
		\includegraphics[width=0.99\textwidth]{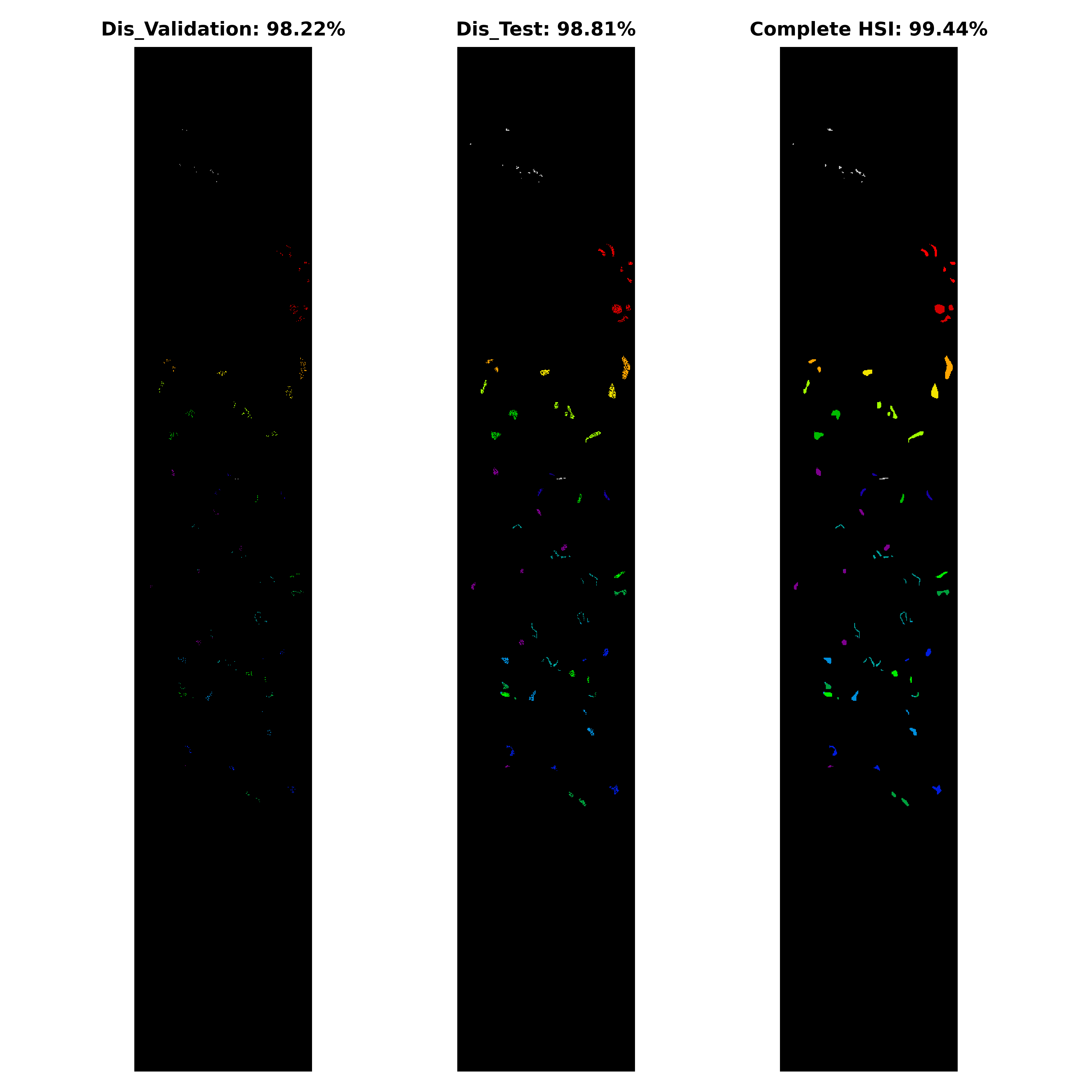}
		\caption{Hybrid CNN}
		\label{Fig10C}
	\end{subfigure} 
	\begin{subfigure}{0.24\textwidth}
		\includegraphics[width=0.99\textwidth]{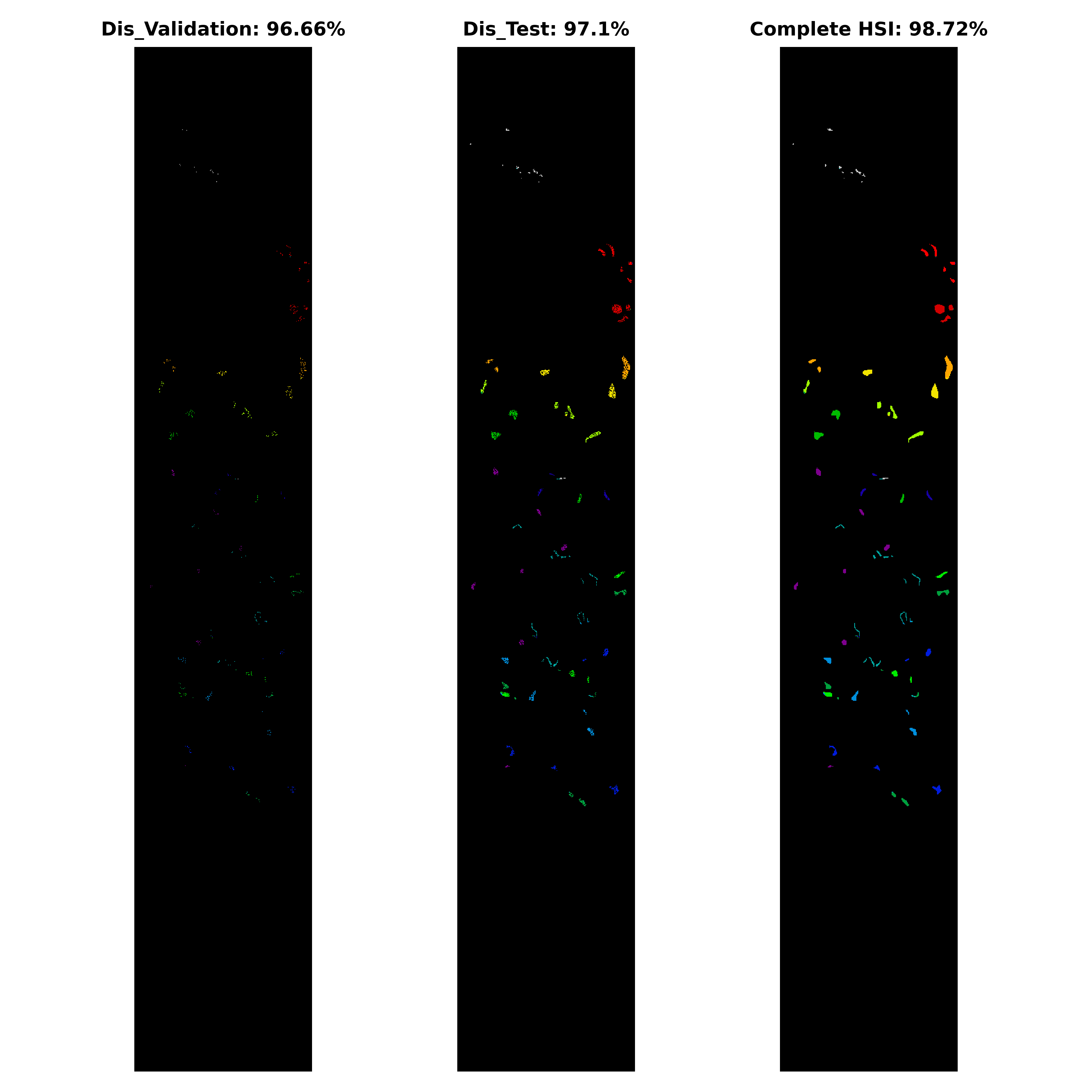}
		\caption{2D IN}
		\label{Fig10D}
	\end{subfigure} 
	\begin{subfigure}{0.24\textwidth}
		\includegraphics[width=0.99\textwidth]{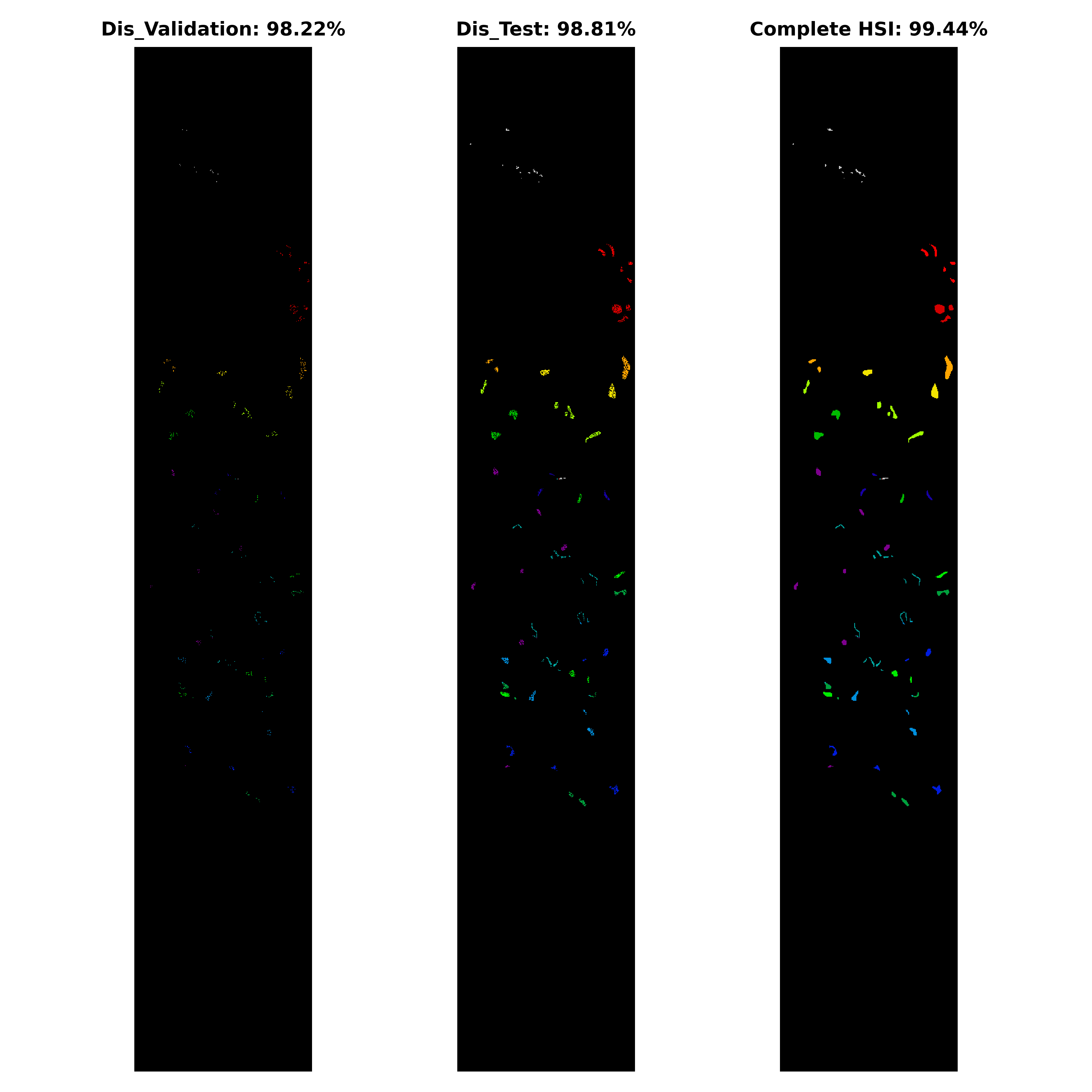}
		\caption{3D IN}
		\label{Fig10E}
	\end{subfigure} 
	\begin{subfigure}{0.24\textwidth}
		\includegraphics[width=0.99\textwidth]{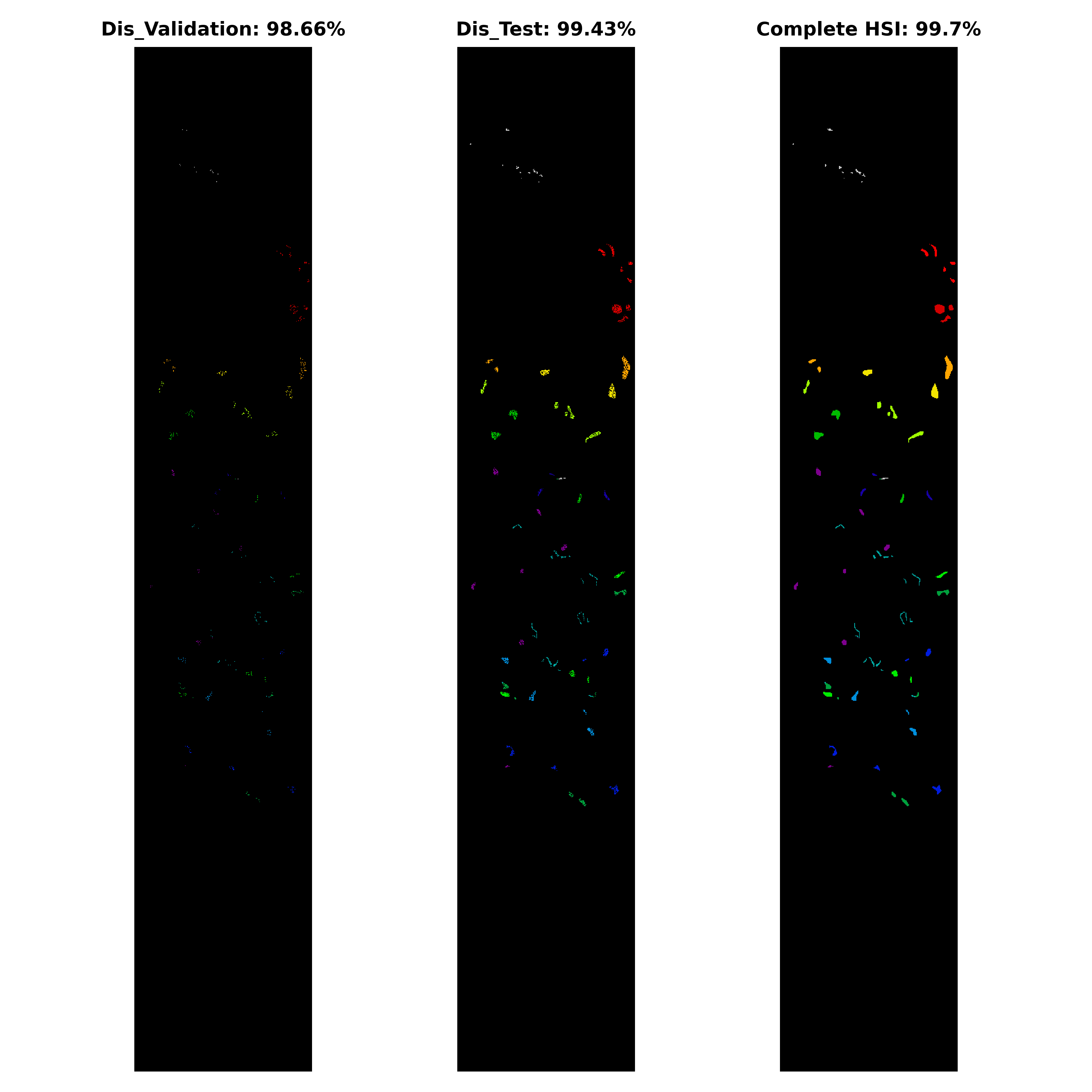}
		\caption{Hybrid IN}
		\label{Fig10F}
	\end{subfigure} 
	\begin{subfigure}{0.24\textwidth}
		\includegraphics[width=0.85\textwidth]{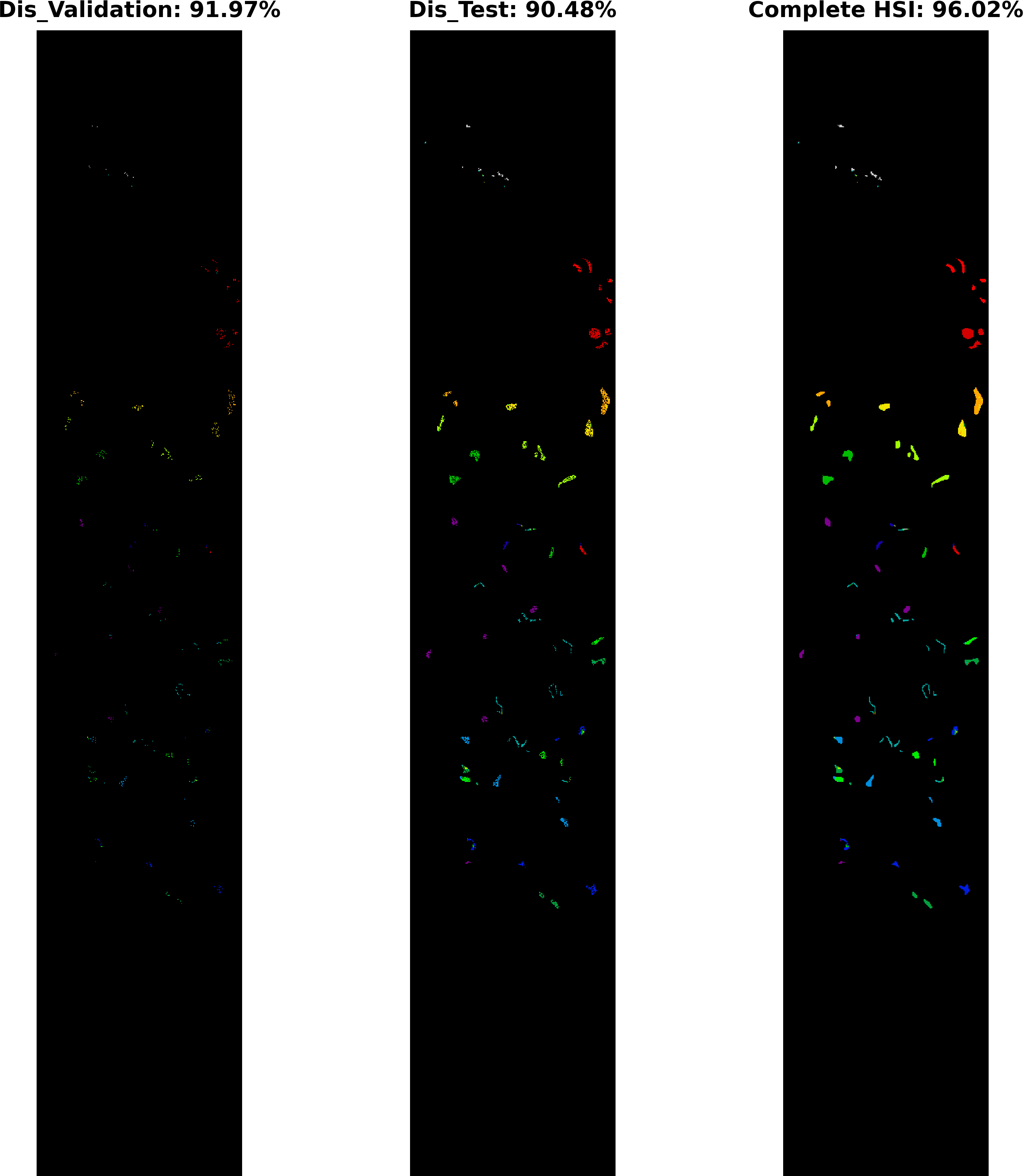}
		\caption{GCN}
		\label{Fig10G}
	\end{subfigure} 
	\begin{subfigure}{0.24\textwidth}
		\includegraphics[width=0.99\textwidth]{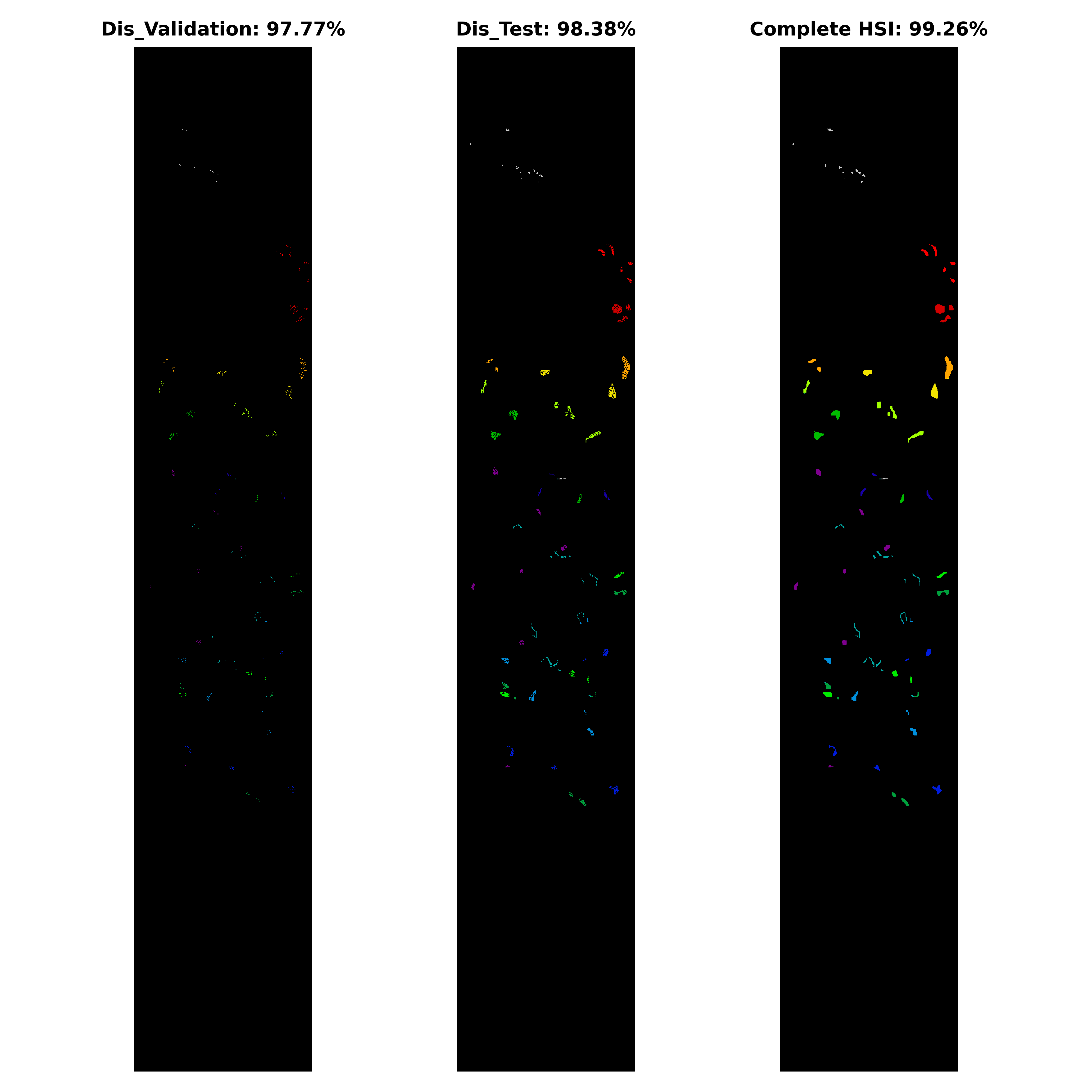}
		\caption{Transformer}
		\label{Fig10H}
	\end{subfigure} 
	\caption{\textbf{Botswana Dataset:} }
	\label{Fig10}
\end{figure*}
%%%%%%%%%%%%%%%%%%%%%%%%%%

%%%%%%%%%%%%%%%%%%%%%%%%%%
\begin{table*}[!hbt]
    \centering
    \caption{Salinas Dataset: Per class comparative results of various SOTA models are showcased on disjoint validation and test sets. Additionally, results on the entire HSI dataset serving as the test set are also presented. The comparative methods include 3D CNN \cite{ahmad2020fast}, Hybrid Inception Net (Hybrid IN) \cite{firat2023hybrid}, 3D Inception Net (3D IN) \cite{zhang2023improved}, 2D Inception Net (2D IN) \cite{xiong2018ai}, 2D CNN \cite{wu2022convolutional}, Hybrid CNN \cite{ghaderizadeh2021hyperspectral}, Attention Graph CNN (Attention GCN) \cite{10409250}, and Spatial-Spectral Transformer \cite{ahmad2024waveformer}. The geographical maps for each model for disjoint validation, test, and complete test are presented in Figure \ref{Fig11}.}
    \resizebox{\textwidth}{!}{\begin{tabular}{c|ccc|ccc|ccc|ccc|ccc|ccc|ccc|ccc} \hline 
    
    \multirow{2}{*}{\textbf{Class}} & \multicolumn{3}{c|}{\textbf{2D CNN}} & \multicolumn{3}{c|}{\textbf{3D CNN}} & \multicolumn{3}{c|}{\textbf{Hybrid CNN}} & \multicolumn{3}{c|}{\textbf{2D IN}} & \multicolumn{3}{c|}{\textbf{3D IN}} & \multicolumn{3}{c|}{\textbf{Hybrid IN}} &  \multicolumn{3}{c|}{\textbf{Attention GCN}} & \multicolumn{3}{c}{\textbf{SSViT}} \\ \cline{2-25}

    & Va & Te & HSI & Va & Te & HSI & Va & Te & HSI & Va & Te & HSI & Va & Te & HSI & Va & Te & HSI & Va & Te & HSI & Va & Te & HSI \\ \hline 

Weeds 1 & 100 & 100 & 100 & 100 & 100 & 100 & 100 & 100 & 100 & 100 & 100 & 100 & 100 & 100 & 100 & 100 & 100 & 100 & 100 & 100 & 100 & 100 & 100 & 100\\
Weeds 2 & 100 & 100 & 100 & 99.64 & 100 & 99.95 & 100 & 100 & 100 & 100 & 100 & 100 & 100 & 100 & 100 & 100 & 100 & 100 & 100 & 100 & 100 & 100 & 100 & 100\\
Fallow & 100 & 100 & 100 & 99.32 & 99.64 & 99.65 & 100 & 100 & 100 & 99.66 & 99.78 & 99.80 & 100 & 100 & 100 & 100 & 100 & 100 & 98.65 & 99.71 & 99.60 & 99.32 & 99.71 & 99.70\\
Fallow rough plow & 100 & 99.90 & 99.93 & 100 & 100 & 100 & 100 & 100 & 100 & 100 & 99.80 & 99.86 & 100 & 100 & 100 & 99.52 & 100 & 99.93 & 98.56 & 99.28 & 99.21 & 99.52 & 99.90 & 99.86\\
Fallow smooth & 97.26 & 98.19 & 98.32 & 99.25 & 99.20 & 99.33 & 98.76 & 99.20 & 99.07 & 97.26 & 97.55 & 97.87 & 99.75 & 100 & 99.96 & 99.75 & 99.95 & 99.93 & 98.01 & 97.87 & 98.21 & 98.76 & 98.51 & 98.77\\
Stubble & 100 & 100 & 100 & 100 & 100 & 100 & 100 & 100 & 100 & 100 & 100 & 100 & 100 & 100 & 100 & 100 & 100 & 100 & 100 & 100 & 100 & 100 & 100 & 100\\
Celery & 100 & 100 & 100 & 100 & 100 & 100 & 100 & 100 & 100 & 99.81 & 99.80  & 99.83 & 100 & 100 & 100 & 100 & 100 & 100 & 100 & 100 & 100 & 100 & 100 & 100\\
Grapes untrained & 99.53 & 99.21 & 99.38 & 99.59 & 99.54 & 99.62 & 99.88 & 99.68 & 99.76 & 98.11 & 98.18 & 98.45 & 99.88 & 99.85 & 99.88 & 99.82 & 99.67 & 99.74 & 98.17 & 98.16 & 98.35 & 97.99 & 97.68 & 98.09\\
Soil vinyard develop & 100 & 100 & 100 & 100 & 100 & 100 & 100 & 100 & 100 & 100 & 100 & 100 & 100 & 100 & 100 & 100 & 100 & 100 & 100 & 99.98 & 99.98 & 100 & 100 & 100\\
Corn Weeds & 100 & 100 & 100 & 100 & 100 & 100 & 100 & 100 & 100 & 100 & 100 & 100 & 100 & 99.96 & 99.97 & 100 & 100 & 100 & 100 & 99.91 & 99.94 & 100 & 100 & 100\\
Lettuce 4wk & 99.38 & 99.73 & 99.72 & 100 & 100 & 100 & 100 & 100 & 100 & 100 & 99.86 & 99.91 & 100 & 100 & 100 & 100 & 100 & 100 & 98.75 & 98.80 & 98.88 & 100 & 100 & 100\\
Lettuce 5wk & 100 & 100 & 100 & 100 & 100 & 100 & 100 & 100 & 100 & 100 & 100 & 100 & 100 & 100 & 100 & 100 & 100 & 100 & 100 & 100 & 100 & 100 & 100 & 100\\
Lettuce 6wk & 100 & 100 & 100 & 100 & 100 & 100 & 100 &  100 & 100 & 100 & 100 & 100 & 100 & 100 & 100 & 100 & 100 & 100 & 99.28 & 99.84 & 99.78 & 100 & 100 & 100\\
Lettuce 7wk & 99.38 & 100 & 99.91 & 100 & 100 & 100 & 100 & 100 & 100 & 100 & 100 & 100 & 100 & 100 & 100 & 100 & 100 & 100 & 100 & 100 & 100 & 100 & 100 & 100\\
Vinyard untrained & 97.25 & 96.91 & 97.37 & 99.82 & 99.84 & 99.86 & 99.91 & 99.86 & 99.89 & 95.41 & 95.48 & 96.15 & 99.72 & 99.88 & 99.88 & 100 & 99.72 & 99.81 & 95.60 & 94.99 & 95.46 & 97.80 & 96.76 & 97.40\\
Vinyard trellis & 99.63 & 99.53 & 99.61 & 100 & 99.84 & 99.89 & 99.26 & 99.76 & 99.72 & 99.26 & 99.68 & 99.67 & 100 & 100 & 100 & 100 & 100 & 100 & 99.89 & 99.13 & 99.22 & 100 & 100 & 100\\ \hline 

\textbf{Kappa} & 99.29 & 99.23 & 99.59 & 99.78 & 99.81 & 99.89 & 99.86 & 99.85 & 99.92 & 98.67 & 98.72 & 99.31 & 99.92 & 99.94 & 99.97 & 99.93 & 99.88 & 99.94 & 98.63 & 98.60 & 99.20 & 99.09 & 98.88 & 99.42\\
\textbf{OA} & 99.36 & 99.31 & 99.71 & 99.80 & 99.83 & 99.93 & 99.88 & 99.87 & 99.94 & 98.81 & 98.85 & 99.52 & 99.93 & 99.95 & 99.98 & 99.94 & 99.89 & 99.96 & 98.77 & 98.75 & 99.45 & 99.19 & 98.99 & 99.60\\
\textbf{AA} & 99.53 & 99.59 & 99.64 & 99.85 & 99.88 & 99.89 & 99.86 & 99.91 & 99.90 & 99.35 & 99.38 & 99.47 & 99.96 & 99.98 & 99.98 & 99.94 & 99.96 & 99.96 & 99.12 & 99.23 & 99.29 & 99.59 & 99.53 & 99.61\\
\textbf{Time (S)} & 1.23 & 5.56 & 44.41 & 1.45 & 5.56 & 47.61 & 0.82 & 5.56 & 47.67 & 2.84 & 5.67 & 43.78 & 5.33 & 14.52 & 72.51 & 2.85 & 9.15 & 61.40 & 3.07 & 5.93 & 49.99 & 2.87 & 20.92 & 78.67 \\ \hline 

    \end{tabular}
    \label{Tab10}}
\end{table*}
%%%%%%%%%%%%%%%%%%%%%%%%%%
\begin{figure*}[!hbt]
    \centering
	\begin{subfigure}{0.24\textwidth}
		\includegraphics[width=0.99\textwidth]{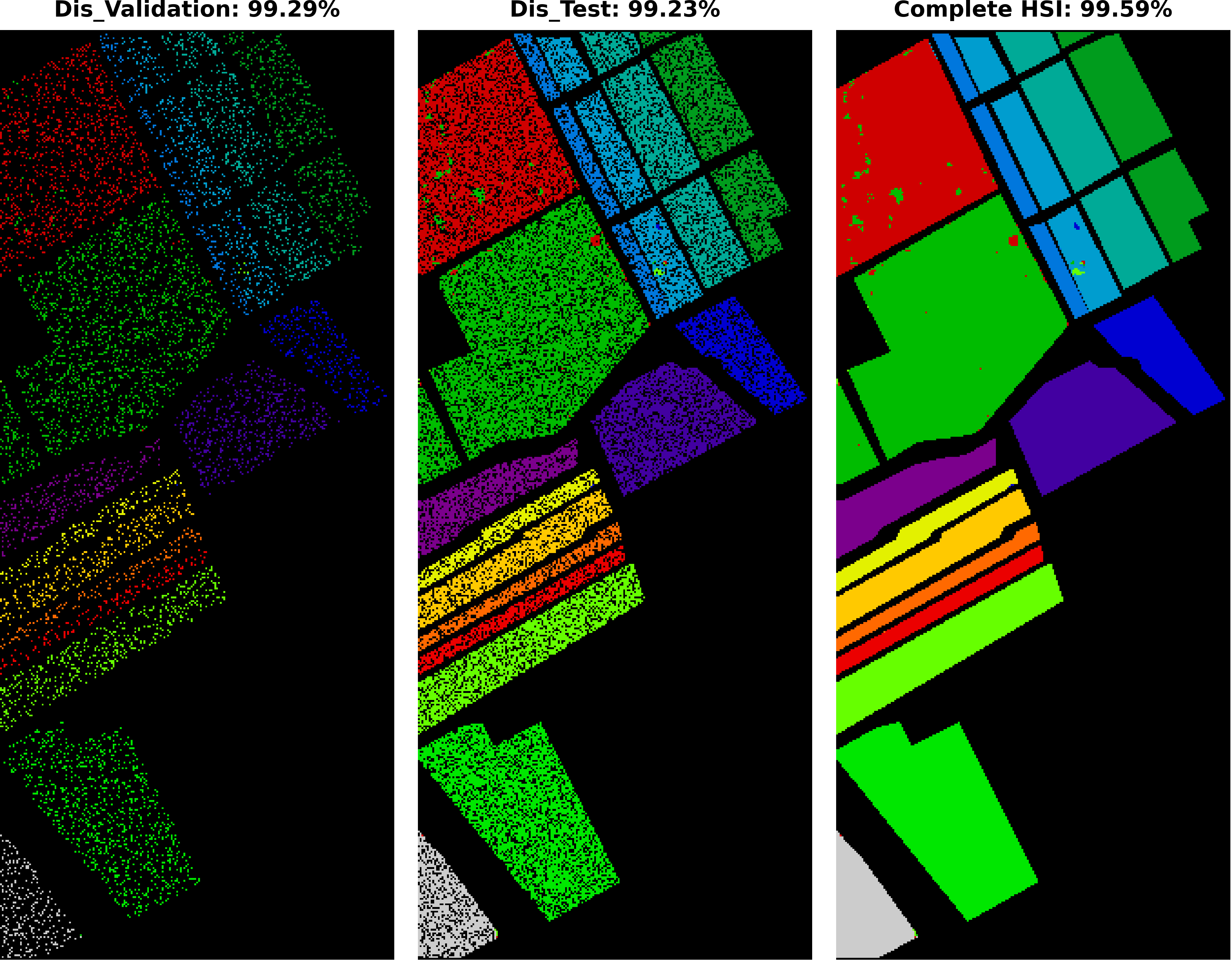}
		\caption{2D CNN} 
		\label{Fig11A}
	\end{subfigure}
	\begin{subfigure}{0.24\textwidth}
		\includegraphics[width=0.99\textwidth]{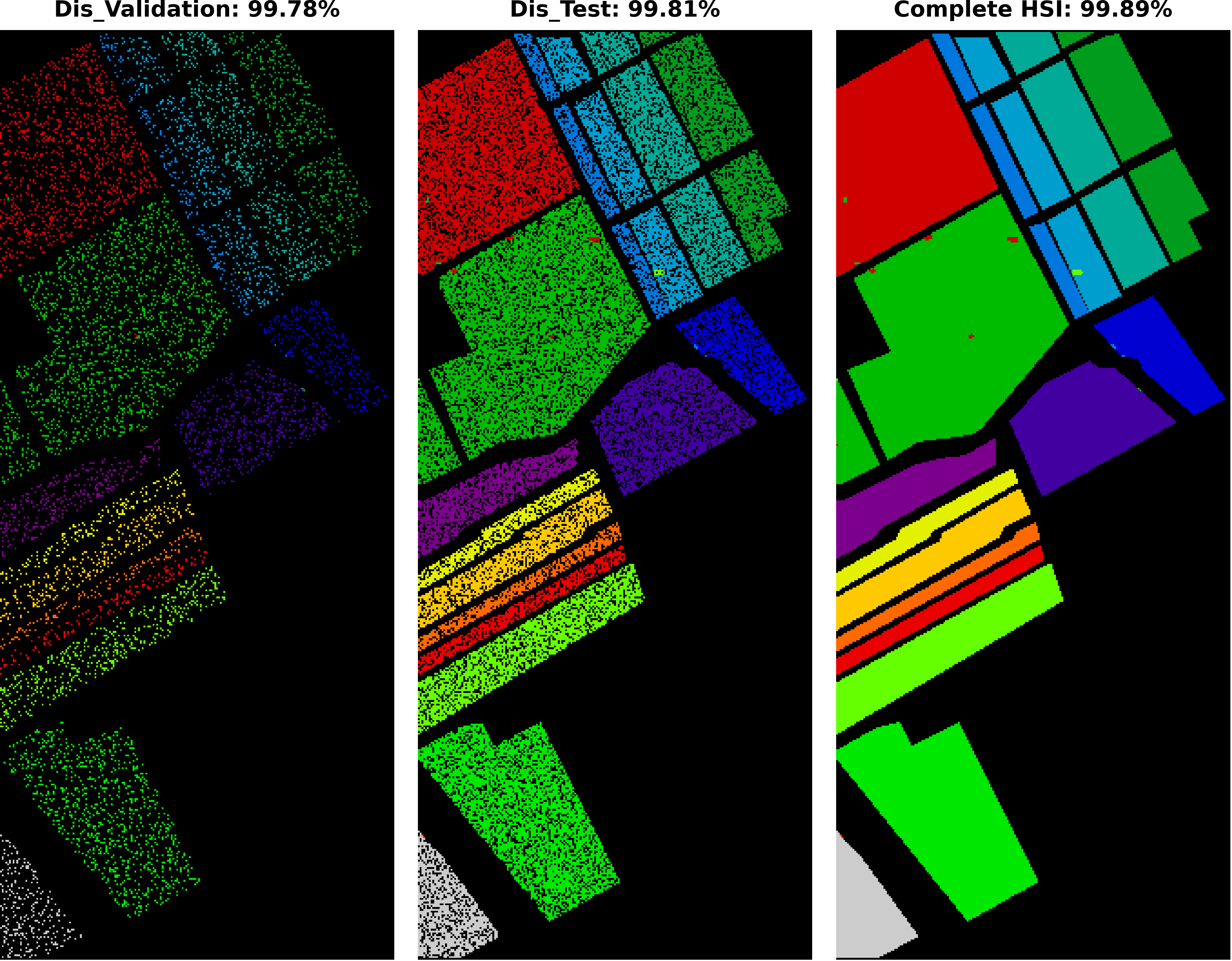}
		\caption{3D CNN}
		\label{Fig11B}
	\end{subfigure} 
	\begin{subfigure}{0.24\textwidth}
		\includegraphics[width=0.99\textwidth]{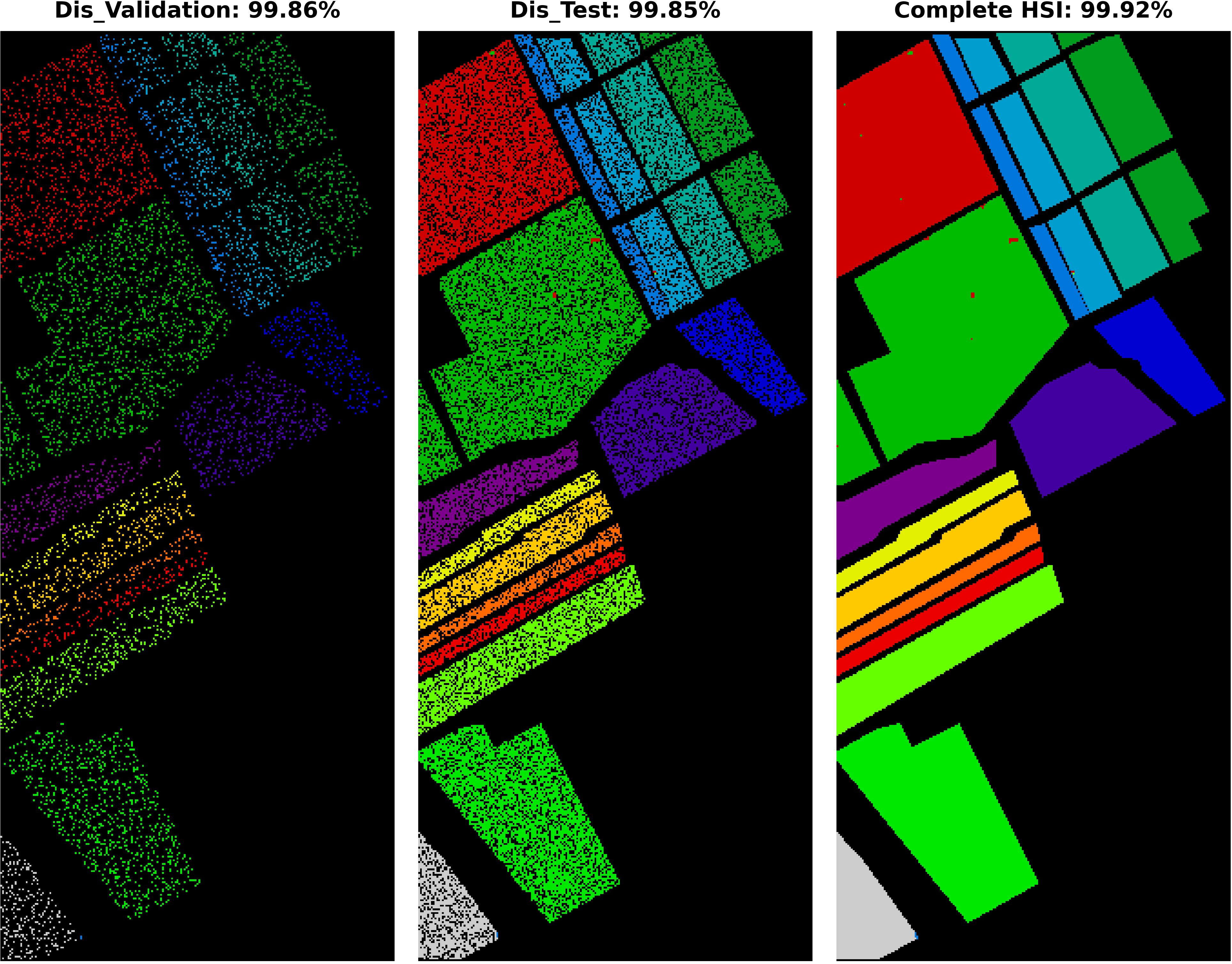}
		\caption{Hybrid CNN}
		\label{Fig11C}
	\end{subfigure} 
	\begin{subfigure}{0.24\textwidth}
		\includegraphics[width=0.99\textwidth]{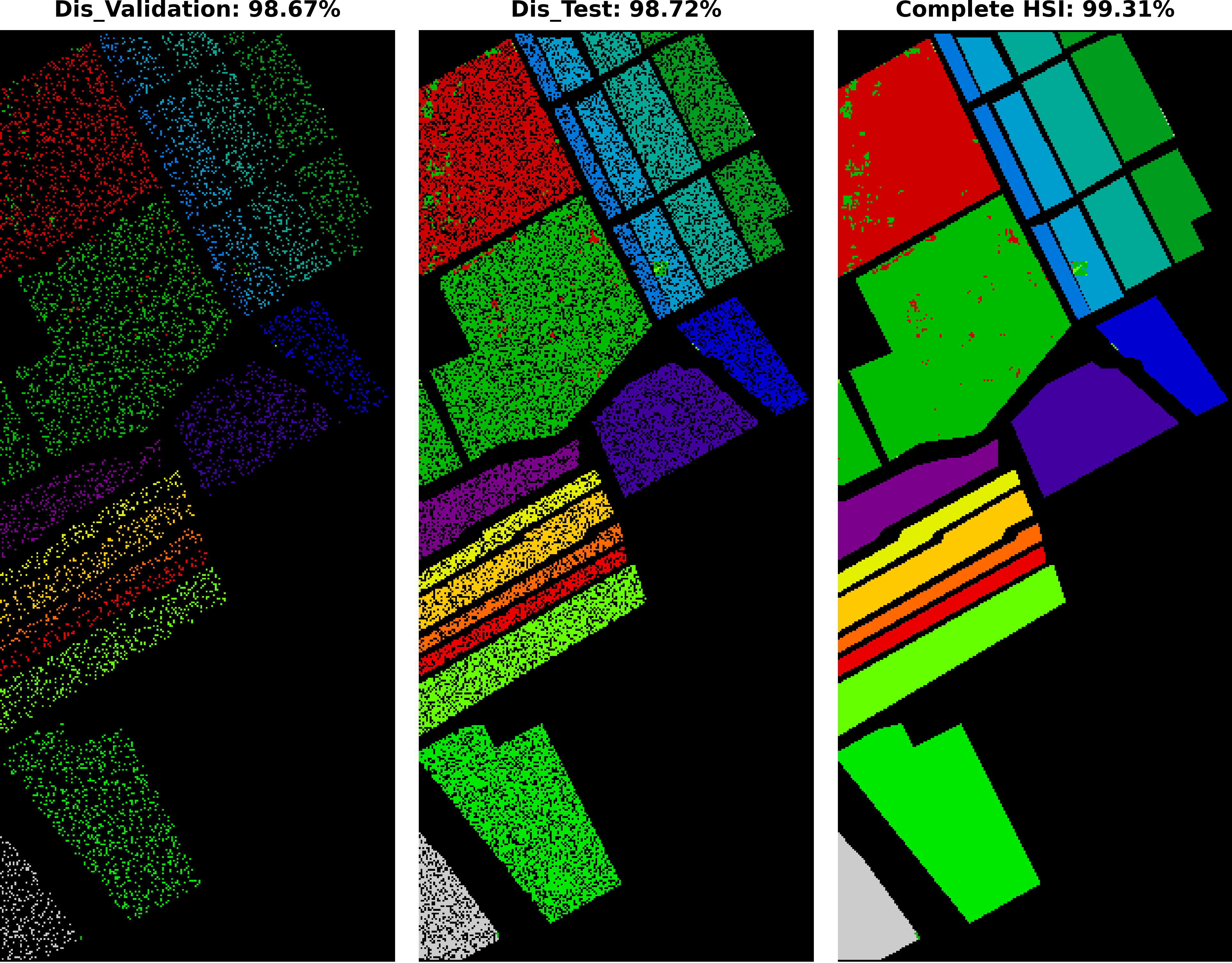}
		\caption{2D IN}
		\label{Fig11D}
	\end{subfigure} 
	\begin{subfigure}{0.24\textwidth}
		\includegraphics[width=0.99\textwidth]{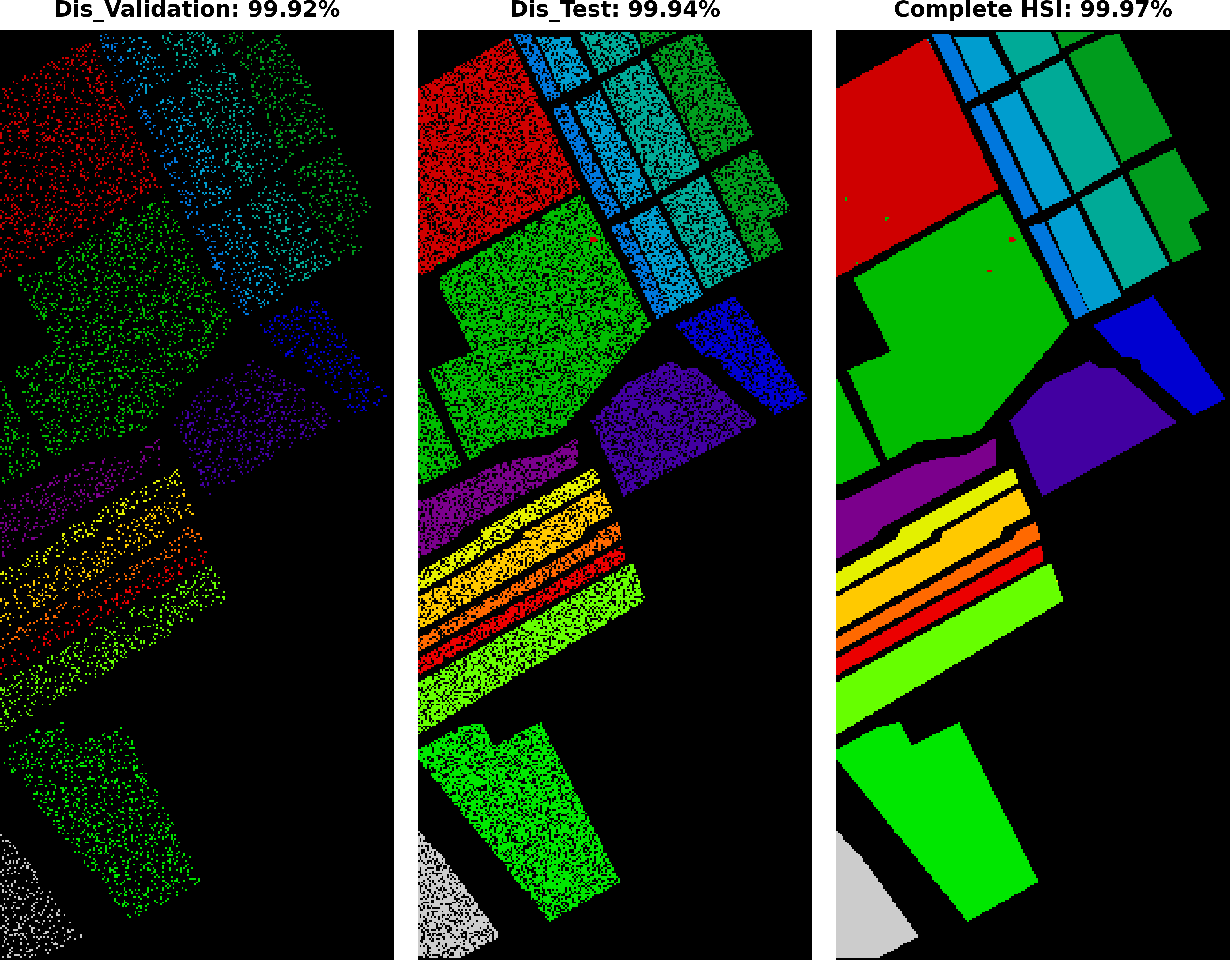}
		\caption{3D IN}
		\label{Fig11E}
	\end{subfigure} 
	\begin{subfigure}{0.24\textwidth}
		\includegraphics[width=0.99\textwidth]{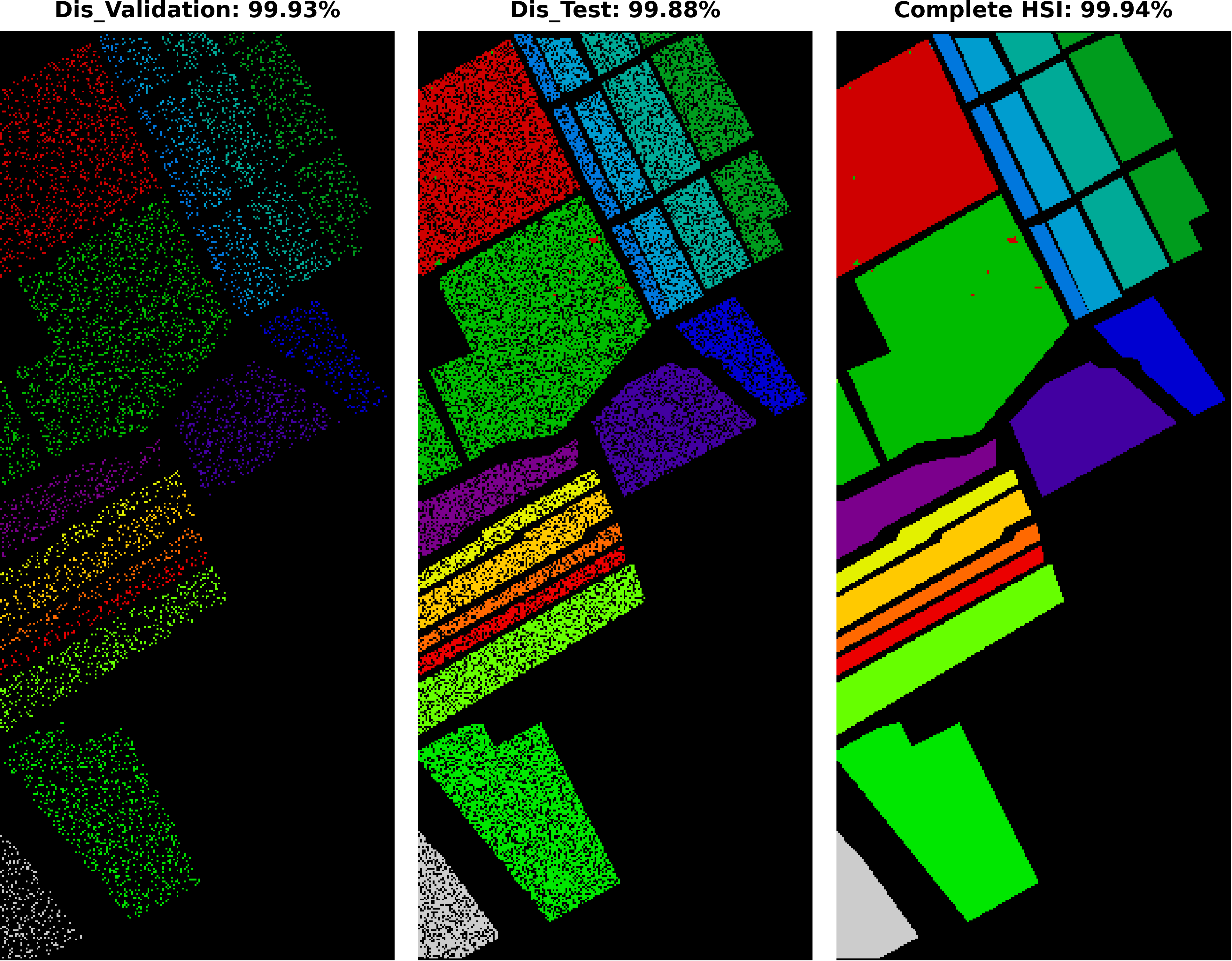}
		\caption{Hybrid IN}
		\label{Fig11F}
	\end{subfigure} 
	\begin{subfigure}{0.24\textwidth}
		\includegraphics[width=0.99\textwidth]{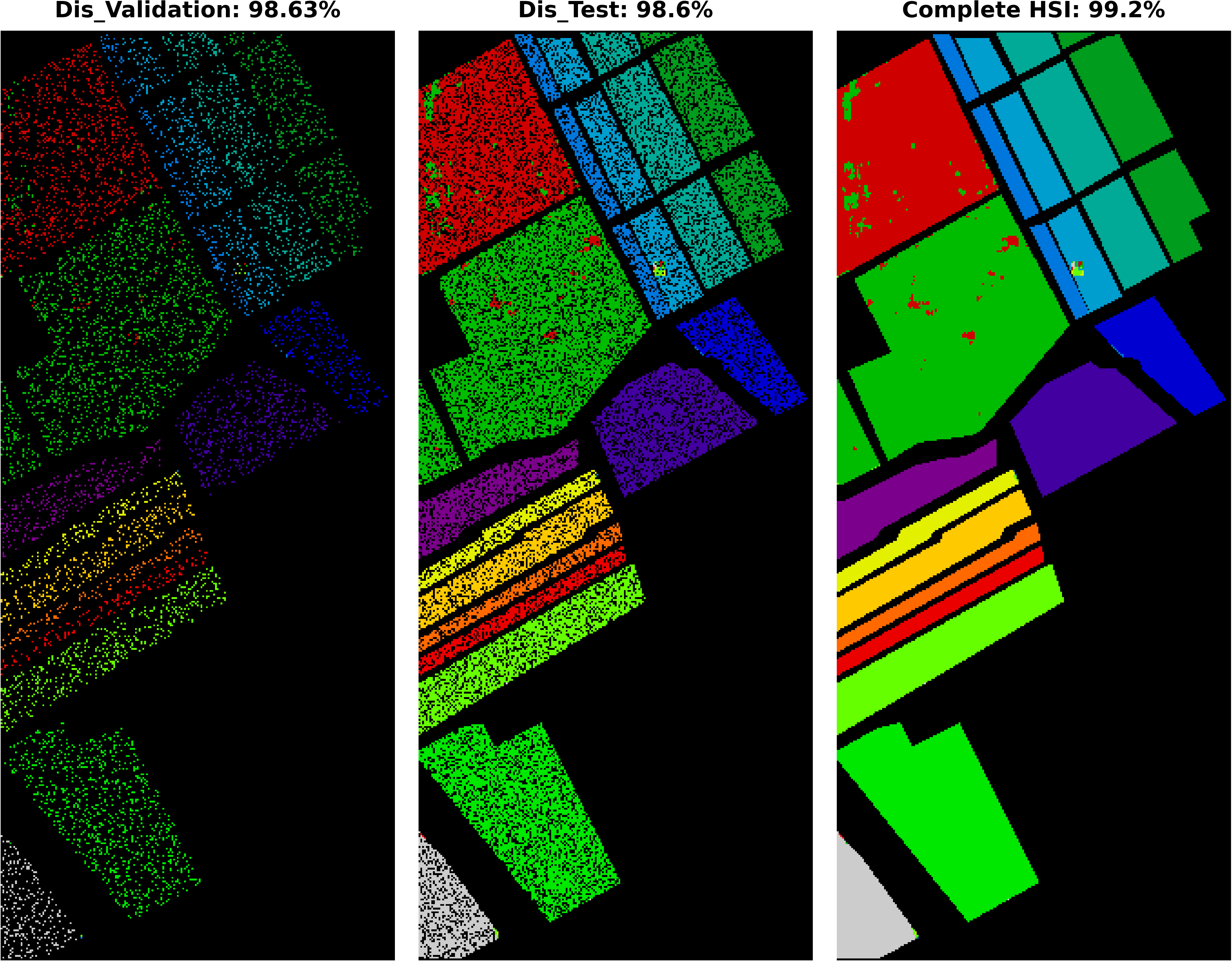}
		\caption{GCN}
		\label{Fig11G}
	\end{subfigure} 
	\begin{subfigure}{0.24\textwidth}
		\includegraphics[width=0.99\textwidth]{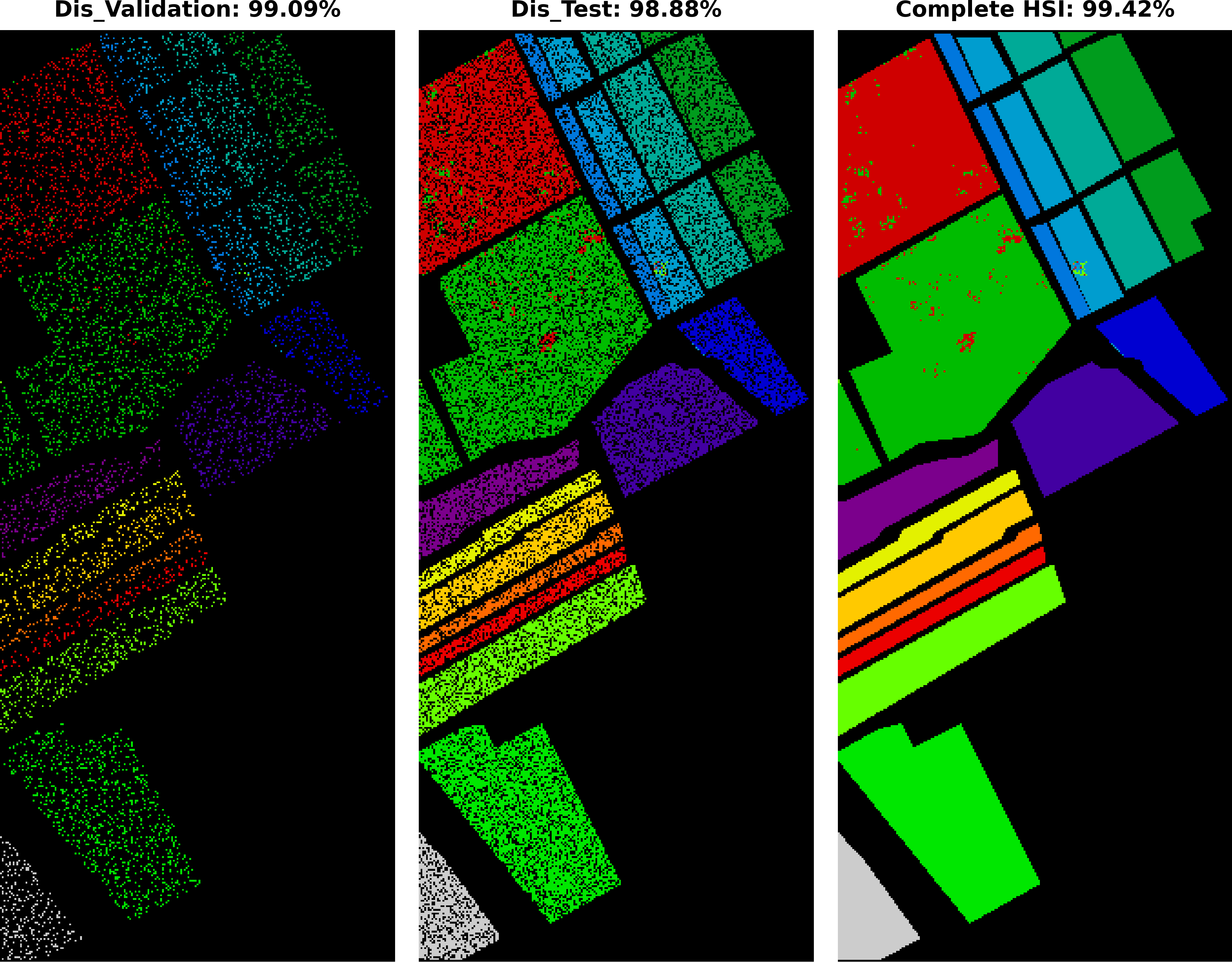}
		\caption{Transformer}
		\label{Fig11H}
	\end{subfigure} 
\caption{\textbf{Salinas Dataset:} Land cover maps for disjoint validation, test, and the entire HSI used as a test set are provided. Comprehensive class-wise results can be found in Table \ref{Tab10}.}
\label{Fig11}
\end{figure*}
%%%%%%%%%%%%%%%%%%%%%%%%%%

The hybrid Inception Net architecture consists of three blocks with the following configurations. In the first block, three 3D convolutional layers are used. The first layer has a $(5 \times 5 \times 7)$ kernel with 30 filters, the second layer uses a $(3 \times 3 \times 5)$ kernel with 20 filters, and the third layer employs a $(3 \times 3 \times 3)$ kernel with 10 filters. The same padding is applied in all three layers. Following the convolutional layer, a reshaped layer is used to convert the features from 3D to 2D. Next, a 2D max-pooling layer with a $(3 \times 3)$ filter is applied, followed by three 2D convolutional layers. Each of these layers uses a $(1 \times 1)$ kernel with 16, 32, and 64 filters, respectively. The same padding is used for all three layers. The same configuration is repeated for the second and third blocks, with the numbers of filters set to 40, 20, and 10 for the 3D convolutional layers, and 16, 32, and 64 for the 2D convolutional layers in the second block, and 60, 30, and 10 for the 3D convolutional layers, and 16, 32, and 64 for the 2D convolutional layers in the third block. Afterward, the outputs from all three blocks are concatenated, and a convolutional layer with a $(1 \times 1 \times 1)$ kernel and 128 filters is applied. Following the convolutional layer, two dense layers are deployed with a 0.4\% dropout rate. Finally, a classification layer is added with the number of output units corresponding to the number of classes in the HSI dataset.

The Attention Graph-based CNN \cite{10409250} and Transformers \cite{ahmad2024waveformer} models are trained according to the settings specified in their respective papers. The Transformer model, in particular, is used without the wavelet transformation and consists of 4 layers with 8 heads to compute the final maps. A dropout rate of 0.1 is applied to the classification layers. For more detailed information, please refer to the original papers.

%%%%%%%%%%%%%%%%%%%%%%%%%%
\subsection{Experimental Results and Discussion}

This section provides a detailed exploration of experimental results in comparison to the state-of-the-art (SOTA) works published in recent years. While many recent research endeavors present extensive experimental outcomes to highlight the strengths and weaknesses of their approaches, it is noteworthy that the experimental results in the literature may follow diverse protocols. For instance, the selection of training, validation, and test samples might be randomly done, and the percentage distribution may be identical. However, there could be variations in the geographical locations of each model, as these models may have undergone training, validation, and testing at different times. Comparative models may have been executed in multiple instances, either sequentially or in parallel, introducing a new set of training, validation, and test samples with the same number or percentage. Consequently, to ensure a fair comparison between the works proposed in the literature and the current study, it is imperative to employ identical experimental settings and execute them with the same set of training, validation, and test samples. This approach ensures a consistent and unbiased evaluation of the proposed methodologies against existing benchmarks.

A prevalent concern in the majority of recent literature is the presence of overlapping training and test samples. When training and validation samples are randomly selected, with or without considering the aforementioned point, the data split often includes overlapping samples. This situation introduces bias to the model, as overlapping implies the model has already encountered the training and validation samples, leading to inflated accuracy metrics. To prevent this issue, this study ensures that, despite the random selection of samples, the intersection between training, test, and validation samples remains consistently empty for all competing methods. This measure aims to maintain the integrity of the model evaluation process and uphold the reliability of accuracy assessments.

To ensure a robust and fair evaluation, the HSI datasets are split into disjoint training, validation, and test sets. Following the proposed method, we begin by dividing the HSI dataset into disjoint training, validation, and test sets. Each model is then trained on the training set and tuned on the validation set to optimize performance. Subsequently, the models are evaluated on the disjoint test set and the complete HSI dataset to assess their generalization capabilities. The experimental results demonstrate the effectiveness of the proposed method in improving the classification accuracy of HSIC as shown in Tables \ref{Tab6}, \ref{Tab7}, \ref{Tab8}, \ref{Tab9}, and \ref{Tab10} and Figures \ref{Fig7}, \ref{Fig8}, \ref{Fig9}, \ref{Fig10}, and \ref{Fig11}. Among the deep learning models considered, 3D CNN \cite{ahmad2020fast} and Hybrid Inception Net \cite{firat2023hybrid} achieve the highest classification accuracy, indicating their suitability for HSIC. Additionally, the results highlight the importance of using a large and diverse training dataset to achieve optimal performance.

The comparative methods frequently misclassify samples with similar spatial structures, exemplified by the misclassification of Meadows and Bare Soil classes in the Pavia University dataset, as illustrated in Figure \ref{Fig8}. Furthermore, the overall accuracy (OA) for the Grapes Untrained class is lower compared to other classes due to the aforementioned reasons as shown in Table \ref{Tab7}. In summary, higher accuracy can be attained by employing a greater number of labeled samples (complete HSI dataset as the test set), as depicted in Figures \ref{Fig7}, \ref{Fig8}, \ref{Fig9}, \ref{Fig10}, and \ref{Fig11} and Tables \ref{Tab6}, \ref{Tab7}, \ref{Tab8}, \ref{Tab9}, and \ref{Tab10}, nevertheless, the elevated accuracy is accompanied by the drawbacks of bias, redundancy, and diminished generalization performance. Tables \ref{Tab6}, \ref{Tab7}, \ref{Tab8}, \ref{Tab9}, and \ref{Tab10} also illustrate the computational time required to process and evaluate the HSI datasets used in this study. As depicted in the Tables, the time exhibits a gradual increase with the growing number of samples, i.e., Disjoint validation, disjoint test, and complete HSI dataset as a test set. 

%%%%%%%%%%%%%%%%%%%%%%%%%%
\section{Conclusion}
This paper introduces a novel technique for generating disjoint train, validation, and test splits in Hyperspectral Image Classification (HSIC). By efficiently partitioning the ground truth data, the proposed technique guarantees unbiased performance evaluations and facilitates reliable comparisons between classification models. This technique serves as a valuable tool for generating disjoint splits, ensuring that the subsets are representative of the entire dataset and the classification results are not skewed due to data leakage. While the presented technique offers significant advantages, we acknowledge that there may be limitations and opportunities for further improvements. Future research could explore alternative strategies for data splitting that consider additional factors, such as class imbalance or spatial coherence, to enhance the representativeness and generalizability of the subsets. By addressing these aspects, we can strive to develop even more robust and effective data-splitting techniques for HSIC.

%%%%%%%%%%%%%%%%%%%%%%%%%%%%%%%
\bibliographystyle{IEEEtran}
\bibliography{IEEEabrv,report}
%%%%%%%%%%%%%%%%%%%%%%%%%%%%%%%
\end{document}